\documentclass[Afour,sageh,times]{sagej}

\usepackage[utf8]{inputenc}

\usepackage[T1]{fontenc}

\usepackage[english]{babel}

\usepackage{microtype}

\usepackage[linesnumbered,vlined,ruled,commentsnumbered]{algorithm2e}

\SetVlineSkip{0.4em}

\DontPrintSemicolon{}

\SetKwIF{If}{ElseIf}{Else}{if}{}{else if}{else}{endif}

\SetAlgoSkip{}

\SetAlCapHSkip{0cm}
\usepackage{xcolor}
\definecolor{espblack}{RGB}{0,0,0}
\definecolor{espwhite}{RGB}{255,255,255}
\definecolor{espgray}{RGB}{206,206,206}
\definecolor{esplightgray}{RGB}{224,224,224}
\definecolor{espdarkgray}{RGB}{168,168,168}
\definecolor{espblue}{RGB}{11,93,174}
\definecolor{esplightblue}{RGB}{59,175,236}
\definecolor{espdarkblue}{RGB}{6,26,64}
\definecolor{espred}{RGB}{206,62,21}
\definecolor{esplightred}{RGB}{206,62,21}
\definecolor{espdarkred}{RGB}{61,19,8}
\definecolor{espyellow}{RGB}{232,163,26}
\definecolor{espgreen}{RGB}{100,161,27}
\definecolor{esplightgreen}{RGB}{149,198,35}
\definecolor{espdarkgreen}{RGB}{49,92,43}
\definecolor{esppurple}{RGB}{106,20,125}
\definecolor{esplightpurple}{RGB}{197,137,232}
\definecolor{espdarkpurple}{RGB}{50,14,59}
\usepackage{xcolor}
\definecolor{mpsblack}{RGB}{0,0,0}
\definecolor{mpswhite}{RGB}{255,255,255}
\definecolor{mpsgray}{RGB}{100,100,100}
\definecolor{mpsblue}{RGB}{11,93,174}
\definecolor{mpslightblue}{RGB}{59,175,236}
\definecolor{mpsdarkblue}{RGB}{6,26,64}
\definecolor{mpsred}{RGB}{206,62,21}
\definecolor{mpslightred}{RGB}{206,62,21}
\definecolor{mpsdarkred}{RGB}{61,19,8}
\definecolor{mpsyellow}{RGB}{232,163,26}
\definecolor{mpsgreen}{RGB}{100,161,27}
\definecolor{mpslightgreen}{RGB}{149,198,35}
\definecolor{mpsdarkgreen}{RGB}{49,92,43}
\definecolor{mpspurple}{RGB}{106,20,125}
\definecolor{mpslightpurple}{RGB}{197,137,232}
\definecolor{mpsdarkpurple}{RGB}{50,14,59}
\usepackage{tikz}

\usepackage{tikzscale}

\usepackage{pgfplots}

\usepackage{pgfplotstable}

\usetikzlibrary{calc}

\usetikzlibrary{positioning}

\usepgfplotslibrary{statistics}

\usepgfplotslibrary{fillbetween}

\pgfplotsset{
  compat=1.15,
  mps basic/.style={
    xlabel near ticks,
    xlabel style={font=\footnotesize},
    ylabel near ticks,
    ylabel style={font=\footnotesize},
    xmajorgrids,
    major x grid style={dotted},
    ymajorgrids,
    major y grid style={dotted},
    tick label style={font=\footnotesize}
  },
  mps scientific x/.style={
    x tick label style={
      /pgf/number format/sci
    }
  },
  mps scientific y/.style={
    y tick label style={
      /pgf/number format/sci
    }
  },
  mps fixed x/.style={
    x tick label style={
      /pgf/number format/.cd,
      fixed,
      fixed zerofill,
      precision=6,
      /tikz/.cd
    }
  },
  mps fixed y/.style={
    y tick label style={
      /pgf/number format/.cd,
      fixed,
      fixed zerofill,
      precision=6,
      /tikz/.cd
    }
  }
}
\usepackage{amsmath, amssymb}

\usepackage{bm}

\usepackage{mathtools}

\usepackage{etoolbox}

\usepackage{nicefrac}

\mathtoolsset{mathic}

\providecommand\given{}

\DeclarePairedDelimiterX\set[1]{\{}{\}}{%
  \renewcommand\given{\SetSymbol[\delimsize]}%
  \ifblank{#1}{\:\ldots\:}{#1}
}

\DeclarePairedDelimiterX{\abs}[1]{\vert}{\vert}{
  \ifblank{#1}{\:\cdot\:}{#1}
}

\DeclarePairedDelimiterX{\norm}[1]{\lVert}{\rVert}{
  \ifblank{#1}{\:\cdot\:}{#1}
}
\DeclarePairedDelimiterXPP{\lnorm}[1]{}{\lVert}{\rVert}{_{2}}{
  \ifblank{#1}{\:\cdot\:}{#1}
}
\DeclarePairedDelimiterXPP{\infnorm}[1]{}{\lVert}{\rVert}{_{\infty}}{
  \ifblank{#1}{\:\cdot\:}{#1}
}
\DeclarePairedDelimiterXPP{\pnorm}[2]{}{\lVert}{\rVert}{
  \ifblank{#2}{_{p}}{_{#2}}}{
  \ifblank{#1}{\:\cdot\:}{#1}
}

\DeclarePairedDelimiterX{\inner}[2]{\langle}{\rangle}{
  \ifblank{#1}{\:\cdot\:}{#1},\ifblank{#2}{\:\cdot\:}{#2}
}

\DeclarePairedDelimiterXPP{\expv}[1]{\mathbb{E}}{[}{]}{}{
  \renewcommand\given{\nonscript\:\delimsize\vert\nonscript\:\mathopen{}}
  \ifblank{#1}{\:\cdot\:}{#1}
}
\usepackage{amsthm}

\newcounter{mpsdefinition}
\newtheorem{definition}[mpsdefinition]{Definition}

\usepackage{caption}

\usepackage{subcaption}

\usepackage{adjustbox}

\usepackage{rotating}

\usepackage{placeins}
\input{preamble/mpsmiscellaneous}

\usepackage{amsmath, amssymb}

\usepackage{bm}

\usepackage{mathtools}

\mathtoolsset{mathic}

\usepackage{etoolbox}

\usepackage{nicefrac}

\usepackage{xparse}

\newcommand{\DeclareVariable}[2]{
  \NewDocumentCommand#1{ o o } {
    \IfNoValueTF{##1} {
      #2
    } {
      \ifblank{##1} {
        \IfNoValueTF{##2} {
          #2
        } {
          {#2}^{##2}
        }
      } {
        \IfNoValueTF{##2} {
          {#2}_{##1}
        } {
          {#2}_{##1}^{##2}
        }
      }
    }
  }
}

\NewDocumentCommand{\DeclareNullaryFunction}%
{ m m O{\left\lparen} O{\right\rparen} }{
  \NewDocumentCommand#1{ s o o } {
    \IfBooleanTF{##1} {
      \IfNoValueTF{##2} {
        {#2}
      } {
        \IfNoValueTF{##3} {
          {#2}_{##2}
        } {
          {#2}_{##2}^{##3}
        }
      }
    } {
      \IfNoValueTF{##2} {
        #2#3#4
      } {
        \IfNoValueTF{##3} {
            {#2}_{##2}#3 #4
        } {
            {#2}_{##2}^{##3}#3 #4
        }
      }
    }
  }
}

\NewDocumentCommand{\DeclareUnaryFunction}%
{ m m O{\left\lparen} O{\right\rparen} }{
  \NewDocumentCommand#1{ s o o m } {
    \IfBooleanTF{##1} {
      \IfNoValueTF{##2} {
        {#2}
      } {
        \IfNoValueTF{##3} {
          {#2}_{##2}
        } {
          {#2}_{##2}^{##3}
        }
      }
    } {
      \IfNoValueTF{##2} {
        \ifblank{##4}{
          #2#3 \;\cdot\; #4
        } {
          #2#3 ##4 #4
        }
      } {
        \IfNoValueTF{##3} {
          \ifblank{##4}{
            {#2}_{##2}#3 \;\cdot\;#4
          } {
            {#2}_{##2}#3 ##4 #4
          }
        } {
          \ifblank{##4}{
            {#2}_{##2}^{##3}#3 \;\cdot\; #4
          } {
            {#2}_{##2}^{##3}#3 ##4 #4
          }
        }
      }
    }
  }
}

\NewDocumentCommand{\DeclareBinaryFunction}%
{ m m O{\left\lparen} O{\right\rparen} }{
  \NewDocumentCommand#1{ s o o m m } {
    \IfBooleanTF{##1} {
      \IfNoValueTF{##2} {
        {#2}
      } {
        \IfNoValueTF{##3} {
          {#2}_{##2}
        } {
          {#2}_{##2}^{##3}
        }
      }
    } {
      \IfNoValueTF{##2} {
        \ifblank{##4}{
          \ifblank{##5}{
            #2#3\;\cdot\;,\;\cdot\;#4
          } {
            #2#3\;\cdot\;, ##5#4
          }
        } {
          \ifblank{##5}{
            #2#3##4,\;\cdot\;#4
          } {
            #2#3##4, ##5#4
          }
        }
      } {
        \IfNoValueTF{##3} {
          \ifblank{##4}{
            \ifblank{##5}{
              {#2}_{##2}#3\;\cdot\;,\;\cdot\;#4
            } {
              {#2}_{##2}#3\;\cdot\;, ##5#4
            }
          } {
            \ifblank{##5}{
              {#2}_{##2}#3##4,\;\cdot\;#4
            } {
              {#2}_{##2}#3##4, ##5#4
            }
          }
        } {
          \ifblank{##4}{
            \ifblank{##5}{
              {#2}_{##2}^{##3}#3\;\cdot\;,\;\cdot\;#4
            } {
              {#2}_{##2}^{##3}#3\;\cdot\;, ##5#4
            }
          } {
            \ifblank{##5}{
              {#2}_{##2}^{##3}#3##4,\;\cdot\;#4
            } {
              {#2}_{##2}^{##3}#3##4, ##5#4
            }
          }
        }
      }
    }
  }
}

\newcommand{\DeclareOperator}[2]{
  \NewDocumentCommand#1{ m } {
    \mathop{}\!\mathrm{#2}##1
  }
}

\DeclareVariable{\adsolcostsrc}{\state[\mathrm{s}][\hat{s}]}
\DeclareVariable{\adsolcosttgt}{\state[\mathrm{t}][\hat{s}]}
\DeclareVariable{\atdedgequeue}{\queue[\edges][\mathrm{ATD}^{*}]}
\DeclareVariable{\atdvertexqueue}{\queue[\vertices][\mathrm{ATD}^{*}]}
\DeclareVariable{\bestfwdsrcstate}{\state[\mathrm{s}, \fwdsymbol]}
\DeclareVariable{\bestfwdtgtstate}{\state[\mathrm{t}, \fwdsymbol]}
\DeclareVariable{\bestrevsrcstate}{\state[\mathrm{s}, \revsymbol]}
\DeclareVariable{\bestrevtgtstate}{\state[\mathrm{t}, \revsymbol]}
\DeclareVariable{\bestsrcstate}{\state[\mathrm{s}][*]}
\DeclareVariable{\beststate}{\state[][*]}
\DeclareVariable{\besttgtstate}{\state[\mathrm{t}][*]}
\DeclareVariable{\cdresolution}{d}
\DeclareVariable{\childstate}{\state[\mathrm{c}]}
\DeclareVariable{\closedrevvertices}{\vertices[\revsymbol, \mathrm{closed}]}
\DeclareVariable{\closedvertices}{\vertices[\mathrm{closed}]}
\DeclareVariable{\currentcost}{\cost*[\mathrm{current}][]{}}
\DeclareVariable{\dimension}{n}
\DeclareVariable{\edges}{E}
\DeclareVariable{\fwdedges}{\edges[\fwdsymbol]}
\DeclareVariable{\fwdqueue}{\queue[\fwdsymbol]}
\DeclareVariable{\fwdsymbol}{\mathcal{F}}
\DeclareVariable{\fwdvertices}{\vertices[\fwdsymbol]}
\DeclareVariable{\goalstates}{\states[\mathrm{goal}]}
\DeclareVariable{\goalstate}{\state[\mathrm{goal}]}
\DeclareVariable{\inadremeffortsrc}{\state[\mathrm{s}][\bar{r}]}
\DeclareVariable{\inadremefforttgt}{\state[\mathrm{t}][\bar{r}]}
\DeclareVariable{\inadsolcostsrc}{\state[\mathrm{s}][\bar{s}]}
\DeclareVariable{\inadsolcosttgt}{\state[\mathrm{t}][\bar{s}]}
\DeclareVariable{\inconsvertices}{\vertices[\mathrm{inconsistent}]}
\DeclareVariable{\inflationfactor}{\varepsilon_{\mathrm{i}}}
\DeclareVariable{\informedset}{\states[\hat{f}]}
\DeclareVariable{\invedges}{\edges[\mathrm{invalid}]}
\DeclareVariable{\invstates}{\states[\mathrm{invalid}]}
\DeclareVariable{\key}{\mathtt{key}}
\DeclareVariable{\lbcost}{\cost*[\mathrm{lb}][]{}}
\DeclareVariable{\motionarg}{t}
\DeclareVariable{\motions}{\Sigma}
\DeclareVariable{\neighborvertices}{\vertices[\mathrm{neighbors}]}
\DeclareVariable{\numinformedstates}{q}
\DeclareVariable{\numstatesperbatch}{m}
\DeclareVariable{\optimalcost}{\cost*[][*]{}}
\DeclareVariable{\optimalmotion}{\motion*[][*]{}}
\DeclareVariable{\outedges}{\edges[\mathrm{out}]}
\DeclareVariable{\outvertices}{\vertices[\mathrm{out}]}
\DeclareVariable{\parentstate}{\state[\mathrm{p}]}
\DeclareVariable{\pathlength}{l}
\DeclareVariable{\queue}{\mathcal{Q}}
\DeclareVariable{\revedges}{\edges[\revsymbol]}
\DeclareVariable{\revqueue}{\queue[\revsymbol]}
\DeclareVariable{\revsymbol}{\mathcal{R}}
\DeclareVariable{\revvertices}{\vertices[\revsymbol]}
\DeclareVariable{\sampledstates}{\states[\mathrm{sampled}]}
\DeclareVariable{\searchtree}{\treesymbol}
\DeclareVariable{\smalldist}{\delta}
\DeclareVariable{\sourcestate}{\state[\mathrm{s}]}
\DeclareVariable{\sparsecdresolution}{d}
\DeclareVariable{\startstate}{\state[\mathrm{start}]}
\DeclareVariable{\states}{X}
\DeclareVariable{\state}{\bm{\mathrm{x}}}
\DeclareVariable{\targetstate}{\state[\mathrm{t}]}
\DeclareVariable{\treesymbol}{\mathcal{T}}
\DeclareVariable{\truncationfactor}{\varepsilon_{\mathrm{t}}}
\DeclareVariable{\valstates}{\states[\mathrm{valid}]}
\DeclareVariable{\vertices}{V}

\DeclareUnaryFunction{\adctc}{\hat{g}}
\DeclareUnaryFunction{\adctglabel}{\hat{h}}[\left\lbrack][\right\rbrack]
\DeclareUnaryFunction{\adctg}{\hat{h}}
\DeclareUnaryFunction{\adrevctc}{\hat{h}}[\left\lbrack][\right\rbrack]
\DeclareUnaryFunction{\adsolcost}{\hat{f}}
\DeclareUnaryFunction{\aitrevkey}{\key[\revsymbol][\mathrm{AIT}^{*}]\hspace*{-0.4em}}
\DeclareUnaryFunction{\atdvertexkey}{\key[\vertices][\mathrm{ATD}^{*}]\hspace*{-0.4em}}
\DeclareUnaryFunction{\clearance}{\delta}
\DeclareUnaryFunction{\closure}{\mathrm{closure}}
\DeclareUnaryFunction{\concost}{\hat{h}_{\mathrm{con}}}[\left\lbrack][\right\rbrack]
\DeclareUnaryFunction{\cost}{c}
\DeclareUnaryFunction{\expcost}{\hat{h}_{\mathrm{exp}}}[\left\lbrack][\right\rbrack]
\DeclareUnaryFunction{\fwdctc}{g_{\fwdsymbol}}
\DeclareUnaryFunction{\homotopy}{H}
\DeclareUnaryFunction{\hyperball}{B}
\DeclareUnaryFunction{\inadctglabel}{\bar{h}}[\left\lbrack][\right\rbrack]
\DeclareUnaryFunction{\inadefforttocome}{\bar{d}}
\DeclareUnaryFunction{\inadefforttogolabel}{\bar{e}}[\left\lbrack][\right\rbrack]
\DeclareUnaryFunction{\inadrevctc}{\bar{h}}[\left\lbrack][\right\rbrack]
\DeclareUnaryFunction{\inadrevetc}{\bar{e}}[\left\lbrack][\right\rbrack]
\DeclareUnaryFunction{\measure}{\lambda}
\DeclareUnaryFunction{\motion}{\sigma}
\DeclareUnaryFunction{\parentctc}{g_{\mathrm{p}}}[\left\lbrack][\right\rbrack]
\DeclareUnaryFunction{\prob}{P}
\DeclareUnaryFunction{\revctc}{g_{\revsymbol}}
\DeclareUnaryFunction{\revtotpotsolcost}{f_{\revsymbol}}
\DeclareUnaryFunction{\se}{\mathrm{SE}}
\DeclareUnaryFunction{\totalvariation}{\mathrm{TV}}
\DeclareUnaryFunction{\treectc}{g_{\treesymbol}}

\DeclareBinaryFunction{\adedgecost}{\hat{c}}
\DeclareBinaryFunction{\adsolcostlabel}{\hat{s}}
\DeclareBinaryFunction{\aitfwdkey}{\key[\fwdsymbol][\mathrm{AIT}^{*}]\hspace*{-0.4em}}
\DeclareBinaryFunction{\edgecost}{c}
\DeclareBinaryFunction{\edge}{}
\DeclareBinaryFunction{\eitfwdkey}{\key[\fwdsymbol][\mathrm{EIT}^{*}]\hspace*{-0.4em}}
\DeclareBinaryFunction{\eitrevkey}{\key[\revsymbol][\mathrm{EIT}^{*}]\hspace*{-0.4em}}
\DeclareBinaryFunction{\inadedgecost}{\bar{c}}
\DeclareBinaryFunction{\inadedgeeffort}{\bar{e}}
\DeclareBinaryFunction{\inadremeffortlabel}{\bar{r}}
\DeclareBinaryFunction{\inadsolcostlabel}{\bar{s}}
\DeclareBinaryFunction{\fwdtotpotsolcost}{f_{\fwdsymbol}}

\DeclareOperator{\diff}{d}

\usepackage[linesnumbered,vlined,ruled,commentsnumbered]{algorithm2e}

\SetVlineSkip{0.4em}

\SetInd{0.1em}{0.5em}

\DontPrintSemicolon{}

\SetKwIF{If}{ElseIf}{Else}{if}{}{else if}{else}{endif}

\SetAlgoSkip{}

\SetAlCapHSkip{0cm}

\SetAlgoCaptionLayout{small}

\SetCommentSty{mpsalgohidecommentfont}

\SetNlSty{texttt}{}{}

\usepackage{etoolbox}

\makeatletter
\patchcmd\algocf@Vline{\vrule}{\vrule \kern-0.4pt}{}{}
\patchcmd\algocf@Vsline{\vrule}{\vrule \kern-0.4pt}{}{}
\makeatother

\makeatletter
\patchcmd{\@algocf@start}% <cmd>
  {-1.5em}% <search>
  {0pt}% <replace>
  {}{}% <success><failure>
\makeatother

\newcommand{\neareststates}[1]{\texttt{nearest}\left(#1, r \text{ or } k\right)}
\newcommand{\setadd}{\raisebox{0pt}[\height-2pt]{ \(\xleftarrow{\scriptscriptstyle +} \) }}
\newcommand{\setsubtract}{\raisebox{0pt}[\height-2pt]{ \( \xleftarrow{\scriptscriptstyle -} \) }}

\DeclareNullaryFunction{\improveapproximation}{\texttt{improve\_approximation}}
\DeclareNullaryFunction{\continuefwdsearch}{\texttt{continue\_forward\_search}}
\DeclareNullaryFunction{\continuerevsearch}{\texttt{continue\_reverse\_search}}
\DeclareNullaryFunction{\iteraterevsearch}{\texttt{iterate\_reverse\_search}}
\DeclareNullaryFunction{\iteratefwdsearch}{\texttt{iterate\_forward\_search}}
\DeclareNullaryFunction{\initializequeues}{\texttt{initialize\_queues}}
\DeclareNullaryFunction{\updateinflationfactor}{\texttt{update\_inflation\_factor}}
\DeclareNullaryFunction{\updatetruncationfactor}{\texttt{update\_truncation\_factor}}
\DeclareNullaryFunction{\updatecdresolution}{\texttt{update\_sparse\_cd\_resolution}}

\DeclareUnaryFunction{\parent}{\texttt{parent}}
\DeclareUnaryFunction{\fwdparent}{\texttt{parent}_{\mathcal{F}}}
\DeclareUnaryFunction{\revparent}{\texttt{parent}_{\mathcal{R}}}
\DeclareUnaryFunction{\fwdchildren}{\texttt{children}_{\mathcal{F}}}
\DeclareUnaryFunction{\revchildren}{\texttt{children}_{\mathcal{R}}}
\DeclareUnaryFunction{\treechildren}{\texttt{children}_{\mathcal{T}}}
\DeclareUnaryFunction{\samplestates}{\texttt{sample}}
\DeclareUnaryFunction{\prunestates}{\texttt{prune}}
\DeclareUnaryFunction{\neighborstates}{\texttt{neighbors}}
\DeclareUnaryFunction{\expandstate}{\texttt{expand}}
\DeclareUnaryFunction{\expandedge}{\texttt{expand}}
\DeclareUnaryFunction{\updateheuristic}{\texttt{update\_heuristic}}
\DeclareUnaryFunction{\updateheuristics}{\texttt{update\_heuristics}}
\DeclareUnaryFunction{\updatestate}{\texttt{update\_state}}
\DeclareUnaryFunction{\getbestedge}{\texttt{get\_best\_edge}}
\DeclareUnaryFunction{\getbestfwdedge}{\texttt{get\_best\_forward\_edge}}
\DeclareUnaryFunction{\getbestrevedge}{\texttt{get\_best\_reverse\_edge}}
\DeclareUnaryFunction{\getbestvertex}{\texttt{get\_best\_vertex}}
\DeclareUnaryFunction{\lowerbound}{\texttt{lower\_bound}}
\DeclareUnaryFunction{\isvalid}{\texttt{collision\_free}}
\DeclareUnaryFunction{\isconsistent}{\texttt{is\_consistent}}
\DeclareUnaryFunction{\couldbevalid}{\texttt{no\_sparse\_collisions\_detected}}
\DeclareUnaryFunction{\expandormarkinconsistentstate}{\texttt{expand\_or\_mark\_inconsistent}}
\DeclareUnaryFunction{\invalidaterevbranch}{\texttt{invalidate\_rev\_branch}}

\DeclareBinaryFunction{\continuesearch}{\texttt{continue\_search}}

\acrodef{AA*}{Adaptive A*}
\acrodef{A-MHA*}{Anytime Multi-Heuristic A*}
\acrodef{ABIT*}{Advanced BIT*}
\acrodef{AEES}{Anytime Explicit Estimation Search}
\acrodef{AIT*}{Adaptively Informed Trees}
\acrodef{ARA*}{Anytime Repairing A*}
\acrodef{BEAST}{Bayesian Effort-Aided Search Trees}
\acrodef{BIT*}{Batch Informed Trees}
\acrodef{BVP}{Boundary Value Problem}
\acrodef{CL-RRTsharp}[CL-RRT\textsuperscript{\#}]{Closed Loop-RRT\textsuperscript{\#}}
\acrodef{DWA*}{Dynamically Weighted A*}
\acrodef{EES}{Explicit Estimation Search}
\acrodef{EIT*}{Effort Informed Trees}
\acrodef{FCL}{Flexible Collision Library}
\acrodef{FMT*}{Fast Marching Trees}
\acrodef{GBRRT}{Generalized Bidirectional RRT}
\acrodef{GLS}{Generalized Lazy Search}
\acrodef{HA*}{Hierarchical A*}
\acrodef{HCA*}{Hierarchical Cooperative A*}
\acrodef{WHCA*}{Windowed HCA*}
\acrodef{hRRT}{Heuristically-Guided RRT}
\acrodef{IST}{Informed Subdivision Tree}
\acrodef{LazySP}{Lazy Shortest Path}
\acrodef{LBT-RRT}{Lower Bound Tree-RRT}
\acrodef{LPA*}{Lifelong Planning A*}
\acrodef{LRHA*}{Lazy Receding Horizon A*}
\acrodef{MPLB}{Motion Planning using Lower Bounds}
\acrodef{OBB}{Oriented Bounding Box}
\acrodef{OMPL}{Open Motion Planning Library}
\acrodef{OpenRAVE}{Open Robotics Automation Virtual Environment}
\acrodef{PRM}{Probabilistic Roadmaps}
\acrodef{PRM*}{PRM*}
\acrodef{QMP}{Quotient-space RoadMap Planner}
\acrodef{QRRT}{Quotient-space RRT}
\acrodef{RIOT}{Region Informed Optimal Trees}
\acrodef{RGG}{Random Geometric Graph}
\acrodef{RRA*}{Reverse Resumable A*}
\acrodef{RRG}{Rapidly-exploring Random Graphs}
\acrodef{RRT}{Rapidly-exploring Random Trees}
\acrodef{RRT*}{RRT*}
\acrodef{RRTsharp}[RRT\textsuperscript{\#}]{RRT\textsuperscript{\#}}
\acrodef{RRTX}[RRT\textsuperscript{X}]{RRT\textsuperscript{X}}
\acrodef{RSS}{Rectangle Swept Sphere}
\acrodef{SBL}{Single-Query, Bi-Directional, Lazy Collision Detection}
\acrodef{TLPA*}{Truncated LPA*}
\acrodef{UKR}{Uni\-compartmental Knee Replacement}
\acrodef{WAM}{Whole-Arm Manipulator}

\usetikzlibrary{decorations.pathreplacing,shapes.misc}
\tikzset{
  cross/.style={cross out, draw=black, minimum size=2*(#1-\pgflinewidth), inner sep=0pt, outer sep=0pt},
  cross/.default={4pt},
  start/.style={draw = black, fill = black, circle, inner sep = 0pt, minimum size=3pt},
  goal/.style={draw = black, fill = white, circle, inner sep = 0pt, minimum size=3pt},
  vertex/.style={draw = black, fill = black, circle, inner sep = 0pt, minimum size=0.4pt},
  rgg edge/.style={densely dotted, very thin, espblack},
  forward edge/.style={espblack, thin},
  reverse edge/.style={espgray, thin},
  invalid edge/.style={espblack, thick, densely dotted},
  obstacle/.style={fill = espdarkgray, draw = espdarkgray},
  antiobstacle/.style={draw = none, fill = white},
  boundary/.style={draw = black, fill = none},
  solution/.style={espyellow, very thick}
}

\newcommand{\tikzinlinestart}{\,\protect\tikz[baseline]{\protect\node[start, yshift = 0.25em] {};}\,}
\newcommand{\tikzinlinegoal}{\,\protect\tikz[baseline]{\protect\node[goal, yshift = 0.25em] {};}\,}
\newcommand{\tikzinlinestate}{\,\protect\tikz[baseline]{\protect\node[vertex, yshift = 0.25em] {};}\,}
\newcommand{\tikzinlineobstacle}{\,\protect\tikz[baseline]{\protect\draw[obstacle, yshift = -0.15em] (0,0) rectangle (0.4em, 0.8em);}\,}

\newcommand{\tikzinlinereverseedge}{\,\protect\tikz[baseline]{\protect\draw[reverse edge, yshift = 0.25em] (0,0) -- (1.1em, 0);}\,}
\newcommand{\tikzinlineforwardedge}{\,\protect\tikz[baseline]{\protect\draw[forward edge, yshift = 0.25em] (0,0) -- (1.1em, 0);}\,}
\newcommand{\tikzinlinesolution}{\,\protect\tikz[baseline]{\protect\draw[solution, yshift = 0.25em] (0,0) -- (1.1em, 0);}\,}

\newcommand{\successplotheight}{0.45\textwidth}
\newcommand{\costplotheight}{0.55\textwidth}

\usetikzlibrary{external}
\graphicspath{{/home/marlin/phd/EITstar/}}

\usepackage{booktabs}
\usepackage{multirow}

\usepackage{ifthen}

\usepackage[hidelinks]{hyperref}

\usepackage{xparse}
\NewDocumentCommand\reft{ o m }{\hyperref[#2]{\IfNoValueTF{#1}{\ref*{#2}}{#1~\ref*{#2}}}}%
\NewDocumentCommand\refp{ o m
}{\hyperref[#2]{\IfNoValueTF{#1}{(\ref*{#2})}{(#1~\ref*{#2})}}}%
\NewDocumentCommand\subreft{ o m }{\hyperref[#2]{\IfNoValueTF{#1}{\subref*{#2}}{#1~\subref*{#2}}}}%
\NewDocumentCommand\subrefp{ o m }{\hyperref[#2]{\IfNoValueTF{#1}{(\subref*{#2})}{(#1~\subref*{#2})}}}%

\NewDocumentCommand\lnereft{ m }{\reft[line]{#1}}

\usepackage[title]{appendix}

\begin{document}

% Title and authors
\runninghead{Strub and Gammell}
\title{AIT* and EIT*: Asymmetric bidirectional sampling-based path planning}
\author{Marlin P. Strub\affilnum{1} and Jonathan D. Gammell\affilnum{1}\vspace*{-1.1em}}
\corrauth{Marlin P. Strub, University of Oxford, Oxford OX2 6NN, United Kingdom}

% Affiliation
\affiliation{\affilnum{1}University of Oxford, Oxford, United Kingdom}%
\email{mstrub@robots.ox.ac.uk}

% Keywords
\keywords{sampling-based path planning, optimal path planning, heuristics,
  problem-specific heuristics, informed search, informed bidirectional search\vspace*{-0.5em}}

% Abstract (This has to be before the title.)
\begin{abstract}
  Optimal path planning is the problem of finding a valid sequence of states
  between a start and goal that optimizes an objective. Informed path planning
  algorithms order their search with problem-specific knowledge expressed as
  heuristics and can be orders of magnitude more efficient than uninformed
  algorithms. Heuristics are most effective when they are both accurate and
  computationally inexpensive to evaluate, but these are often conflicting
  characteristics. This makes the selection of appropriate heuristics difficult
  for many problems.

  This paper presents two almost-surely asymptotically optimal sampling-based
  path planning algorithms to address this challenge, \ac{AIT*} and
  \ac{EIT*}. These algorithms use an asymmetric bidirectional search in which
  both searches continuously inform each other. This allows \ac{AIT*} and
  \ac{EIT*} to improve planning performance by simultaneously calculating and
  exploiting increasingly accurate, problem-specific heuristics.

  The benefits of \ac{AIT*} and \ac{EIT*} relative to other sampling-based
  algorithms are demonstrated on twelve problems in abstract, robotic, and
  biomedical domains optimizing path length and obstacle clearance. The
  experiments show that \ac{AIT*} and \ac{EIT*} outperform other algorithms on
  problems optimizing obstacle clearance, where \textit{a priori} cost
  heuristics are often ineffective, and still perform well on problems
  minimizing path length, where such heuristics are often effective.
\end{abstract}

%%% Local Variables:
%%% mode: latex
%%% TeX-master: "../main"
%%% End:
\acresetall%
% RRT* and PRM* should never be expanded as acronyms.
\acused{RRT*}
\acused{PRM*}

% Make the title
\maketitle

% Introduction
\section{Introduction}%
\label{sec:introduction}

\begin{figure*}[t]
  % \begin{subfigure}[b][0.18\textwidth][c]{1em}%
  %   \begin{sideways}
  %     \centering%
  %     Path Length
  %   \end{sideways}
  % \end{subfigure}%
  \begin{center}
    Path length
  \end{center}
  \vspace{-1.0em}
  \begin{subfigure}[b]{0.195\textwidth}%
    \input{figures/1-introduction/evaluated_edges/path_length/rrtstar/tikzpictures/000154}%
    \caption{RRT*}%
    \label{fig:example-rrtstar-path-length}%
  \end{subfigure}%
  \hfill%
  \begin{subfigure}[b]{0.195\textwidth}%
    \input{figures/1-introduction/evaluated_edges/path_length/fmtstar/tikzpictures/000000}%
    \caption{FMT*}%
    \label{fig:example-fmtstar-path-length}%
  \end{subfigure}%
  \hfill%
  \begin{subfigure}[b]{0.195\textwidth}%
    \input{figures/1-introduction/evaluated_edges/path_length/bitstar/tikzpictures/000283}%
    \caption{BIT*}%
    \label{fig:example-bitstar-path-length}%
  \end{subfigure}%
  \hfill%
  \begin{subfigure}[b]{0.195\textwidth}%
    \input{figures/1-introduction/evaluated_edges/path_length/aitstar/tikzpictures/000766}%
    \caption{AIT*}%
    \label{fig:example-aitstar-path-length}%
  \end{subfigure}%
  \hfill%
  \begin{subfigure}[b]{0.195\textwidth}%
    \input{figures/1-introduction/evaluated_edges/path_length/eitstar/tikzpictures/011090}%
    \caption{EIT*}%
    \label{fig:example-eitstar-path-length}%
  \end{subfigure}
  \begin{center}
    \vspace{-0.6em} Obstacle clearance
  \end{center}
  \vspace{-0.6em}
  % \begin{subfigure}[b][0.23\textwidth][c]{1em}%
  %   \begin{sideways}
  %     \centering%
  %     \hspace{1em}Clearance
  %   \end{sideways}
  % \end{subfigure}%
  \begin{subfigure}[b]{0.195\textwidth}%
    \input{figures/1-introduction/evaluated_edges/reciprocal_clearance/rrtstar/tikzpictures/000168}%
    \caption{RRT*}%
    \label{fig:example-rrtstar-obstacle-clearance}%
  \end{subfigure}%
  \hfill%
  \begin{subfigure}[b]{0.195\textwidth}%
    \input{figures/1-introduction/evaluated_edges/reciprocal_clearance/fmtstar/tikzpictures/000000}%
    \caption{FMT*}%
    \label{fig:example-fmtstar-obstacle-clearance}%
  \end{subfigure}%
  \hfill%
  \begin{subfigure}[b]{0.195\textwidth}%
    \input{figures/1-introduction/evaluated_edges/reciprocal_clearance/bitstar/tikzpictures/002225}%
    \caption{BIT*}%
    \label{fig:example-bitstar-obstacle-clearance}%
  \end{subfigure}%
  \hfill%
  \begin{subfigure}[b]{0.195\textwidth}%
    \input{figures/1-introduction/evaluated_edges/reciprocal_clearance/aitstar/tikzpictures/002903}%
    \caption{AIT*}%
    \label{fig:example-aitstar-obstacle-clearance}%
  \end{subfigure}%
  \hfill%
  \begin{subfigure}[b]{0.195\textwidth}%
    \input{figures/1-introduction/evaluated_edges/reciprocal_clearance/eitstar/tikzpictures/011244}%
    \caption{EIT*}%
    \label{fig:example-eitstar-obstacle-clearance}%
  \end{subfigure}%
  \vspace{-0.3em}
  \caption{An illustration of the search trees constructed by \acs{RRT*},
    \acs{FMT*}, \acs{BIT*}, \acs{AIT*}, and \acs{EIT*} to find an initial
    solution when optimizing path length
    (\subref{fig:example-rrtstar-path-length}--\subref{fig:example-eitstar-path-length})
    and obstacle clearance
    (\subref{fig:example-rrtstar-obstacle-clearance}--\subref{fig:example-eitstar-obstacle-clearance}). The
    start and goal are represented by a black dot~(\tikzinlinestart) and
    circle~(\tikzinlinegoal), respectively. Sampled states are represented by
    small black dots~(\tikzinlinestate). State space obstacles are indicated
    with gray rectangles~(\tikzinlineobstacle). The initial solutions are shown
    in yellow~(\tikzinlinesolution) and the search trees constructed to find
    them are shown in black~(\tikzinlineforwardedge). Any edge in these search
    trees that is not part of the initial solution delayed finding
    it. \acs{RRT*} randomly explores the state space and fully evaluates many
    edges that are not part of the initial solution for both
    objectives~(\subref{fig:example-rrtstar-path-length},
    \subref{fig:example-rrtstar-obstacle-clearance}). \acs{FMT*} orders its
    search with increasing cost-to-come of the vertices and not only fully
    evaluates many edges that are not part of the initial solution for both
    objectives but also fails to find a solution when optimizing obstacle
    clearance because its connection strategy depends on edge
    cost~(\subref{fig:example-fmtstar-path-length},
    \subref{fig:example-fmtstar-obstacle-clearance}). \acs{BIT*} orders its
    search on the total potential solution cost according to an admissible cost
    heuristic and evaluates fewer edges that are not part of the initial
    solution~(\subref{fig:example-bitstar-path-length}) but only if an
    informative admissible cost heuristic is
    available~(\subref{fig:example-bitstar-obstacle-clearance}). \acs{AIT*}
    calculates and exploits a problem-specific heuristic and evaluates still
    fewer edges that are not part of the initial
    solution~(\subref{fig:example-aitstar-path-length}) but only when an
    informative admissible cost heuristic can be calculated for the optimization
    objective~(\subref{fig:example-aitstar-obstacle-clearance}). \acs{EIT*}
    calculates and exploits cost and effort heuristics and evaluates few edges
    even when an admissible cost heuristic cannot be calculated for the
    objective~(\subref{fig:example-eitstar-path-length},
    \subref{fig:example-eitstar-obstacle-clearance}).}%
  \label{fig:evaluated-edges}
  \vspace*{-0.5em}
\end{figure*}

%%% Local Variables:
%%% mode: latex
%%% TeX-master: "../../../main"
%%% End:

Path planning algorithms aim to find a sequence of valid states, called a path,
that connects a start to a goal. Sampling-based planners, such as
\accite{PRM}{kavraki_tro1996}, find paths by randomly sampling valid states and
connecting nearby states when these local connections are valid. The resulting
structure can be viewed as a graph embedded in a state space, where each vertex
represents a valid state and each edge a sequence of valid states connecting
two vertices. Multiple planning problems can be solved by adding starts and
goals to this embedded graph and then finding a path between them with a
graph-search algorithm.

A single planning problem is often solved more efficiently with incremental sampling-based planners, such as \accite{RRT}{lavalle_icra1999, lavalle_ijrr2001a} and its asymptotically optimal variant, \acs{RRT*}~\citep{karaman_rss2010,karaman_ijrr2011}. These planners build a search tree of valid paths rooted at the start by incrementally sampling and connecting states when these local connections are valid. This avoids having to specify the sampling resolution \textit{a priori}, but results in a randomly ordered search that spends computational effort on paths that are never part of a solution.

Best-first graph-search algorithms, such as Dijkstra's
algorithm~\citep{dijkstra_nm1959}, can search problems more efficiently by
ordering their search on \emph{partial} solution cost. Informed graph-search
algorithms, such as A*~\citep{hart_tssc1968}, can further increase search
efficiency by leveraging problem-specific information to order their search on
\emph{total} potential solution cost. This information is often expressed as a
heuristic function that estimates the cost to connect any two vertices in a
graph. A heuristic is called \textit{admissible} if it never overestimates the
true cost and \textit{consistent} if it satisfies a specific triangle
inequality. Given an admissible cost heuristic, A* finds an optimal solution,
and given a consistent cost heuristic, A* does so optimally efficiently with
respect to the number of expanded vertices~\citep{hart_tssc1968}.
% To guarantee these properties, A* processes
% states in order of total potential solution cost, coupling search order to
% solution quality guarantee.

The efficient search order of A* is combined with the incremental sampling of
\ac{RRT*} in informed sampling-based planners, such as
\accite{BIT*}{gammell_icra2015,gammell_ijrr2020}. This improves planning
performance, but only when effective cost heuristics are available. A heuristic
is most effective when it is both accurate and computationally inexpensive to
evaluate relative to other search operations. Such heuristics may not exist for
some problems, because they are inaccurate for a given obstacle configuration
or computationally expensive due to complex optimization objectives, or may
not be admissible, which is often required for theoretical performance
guarantees of informed planners.

Problem-specific information not expressible as admissible cost heuristics can
be exploited by more advanced informed graph-search algorithms, such as
\accite{AEES}{thayer_socs2012} and \accite{A-MHA*}{natarajan_socs2019}. These algorithms
decouple search order from solution quality guarantees, which allows them to
balance search efficiency with anytime performance. This is especially
important for robotic systems that operate under hard time constraints.
% \ac{AEES} uses
% multiple heuristics estimating solution cost and search effort to find an
% initial solution quickly and then continuously improves its solution with
% minimal computational effort.

% Computationally expensive search operations in sampling-based planning include
% state expansions and edge evaluations~\citep{hauser_icra2015,kleinbort_afr2020}. State
% expansions often require nearest-neighbor searches and edge evaluations require
% collision detection and edge-cost computation.

This paper presents techniques to inexpensively calculate accurate, admissible,
and problem-specific heuristics and exploit them with sampling-based planning
algorithms. This is achieved with an asymmetric bidirectional search that
considers different information in the forward and reverse searches. These two
searches continuously inform each other by sharing complementary information in
both directions. Algorithm~\ref{alg:conceptual} provides a conceptual overview
of this approach.

The reverse search calculates heuristics for the current sampling-based approximation of a planning problem. It exploits problem-specific information implicit in the observed distribution of valid states by combining \textit{a priori} heuristics between multiple states into more accurate heuristics between each state and the goal. The reverse search is computationally inexpensive because it only combines edge heuristics and avoids full collision detection and true edge cost evaluation.

% Effectiveness vs. Efficiency: Being effective is doing the right things,
% being efficient is doing the things right.

The forward search finds valid paths in the current sampling-based approximation of a planning problem. It does this effectively by exploiting the accurate, problem-specific heuristics calculated by the reverse search. The forward search informs the reverse search when invalid edges were used to calculate the heuristic, causing the reverse search to update the heuristic. The forward search is computationally expensive because it performs full collision detection and edge cost evaluation, but focused on connections likely to yield a solution by the calculated heuristics (Figure~\ref{fig:evaluated-edges}).

This paper presents two almost-surely asymptotically optimal sampling-based
planning algorithms informed by an asymmetric bidirectional search, \ac{AIT*}
and \ac{EIT*}. \ac{AIT*} calculates an increasingly accurate, admissible cost
heuristic with its reverse search and exploits this heuristic with its forward
search. This results in fast initial solution times even when the admissible
cost heuristic available \textit{a priori} is not accurate. The full details of
\ac{AIT*} are presented in \reft[Section]{sec:adaptively-informed-trees}.

\ac{EIT*} builds on \ac{AIT*} by calculating an additional cost and effort
heuristic with its reverse search and exploiting all three heuristics with its
forward search. This results in fast initial solution times even when no
admissible cost heuristic is available. The full details of \ac{EIT*} are
presented in \reft[Section]{sec:effort-informed-trees}.

The benefits of simultaneously calculating and exploiting adaptive heuristics
are demonstrated on twelve problems in abstract, robotic, and biomedical
domains in \reft[Section]{sec:experimental-results}. All domains are tested with
the path-length objective, where informative admissible heuristics are
available \textit{a priori}, and the obstacle-clearance objective, where
informative admissible heuristics are not always available \textit{a
  priori}. The results show that \ac{EIT*} outperforms all other asymptotically
optimal planners on all problems in all domains when optimizing obstacle
clearance. \ac{AIT*} and \ac{EIT*} also perform well when minimizing path
length in comparison to the tested planners when considering success rates,
median initial solution times, and median solution quality over time.

\subsection{Statement of Contributions}%
\label{sec:statement-of-contributions}

This paper expands on ideas first published as \citet{strub_icra2020b}. It makes the
following specific contributions:
\begin{itemize}[leftmargin=\parindent, itemsep=0pt]\setlist{nosep}
\item Presents \ac{EIT*} as an extension of \ac{AIT*} to optimization objectives that are computationally expensive or difficult to approximate with an admissible \textit{a priori} cost heuristic~\refp[Section]{sec:effort-informed-trees}.
\item Proves the almost-sure asymptotic optimality of these algorithms by building on results from the path-planning and graph-search literature~\refp[Section]{sec:analysis}.
\item Demonstrates the effectiveness of these algorithms accross multiple domains and optimization objectives, including problems of robotic manipulation and problems with continuous goal regions~\refp[Section]{sec:experimental-results}.
\end{itemize}

%%% Local Variables:
%%% mode: latex
%%% TeX-master: "../main"
%%% End:

% Background
\section{Background}%
\label{sec:background}

This section first formally defines the optimal path planning problem
\refp[Section]{sec:problem-definition} and then reviews related techniques from
the literature on using heuristics in sampling-based planners and graph-search
algorithms \refp[Section]{sec:literature-review}.

\subsection{Problem Definition}%
\label{sec:problem-definition}

There are two widely studied versions of the path planning problem. The
\textit{feasible} path planning problem is the task of finding a sequence of
valid states, i.e., a path, that leads from a start to a goal. Many feasible
problems have many solutions. The \textit{optimal} path planning problem is the
task of finding the best among these solutions, i.e., a valid path that is
optimal with respect to a given optimization objective. Many optimal problems
have a unique solution. The optimal path planning problem is formally defined
in Definition~\ref{def:optimal-planning}.

\begin{definition}[The Optimal Path Planning Problem~\citep{karaman_ijrr2011}]%
  \label{def:optimal-planning}
  Let the state space of a planning problem be denoted by \( X \), the subset
  of invalid states by \( X_{\mathrm{invalid}} \subset X \), and the subset of
  valid states by
  \( X_{\mathrm{valid}} \coloneqq \mathop{\mathrm{closure}} \left( X \setminus
    X_{\mathrm{invalid}} \right) \). Let the start state and the set of goal
  states be denoted by
  \( \bm{\mathrm{x}}_{\mathrm{start}} \in X_{\mathrm{valid}} \) and
  \( X_{\mathrm{goal}} \subset X_{\mathrm{valid}} \), respectively. Let
  \( \sigma \colon [0, 1] \to X_{\mathrm{valid}} \) be a continuous function
  with bounded total variation, i.e., a valid path, and let the set of all
  valid paths be denoted by \( \Sigma \). Let the optimization objective be
  defined by a cost function, \( c \colon \Sigma \to [0, \infty) \), that maps
  each path to a nonnegative real number.

  The optimal path planning problem is the task of finding a path,
  \( \sigma^{*} \in \Sigma \), from the start to the goal with minimum cost,
  \begin{equation*}
    \sigma^{*} \coloneqq \mathop{\arg\min}_{\sigma \in \Sigma} \set*{ c(\sigma) \given \sigma(0) = \bm{\mathrm{x}}_{\mathrm{start}}, \sigma(1) \in X_{\mathrm{goal}}},
  \end{equation*}
  or reporting failure if no such path exists.
\end{definition}

\begin{algorithm}[t]
  \caption{Asymmetric bidirectional search}%
  \label{alg:conceptual}
  \SetInd{0.0em}{0.4em}
  \SetVlineSkip{0.0em}
  \tt\footnotesize
  \Repeat{\normalfont\tt stopped}{
    improve \acs{RGG} approximation {\color{black!50}(sampling)}\;
    calculate heuristics {\color{black!50}(reverse search)}\;
    \While{\normalfont\tt \acs{RGG} approximation is useful}{
      find solution {\color{black!50}(forward search)}\;
      \If{\normalfont\tt found invalid edge or unprocessed state}{
        update heuristics {\color{black!50}(reverse search)}\;
      }
    }
  }
\end{algorithm}%

%%% Local Variables:
%%% mode: latex
%%% TeX-master: "../../../main"
%%% End:

Sampling-based planners are often evaluated probabilistically as a function of
the number of samples over all possible realizations of a
distribution. Algorithms whose probability of solving the feasible path
planning problem approaches one as the number of samples approaches infinity
are called \emph{probabilistically complete}. Algorithms that asymptotically
solve the optimal planning problem as the number of samples approaches infinity
with a probability of one are called \emph{almost-surely asymptotically
  optimal}~\citep{karaman_ijrr2011}. Almost-sure asymptotic optimality implies
probabilistic completeness and is formally defined in
Definition~\ref{def:asymptotic-optimality}.

\begin{definition}[Almost-sure asymptotic optimality~\citep{karaman_ijrr2011}]%
  \label{def:asymptotic-optimality}
  A sampling-based path planning algorithm is called \emph{almost-surely
    asymptotically optimal} if it has a unity probability of asymptotically
  solving the optimal path planning problem as the number of samples approaches
  infinity (if an optimal solution exists),
  \begin{equation*}
    P\left( \adjustlimits{\mathop{\lim\,\sup}}_{{q \to \infty}}{\;\min}_{\;\sigma \in
        \Sigma_{q}} \{ c(\sigma) \} = c^{*} \right) = 1,
  \end{equation*}
  where \( q \) is the number of samples, \( \Sigma_{q} \subset \Sigma \) is
  the set of valid paths from the start to the goal found by the planner from
  those samples, \( c \colon \Sigma \to [0, \infty) \) is the cost function,
  and \( c^{*} \) is the optimal solution cost.
\end{definition}

Sampling-based planners can also use deterministic sampling
strategies~\citep[e.g.,][]{branicky_icra2001, lavalle_ijrr2004}, which can result in deterministic optimality guarantees~\citep{janson_ijrr2018}.
The finite-time properties of asymptotically optimal planners are analyzed by
\citet{dobson_iros2013}, \citet{janson_ijrr2018}, and \citet{tsao_icra2020}.

Formally analyzing sampling-based planners requires assumptions about
the path planning problem~\citep[e.g.,][]{gammell_arcras2021}. The analysis of
\ac{AIT*} and \ac{EIT*} builds on the probabilistic results of \citet{karaman_ijrr2011} and makes the same assumptions (Sections~\ref{sec:state-space-assumptions}--\ref{sec:solution-assumptions}).

\subsubsection{State Space Assumption}%
\label{sec:state-space-assumptions}

The state space of the planning problem is assumed to be an open, \( n
\)-dimensional unit (hyper)cube, \( X \coloneqq (0, 1)^{n} \), but problems
with other state spaces can also be searched~\citep{kleinbort_wafr2016, kleinbort_afr2020}.

\subsubsection{Cost Function Assumptions}%
\label{sec:cost-function-assumptions}

Let \( \sigma_{1}, \sigma_{2} \in \Sigma \) be two paths such that
\( \sigma_{1}(1) = \sigma_{2}(0) \), and let
\( (\sigma_{1} | \sigma_{2}) \in \Sigma \) denote their concatenation,
\begin{equation*}
  (\sigma_{1} | \sigma_{2})(t) \coloneqq
  \begin{cases}
    \sigma_{1}(2t) & \text{for } t \in \lbrack 0, \nicefrac{1}{2}
    \rbrack \\
    \sigma_{2}(2t - 1) & \text{for } t \in \lparen\nicefrac{1}{2}, 1\rbrack.
  \end{cases}
\end{equation*}
The cost of any path, \( \sigma = ( \sigma_{1} | \sigma_{2} ) \in \Sigma \), is
assumed to be lower bounded by the cost of any of its segments,
\begin{equation*}
  \forall\; \sigma_{1}, \sigma_{2} \text{ s.t. } \sigma = (
  \sigma_{1} | \sigma_{2} ), \quad c(\sigma) \geq \max\{
c(\sigma_{1}), c(\sigma_{2}) \},
\end{equation*}
and upper bounded by a multiple of its total variation,
\begin{equation*}
  \exists k \in [0, \infty), \quad c(\sigma) \leq k \mathop{\mathrm{TV}}(\sigma),
\end{equation*}
where \( \mathrm{TV}(\sigma) \) denotes the total variation of the
path~\( \sigma \)~\citep{karaman_ijrr2011}.

It is also assumed that only trivial paths consisting of a single state can
have zero cost,
\begin{equation*}
  c(\sigma) = 0 \iff \forall\; t \in [0, 1], \sigma(t) = \sigma(0).
\end{equation*}

\subsubsection{Obstacle Assumption}%
\label{sec:obstacle-assumptions}

The obstacle configuration of the optimal path planning problem is assumed to
allow for a valid path from the start to the goal that remains a fixed
distance, \( \delta > 0 \), from its nearest obstacles for its entire length,
\begin{equation*}
  \exists\; \sigma \in \Sigma, \delta \in (0, \infty), \; \text{s.t. } \forall\; t \in [0,
  1], B_{\delta, n}(\sigma(t)) \subset X_{\mathrm{valid}},
\end{equation*}
where \( B_{\delta, n}(\sigma(t)) \) is an \( n \)-dimensional ball with radius
\( \delta \) centered at \( \sigma(t) \),
\begin{equation*}
  B_{\delta, n}(\bm{\mathrm{x}}) \coloneqq \left\{ \bm{\mathrm{x}}^{\prime} \in X
    \;\big|\; {\| \bm{\mathrm{x}} - \bm{\mathrm{x}}^{\prime} \|}_{2} \leq \delta \right\}.
\end{equation*}
Such a path is said to have \emph{strong \( \delta \)-clearance}.

\subsubsection{Optimal Solution Assumption}
\label{sec:solution-assumptions}

At least one solution of the optimal path planning problem,
\( \sigma^{*} \in \Sigma \), is assumed to be homotopic to a path,
\( \sigma_{\delta} \in \Sigma \), with strong \( \delta \)-clearance,
\begin{equation*}
  \exists\; H \colon [0, 1] \to \Sigma, \quad H(0) = \sigma^{*}, H(1) = \sigma_{\delta},
\end{equation*}
where \( H \) is a homotopic map whose image is the set of all valid paths from
the start to the goal. Such a solution is said to have \emph{weak
  \( \delta \)-clearance}.

\subsection{Literature Review}%
\label{sec:literature-review}

Almost-surely asymptotically optimal planning is a popular area of
research~\citep{gammell_arcras2021}. This section focuses on sampling-based and
graph-search techniques to calculate and/or exploit heuristics to improve
performance. Sampling-based planners that use heuristics to guide the sampling
and/or order the search are reviewed in
\reft[Section]{sec:sampling-based-planning-with-heuristics}. Approaches that
calculate and exploit accurate cost heuristics for graph-search algorithms are
reviewed in \reft[Section]{sec:improved-heuristics-for-informed-search} and
algorithms that use effort heuristics in
\reft[Section]{sec:effort-and-distance-based-heuristics-in-informed-search}.
Using heuristics in sampling-based planning has parallels with lazy collision
detection, which is reviewed in
\reft[Section]{sec:sampling-based-planning-with-lazy-collision-detection}.

\subsubsection{Heuristics in Sampling-Based Planning}%
\label{sec:sampling-based-planning-with-heuristics}

Sampling-based planning algorithms can improve their performance by using
heuristics to bias their sampling and guide their search.

\ac{RRT}-Connect~\citep{kuffner_icra2000} builds on \ac{RRT} by growing two
trees, one rooted in the start and one in the goal state. These trees each
explore the state space around them, but are also guided towards each other
with a \textit{connect heuristic}. This approach can result in very fast
initial solution times, but does not consider the solution cost and can
consequently not improve the solution given more computational
time. Almost-surely asymptotically optimal variants of \ac{RRT}-Connect
exist~\citep{akgun_iros2011,jordan_tr2013,klemm_robio2015,qureshi_ras2015,burget_iros2016} but the
connect heuristic does not guide the search beyond finding an initial solution.

\accite{hRRT}{urmson_iros2003} and \accite{GBRRT}{nayak_arxiv2021} bias the
growth of their trees with cost heuristics. \ac{hRRT} uses \textit{a priori}
heuristics to weigh the Voronoi regions of \ac{RRT}. \ac{GBRRT} is a
bidirectional version of \ac{RRT} that guides the forward tree with heuristics
computed by the reverse tree. These algorithms have improved performance but do
not provide any bounds on the quality of their solution.

Informed \acs{RRT*}~\citep{gammell_iros2014, gammell_tro2018} builds on \acs{RRT*} by using an admissible cost heuristic to ensure that only states that can improve the current solution are processed. This improves \acs{RRT*}'s convergence rate and retains its almost-sure asymptotic optimality, but does not guide the search with the heuristic, does not improve the accuracy of the heuristic as the search progresses, and does not provide any benefits until an initial solution is found. \citet{kunz_icra2016} and \citet{yi_icra2018} extend informed sampling to kinodynamic systems. \citet{joshi_icra2019} present a variant of Informed \acs{RRT*} that uses previous collision detection results and available information in the graph structure to guide the search.

\acs{RRTsharp}~\citep{arslan_icra2013, arslan_icra2015} builds on \acs{RRT*} by
ensuring that all samples are optimally connected to the search tree after each
iteration. It does this efficiently by using an admissible cost heuristic to
update the connections of suboptimally connected samples in order of their
total potential solution cost. This again improves \acs{RRT*}'s convergence
rate and retains its almost-sure asymptotic optimality, but can also not
improve the accuracy of the heuristic as the search progresses and does not
provide any benefits until an initial solution is found.

\citet{sakcak_lcss2020} present a method that incorporates a heuristic into a
version of RRT* that is based on motion primitives~\citep{sakcak_ar2019}. This can
improve the performance on kinodynamic problems but uses a discretization of
the state space that suffers from the \textit{curse of
  dimensionality}~\citep{bellman_book1957}.

\accite{IST}{bekris_stair2008}, A*-RRT~\citep{li_icra2011}, the \( f \)-biasing
method~\citep{kiesel_socs2012}, P-PRM~\citep{le_iros2014}, and
\accite{RIOT}{westbrook_iros2020} all search a simplified approximation of the state
space to calculate an accurate cost heuristic, which is then used to guide a
sampling-based planner. These approaches improve planning performance but
require a preprocessing step.

% A similar approach is the \( f \)-biasing method by \citet{kiesel_socs2012}, in
% which a simplified abstraction of the problem is searched with Dijkstra's
% algorithm~\citep{dijkstra1959} to create a cost heuristic to bias the sampling
% of \ac{RRT} and \ac{RRT*}. This results in better convergence rates but relies
% on a preprocessing step.

% P-PRM~\citep{le_iros2014} runs \ac{PRM} on a simplified configuration space of the
% motion planning problem, and processes the resulting graph with Dijkstra's
% algorithm to generate a cost heuristic in the simplified
% space. It then uses this heuristic to guide its search in the original
% configuration space. This approach can speed up planning times but requyires
% many user-specified parameters and does not guarantee any bounds on the quality
% of its solutions. A similar approach is taken by \accite{RIOT}{westbrook_iros2020},
% which is an almost-surely asymptotically optimal algorithm specialized for
% kinodynamic systems.

% A*\=/\ac{RRT}~\citep{li_icra2011} aims to evenly explore the state space of
% kinodynamic systems. It creates a graph in an obstacle-free version of the
% state space of a kinodynamic system as a preprocessing step. This graph is then
% processed with A* to create a cost heuristic online that captures the
% kinodynamic constraints of the system. This cost heuristic is in turn used as
% the distance metric for \ac{RRT}. A*\=/RRT samples states more uniformly than
% RRT in the state space of kinodynamic systems, but requires a preprocessing
% step and significantly increases the runtime of \ac{RRT}.

\accite{BEAST}{kiesel_iros2017} is similar to these methods in that it runs \ac{PRM}
on a simplified abstraction of the problem and uses the resulting graph to
calculate an effort heuristic for the samples in the simplified space. If
searching the original space reveals that regions in the abstract space cannot
easily be connected, then this effort heuristic is updated in a Bayesian
manner. BEAST tends to find initial solutions faster than other planners but
does not provide any guarantees on the quality of its solutions.

\accite{QMP}{orthey_iros2018},~\accite{QRRT}{orthey_isrr2019}, and their asymptotically optimal variants \ac{QMP}* and \ac{QRRT}*~\citep{orthey_arxiv2020b} solve planning problems with sequences of admissible lower-dimensional simplifications of increasing dimensionality. Paths in lower-dimensional simplifications can guide the sampling of states in higher dimensional simplifications and can be seen as admissible heuristics. This can improve performance by orders of magnitude, especially for high-dimensional problems, but requires the user to manually specify the sequence of lower-dimensional simplifications for each problem.

\accite{MPLB}{salzman_icra2015b} is an anytime adaption of
\ac{FMT*}~\citep{janson_isrr2013, janson_ijrr2015} that incorporates admissible cost
heuristics. \ac{MPLB} uses two passes of Dijkstra's algorithm to restrict the
set of samples to be searched and another pass of Dijkstra's to calculate an
admissible cost heuristic for these samples, all without detecting
collisions. It then uses the resulting cost heuristic in a forward search with
collision detection to find a path. This approach can result in accurate,
admissible cost heuristics and requires few collision detections but does not
update the heuristic when the forward search detects collisions on edges that
were used to compute the heuristic.

\ac{BIT*} samples batches of states and views these states as an increasingly
dense edge-implicit \accite{RGG}{penrose_book2003}. It uses an admissible cost
heuristic to search this graph in order of potential solution quality with
techniques similar to \accite{LPA*}{koenig_ai2004, likhachev_icaps2005b,
  aine_ai2016}. \accite{ABIT*}{strub_icra2020a} speeds up initial solution times
by inflating its heuristic, similar to \accite{ARA*}{likhachev_nips2004}, and
balances exploring the state space with exploiting the current \ac{RGG}
approximation by truncating its search, similar to
\accite{TLPA*}{aine_ai2016}. Both algorithms work best when the cost of a path
correlates well with the computational effort required to validate it and when
accurate cost heuristics are available \textit{a priori}, but require that
these heuristics are admissible and do not improve their accuracy as the search
progresses.

Similar to \ac{AIT*} and \ac{EIT*}, the methods in this section use heuristics
to improve the performance of sampling-based planning algorithms. In contrast
to \ac{AIT*} and \ac{EIT*}, these methods either do not apply heuristics to all
aspects of the search, do not provide any bounds on the quality of their
solution, require a preprocessing step, do not calculate problem-specific
heuristics, or do not improve the accuracy of the heuristic as the search
progresses.

\begin{figure*}
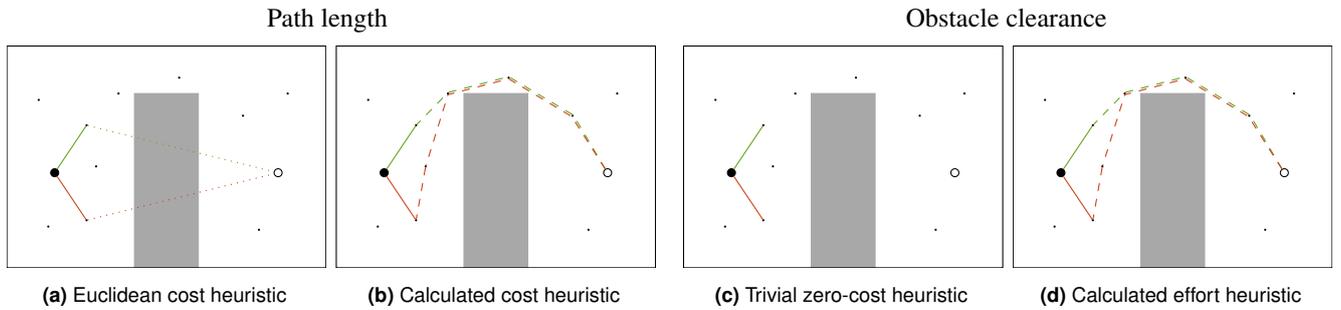

  \hspace{9.5em} Path length \hspace{18.9em} Obstacle clearance\\[0.4em]
  \begin{subfigure}[t]{0.49\columnwidth}
    \begin{tikzpicture}[scale = 4.2]
  \input{figures/3-algorithms/3-0-benefits-of-heuristics/context.tikz}
  \input{figures/3-algorithms/3-0-benefits-of-heuristics/states.tikz}

  \draw [forward edge, espgreen] (start) -- (v1);
  \draw [forward edge, espred] (start) -- (v2);

  \draw [forward edge, espgreen, dotted] (v1) -- (goal);
  \draw [forward edge, espred, dotted] (v2) -- (goal);
\end{tikzpicture}

%%% Local Variables:
%%% mode: latex
%%% TeX-master: "../../../main"
%%% End:
%
    \caption{Euclidean cost heuristic}%
    \label{fig:euclidean-heuristic}
  \end{subfigure}%
  \hspace{0.5em}%
  \begin{subfigure}[t]{0.49\columnwidth}
    \begin{tikzpicture}[scale = 4.2]
  \input{figures/3-algorithms/3-0-benefits-of-heuristics/context.tikz}
  \input{figures/3-algorithms/3-0-benefits-of-heuristics/states.tikz}

  \draw [forward edge, espgreen] (start) -- (v1);
  \draw [forward edge, espred] (start) -- (v2);

  \draw [forward edge, espgreen, dashed] (goal) -- (v6.north east) -- (v5.north) -- (v4.north) -- (v1);
  \draw [forward edge, espred, dashed] (goal) -- (v6.south west) -- (v5.south) -- (v4.south) -- (v3) -- (v2);
\end{tikzpicture}

%%% Local Variables:
%%% mode: latex
%%% TeX-master: "../../../main"
%%% End:
%
    \caption{Calculated cost heuristic}%
    \label{fig:adaptive-cost-heuristic}
  \end{subfigure}%
  \hfill%
  \begin{subfigure}[t]{0.49\columnwidth}
    \begin{tikzpicture}[scale = 4.2]
  \input{figures/3-algorithms/3-0-benefits-of-heuristics/context.tikz}
  \input{figures/3-algorithms/3-0-benefits-of-heuristics/states.tikz}

  \draw [forward edge, espgreen] (start) -- (v1);
  \draw [forward edge, espred] (start) -- (v2);
\end{tikzpicture}

%%% Local Variables:
%%% mode: latex
%%% TeX-master: "../../../../main"
%%% End:
%
    \caption{Trivial zero-cost heuristic}%
    \label{fig:zero-heuristic}
  \end{subfigure}%
  \hspace{0.5em}%
  \begin{subfigure}[t]{0.49\columnwidth}
    \begin{tikzpicture}[scale = 4.2]
  \input{figures/3-algorithms/3-0-benefits-of-heuristics/context.tikz}
  \input{figures/3-algorithms/3-0-benefits-of-heuristics/states.tikz}

  \draw [forward edge, espgreen] (start) -- (v1);
  \draw [forward edge, espred] (start) -- (v2);

  \draw [forward edge, espgreen, dashed] (goal) -- (v6.north east) -- (v5.north) -- (v4.north) -- (v1);
  \draw [forward edge, espred, dashed] (goal) -- (v6.south west) -- (v5.south) -- (v4.south) -- (v3) -- (v2);
\end{tikzpicture}

%%% Local Variables:
%%% mode: latex
%%% TeX-master: "../../../main"
%%% End:
%
    \caption{Calculated effort heuristic}%
    \label{fig:adaptive-effort-heuristic}
  \end{subfigure}%
  \caption{An illustration of how \ac{AIT*} and \ac{EIT*} leverage the observed
    distribution of valid states to calculate accurate, problem-specific
    heuristics. The start and goal are represented by a black
    dot~(\tikzinlinestart) and circle~(\tikzinlinegoal), respectively. Sampled
    states are represented by small black dots~(\tikzinlinestate). The state
    space obstacle is indicated with a gray
    rectangle~(\tikzinlineobstacle). Euclidean distance is often used as a cost
    heuristic when optimizing path length~(\subref{fig:euclidean-heuristic},
    \subref{fig:adaptive-cost-heuristic}). It suggests the green and red edges
    are equally promising, even through the red edge leads to a dead
    end~(\subref{fig:euclidean-heuristic}). Calculating a problem-specific cost
    heuristic with a reverse search reveals that the green edge is more
    promising and can lead the forward search around the obstacle without
    evaluating many unnecessary edges~(\subref{fig:adaptive-cost-heuristic}).
    Some optimization objectives may not easily allow for informative
    admissible heuristics, such as obstacle
    clearance~(\subref{fig:zero-heuristic},
    \subref{fig:adaptive-effort-heuristic}). Most informed search algorithms
    are ordered by cost-to-come in the absence of an informative heuristic,
    which again suggests that the green and red edges are equally
    promising~(\subref{fig:zero-heuristic}). Calculating a problem-specific
    effort heuristic with a reverse search again reveals that the green edge
    can lead to a solution faster than the red edge, even in the absence of an
    informative admissible cost
    heuristic~(\subref{fig:adaptive-effort-heuristic}).}%
  \label{fig:benefits-of-adaptive-heuristics}
\end{figure*}

%%% Local Variables:
%%% mode: latex
%%% TeX-master: "../../../main"
%%% End:

\subsubsection{Improved Heuristics in Graph-Search}%
\label{sec:improved-heuristics-for-informed-search}

Developing and exploiting accurate heuristics is an important area of research
in informed graph-search algorithms. Techniques to calculate more accurate
heuristics have improved performance in various problem domains, e.g.,
the 15-Puzzle~\citep{culberson_cscsi1996}, Rubik's Cube~\citep{korf_ncai1997},
and robot vacuum on a grid~\citep{thayer_icaps2011b}.

Pattern databases~\citep{culberson_cscsi1996, korf_ncai1997, culberson_ci1998}
are precomputed tables of exact solution costs to potentially simplified
subproblems of a problem domain. The highest solution cost of any remaining
subproblem in an ongoing search can be used as an accurate heuristic for an
informed search. Additive pattern databases~\citep{felner_jair2004} are constructed
such that the heuristic remains admissible when the solution costs of all
remaining subproblems are combined, which can result in more accurate
heuristics. This increased accuracy can significantly improve performance, but
is confined to problem domains for which pattern databases can be generated.

\accite{HA*}{holte_ncai1996} uses homo\-morphic transformations of the state space to create abstractions in which multiple states of the original space are mapped to a single state in abstract space. These abstractions are then searched to calculate a heuristic for the original state space. This can result in fewer expanded states, but the presented technique is only shown to work for graphs with uniform edge costs. \accite{HCA*}{silver_aiide2005} and \accite{WHCA*}{silver_aiide2005} are multiagent versions of \ac{HA*} that use \accite{RRA*}{silver_aiide2005} to search the abstraction from the goal to the start. The cost of the optimal paths to states from the goal in the abstract space is used as the heuristic for the corresponding states in the original space. If the search in the original space processes a state whose abstract representation has not been processed by \ac{RRA*}, then \ac{RRA*} is resumed until it finds the optimal path to an abstract state that corresponds to the state being processed by the search in the original space. This results in lower cost paths and better success rates than alternative multi-agent search algorithms, but cannot directly be applied to single-agent planning in continuous spaces.

\accite{AA*}{koenig_aamas2005, sun_aamas2008} is an incremental search algorithm that
calculates an increasingly accurate, admissible cost heuristic for subsequent
searches of a graph with the same goal but different start states. After each
search, the heuristic cost-to-go value of each closed state is updated to be
the difference between the solution cost and the cost-to-come value of that
state. This results in increasingly efficient searches of any problem domain
but does not provide any benefits for the initial search of a graph.

The method in~\citet{thayer_icaps2011b} also generates more accurate heuristics for
any domain. It uses a relationship between the cost-to-go of a state and the
cost-to-go of its best child to define a \emph{single-step error} in the cost
heuristic. The mean single-step error in this heuristic is then calculated
either globally or per branch and used to adjust the heuristic
accordingly. This approach can be used in the initial search but is not
guaranteed to produce admissible heuristics.

The \textit{Add method}~\citep{kaindl_jair1997} uses a bidirectional search in
which a partial reverse search generates heuristics that inform the forward
search. The reverse search reveals errors in the heuristic values of the
processed states, the minimum of which is added to the heuristic values of all
unexpanded states in the forward search. This results in a more informed
heuristic that remains admissible, but increases the heuristic value for all
unexpanded states uniformly and requires a user-defined parameter that
specifies how many states to expand in the reverse search. A version without
this parameter is presented by \citet{wilt_cai2013}.

Similar to \ac{AIT*} and \ac{EIT*}, the methods in this section generate
increasingly accurate heuristics that result in increasingly efficient
searches. In contrast to \ac{AIT*} and \ac{EIT*}, these methods either require
preprocessing, cannot be used for the initial search of a graph, result in
inadmissible heuristics, or increase the heuristic value for all unexpanded
states uniformly and only order by estimated solution cost.

\subsubsection{Effort Heuristics in Graph-Search}%
\label{sec:effort-and-distance-based-heuristics-in-informed-search}

Ordering the search based on (inflated) cost heuristics improves anytime
performance the most when the cost of a path correlates well with the
computational effort required to find it~\citep{wilt_socs2012}. Directly
ordering the search on the com\-putational effort of a path can instead improve performance even when this is not the case. The graph-search literature often uses the number of states that must be expanded to find a solution as a proxy for the total search effort.

\accite{DWA*}{pohl_ijcai1973} aims to improve performance by ordering its
search in a manner that rewards progress away from the start. It multiplies an
admissible cost heuristic by a weighting factor that decreases with increasing
depth in the search tree. This is shown to reduce the number of expanded states
on some problem domains, but requires an \textit{a priori} estimate of the
solution depth and implicitly assumes that every step away from the start is a
step closer to the goal. Revised DWA*~\citep{thayer_icaps2009} removes this
assumption, but still requires an \textit{a priori} estimate of the solution
depth.

A\(_{\varepsilon}^{*}\)~\citep{pearl_pami1982} aims to expand states that are as
close to the goal as possible and could be part of a solution whose cost is
within a user-specified factor of the optimal cost. It always expands the node
with the least number of states left to be expanded, provided it could be part
of a solution within the suboptimality bound according to an admissible cost
heuristic. This works well if loose suboptimality bounds are acceptable or a
very accurate cost heuristic is available, but otherwise forces the search to
expand states with a large estimate of states left to be expanded just to
increase the lower bound on the optimal solution cost.

\accite{EES}{thayer_icaps2011a} aims to always expand the node which most quickly
leads to a solution whose cost is within a user-specified bound of the optimal
cost. It uses an admissible cost heuristic to guarantee the bound on the
suboptimality and inadmissible cost and effort heuristics to guide its
search. This significantly improves search performance in domains where
solution cost and depth can differ, but introduces computational overhead and
algorithmic complexity because \ac{EES} must maintain three queues ordered on
three different quantities. \ac{AEES} is an anytime version of EES and provides
the foundation of the forward search of \ac{EIT*}, which is discussed in
\reft[Section]{sec:eit-forward-search}.

Similar to \ac{AIT*} and \ac{EIT*}, the methods in this section use estimates
of the computational effort required to discover a solution to guide the search
and improve performance. In contrast to \ac{AIT*} and \ac{EIT*}, these methods
do not increase the accuracy of their heuristics as the search progresses.

\subsubsection{Lazy Collision Detection}%
\label{sec:sampling-based-planning-with-lazy-collision-detection}

A byproduct of using heuristics in sampling-based planning to bias the sampling
and guide the search is often that fewer edges have to be fully evaluated
\refp[Figure]{fig:evaluated-edges}. This relates informed path planning
algorithms to algorithms with lazy collision detection that explicitly aim to
minimize the number of collision detections.

Lazy \ac{PRM}~\citep{bohlin_icra2000} and Fuzzy
\ac{PRM}~\citep{nielsen_iros2000} take similar approaches to minimizing
computational effort through lazy collision detection. Both algorithms
initially connect samples without performing any collision detection on the
edges. The resulting graph is processed with an informed graph-search algorithm
to find a path that connects the start and goal states, and only checked for
collision once a path is found. If collisions are detected, then the
corresponding vertices and edges are removed from the graph and the updated
graph is processed again to find a new path between the start and goal
states. This results in few fully evaluated edges and improved planning
performance, but Lazy \ac{PRM} and Fuzzy \ac{PRM} do not provide any guarantee
on the quality of its solution and do not improve the accuracy of the heuristic
used in their graph-search algorithm. Almost-surely asymptotically optimal
variants of similar approaches exist~\citep{hauser_icra2015, kim_icra2018} but these
algorithms also do not improve the accuracy of their heuristics as the search
progresses.

The \ac{SBL} planner~\citep{sanchez_rr2003} combines lazy collision detection with
ideas from RRT-Connect. It grows two trees, similar to RRT-Connect, but only
checks collisions on edges that it believes to be on a path connecting the
start and goal states, like Lazy \ac{PRM} and Fuzzy \ac{PRM}. \ac{SBL} achieves
fast solution times, but does also not provide any guarantees on the quality of
its solution and does not improve its solution given more computational time.

\accite{LBT-RRT}{salzman_icra2014,salzman_tro2016} extends \acs{RRT*} with a graph whose
edges are not fully evaluated and uses this graph to determine which edges to
evaluate next. \ac{LBT-RRT} is almost-surely asymptotically near-optimal and
allows for continuous interpolation between \ac{RRT} and
\accite{RRG}{karaman_ijrr2011} by only rewiring states that are
\( \varepsilon \)-inconsistent. A similar approach is used for replanning in
dynamic environments in
\acs{RRTX}~\citep{otte_afr2015,otte_ijrr2016}. Interpolating \ac{LBT-RRT} between
\ac{RRT} and \ac{RRG} allows for balancing exploration with exploitation, but
\ac{LBT-RRT} can only optimize path length.

The \ac{LazySP} Framework~\citep{dellin_icaps2016} explicitly aims to minimize the number of edges that are checked for collision. It first finds a path from the start to the goal using a heuristic for the edge cost and then uses an \textit{edge selector} to determine the order in which the edges on the potential solution path are checked for collision. This often results in few fully evaluated edges, but is not asymptotically optimal and restarts the search every time an edge is found to be invalid.
\accite{GLS}{mandalika_icaps2019} builds on \ac{LazySP} by presenting a framework that can algorithmically balance edge evaluation with continuing the search, but is also not asymptotically optimal.
\accite{LRHA*}{mandalika_icaps2018} is an example algorithm that fits within the \ac{LazySP} and \ac{GLS} frameworks.

Similar to \ac{AIT*} and \ac{EIT*}, the methods in this section improve
performance by reducing the number of full edge evaluations. In contrast to
\ac{AIT*} and \ac{EIT*}, these methods either do not use heuristics, do not
improve the accuracy of their heuristics, do not guarantee any bounds on the
quality of their solution, do not improve their solution given more computation
time, or can only optimize path length.

%%% Local Variables:
%%% mode: latex
%%% TeX-master: "../main"
%%% End:

% Algorithms
\section{\acs{AIT*} and \acs{EIT*}}%
\label{sec:algorithms}

\begin{table*}
  \vspace*{-0.5em}
  \centering
  \footnotesize
  \begin{tabular}{lllll}
    \toprule
    \multicolumn{2}{l}{Component} & \ac{BIT*}                                   & \ac{AIT*}                                   & \ac{EIT*}\\
    \midrule
    \multicolumn{2}{l}{Approximation}
                                  & Series of \ac{RGG}s                         & Series of \ac{RGG}s                         & Series of \ac{RGG}s\\[0.8em]
    \multirow{4}{*}{\begin{sideways}\parbox{1cm}{\centering Reverse search}\end{sideways}}
            & Purpose             & ---                                         & Calculate admissible cost heuristic          & Calculate in-/admissible cost \& effort heuristics\\
            & Algorithm           & ---                                         & Vertex-queue \ac{LPA*}                      & Edge-queue A*\\
            & Ordering            & ---                                         & \textit{A priori} admissible solution cost  & \textit{A priori} admissible solution cost \& effort\\
            & Validation          & ---                                         & None                                        & Adaptive sparse collision detection\\[0.8em]
    \multirow{4}{*}{\begin{sideways}\parbox{1cm}{\centering Forward search}\end{sideways}}
            & Purpose             & Find valid paths                            & Find valid paths                            & Find valid paths\\
            & Algorithm           & Edge-queue \ac{TLPA*}                       & Edge-queue A*                               & Edge-queue \ac{AEES}\\
            & Ordering            & \textit{A priori} admissible solution cost  & Calculated admissible solution cost          & Calculated in-/admissible solution cost \& effort\\
            & Validation          & Dense collision detection                   & Dense collision detection                   & Dense collision detection\\
    \bottomrule
  \end{tabular}\vspace{0.5em}
  \caption{An overview of the components in \ac{BIT*}, \ac{AIT*}, and
    \ac{EIT*}. All three algorithms approximate the state space with the same
    series of increasingly dense \ac{RGG}s. \ac{BIT*} does not have a reverse
    search and finds valid paths with an edge-queue version of \ac{TLPA*}
    ordered by the total potential solution cost according to an \textit{a
      priori} admissible cost heuristic. \ac{AIT*} calculates a problem-specific
    admissible cost heuristic with a reverse \ac{LPA*} search ordered by the
    total potential solution cost according to an \textit{a priori} admissible
    cost heuristic. It finds valid paths with an edge-queue version of A*
    ordered by the total potential solution cost according to the cost
    heuristic calculated by its reverse search. \ac{EIT*} calculates
    problem-specific admissible and inadmissible cost and effort heuristics
    with a reverse A* search ordered by the total potential solution cost and
    effort according to \textit{a priori} cost and effort heuristics. It finds
    valid paths with an edge-queue version of \ac{AEES} ordered by the total
    potential solution cost and effort according to the heuristics calculated by
    its reverse search. The reverse and forward search algorithms of \ac{AIT*}
    and \ac{EIT*} are detailed in Sections~\ref{sec:adaptively-informed-trees}
    and~\ref{sec:effort-informed-trees}, respectively.}%
  \label{tbl:forward-and-reverse-searches}
  \vspace*{-0.4em}
\end{table*}

  % Transposed version
  % \begin{tabular}{llllllll}
  %   \toprule
  %            &  & \multicolumn{3}{l}{Reverse search}                              & \multicolumn{3}{l}{Forward search}\\
  %   Alg.                    & Approx.                         & Alg.      & Calculated heuristics          & Collision detection & Alg.  & Guiding heuristics                & Collision detection\\
  %   \midrule
  %   \ac{BIT*}             & \ac{RGG}s                & ---       & ---                           & ---                 & \ac{TLPA*} & \textit{A priori} admissible cost & Dense\\[1em]
  %   \ac{AIT*}             & \ac{RGG}s                & \ac{LPA*} & Admissible cost               & ---                 & A*         & Calculated admissible cost         & Dense\\[1em]
  %                         &                          &           & Admissible cost               &                     &            & Calculated admissible cost                   &\\
  %   \ac{EIT*}             & \ac{RGG}s                & A*        & Inadmissible cost             & Adaptive sparse     & \ac{AEES}  & Calculated inadmissible cost                 & Dense\\
  %                         &                          &           & Inadmissible effort           &                     &            & Calculated inadmissible effort               &\\
  %   \bottomrule
  % \end{tabular}\vspace{0.5em}

%%% Local Variables:
%%% mode: latex
%%% TeX-master: "../../../main"
%%% visual-line-mode: nil
%%% truncate-lines: t
%%% End:

\ac{AIT*} and \ac{EIT*} are almost-surely asymptotically optimal path planning algorithms that build on \ac{BIT*}. \ac{BIT*} approximates the state space with a batch of samples, which it views as an edge-implicit \ac{RGG}.
\ac{BIT*} searches this \ac{RGG} in order of the total potential solution
quality of its edges until it can guarantee that it has found the
\textit{resolution-optimal} solution, i.e., the optimal solution in the current
approximation of the state space.
Once the search is finished, \ac{BIT*} improves its \ac{RGG} approximation by adding a new batch of samples. In this way, \ac{BIT*} approximates and searches a continuously valued state space by building and searching a series of increasingly dense, edge-implicit \acp{RGG}.

To find solutions, \ac{BIT*} processes the implicit edges of its current
\ac{RGG} approximation in order of their total potential solution cost, using
incremental search techniques similar to an edge-queue version of
\ac{TLPA*}. It estimates the total potential solution cost of an edge as the
sum of the current cost to come to the source of the edge, a heuristic estimate
of the edge cost, and a heuristic estimate of the cost to go from the target of
the edge. The formal guarantees of \ac{BIT*} require that these cost heuristics
are admissible.

\ac{AIT*} and \ac{EIT*} extend \ac{BIT*} with an asymmetric bi\-directional
search that unifies many of the benefits reviewed in
\reft[Section]{sec:literature-review} by leveraging information implicit in the
observed distribution of valid states
\refp[Figure]{fig:benefits-of-adaptive-heuristics}. The reverse searches of
\ac{AIT*} and \ac{EIT*} calculate accurate heuristics which are exploited by
their forward searches. The forward searches in turn inform the reverse
searches if they used invalid edges to compute the heuristic. In this way, both
searches continuously inform each other with complementary information.

The reverse searches of \ac{AIT*} and \ac{EIT*} evaluate edges approximately
and are therefore computationally inexpensive. The forward searches of
\ac{AIT*} and \ac{EIT*} evaluate edges fully and are therefore computationally
expensive, but focused by the calculated problem-specific heuristics. This
computational asymmetry avoids the inefficiency of naive symmetric
bidirectional informed search, where frontiers of expensive searches pass each
other~\citep[Section 10.2, ][]{pohl_phd1969}.

An overview of the search algorithms used in \ac{AIT*} and \ac{EIT*} is provided in \reft[Table]{tbl:forward-and-reverse-searches}. The rest of this section presents the notation used in this paper \refp[Section]{sec:notation}, the algorithmic details of \ac{AIT*} and \ac{EIT*} (Sections~\ref{sec:adaptively-informed-trees} and~\ref{sec:effort-informed-trees}), and the formal analysis of their asymptotic optimality \refp[Section]{sec:analysis}.

\subsection{Notation}%
\label{sec:notation}

The state space is denoted by
\( X \subseteq \mathbb{R}^{n}, n \in \mathbb{N} \), the invalid states by
\( X_{\mathrm{invalid}} \subset X \), and the valid states by
\( X_{\mathrm{valid}} \coloneqq \mathop{\mathrm{closure}}\left( X \setminus
  X_{\mathrm{invalid}} \right) \). A single state is denoted by
\( \bm{\mathrm{x}} \in X \). The start and goal are denoted by
\( \bm{\mathrm{x}}_{\mathrm{start}} \in X_{\mathrm{valid}} \) and
\( X_{\mathrm{goal}} \subset X_{\mathrm{valid}} \), respectively. The set of
sampled states underlying the \ac{RGG} approximation is denoted by
\( X_{\mathrm{sampled}} \subset X_{\mathrm{valid}} \).

The forward and reverse search trees are denoted by
\( \mathcal{F} = (V_{\mathcal{F}}, E_{\mathcal{F}}) \) and
\( \mathcal{R} = (V_{\mathcal{R}}, E_{\mathcal{R}}) \), respectively. All
vertices in these trees, \( V_{\mathcal{F}} \) and \( V_{\mathcal{R}} \), are
individually associated with a sampled state and are embedded in the valid
region of the state space, \( X_{\mathrm{valid}} \). Up to two vertices can be
associated with a sample (one per search tree), but not every sample must be
associated with a vertex in a tree. All edges in both trees,
\( E_{\mathcal{F}} \subset V_{\mathcal{F}} \times V_{\mathcal{F}} \) and
\( E_{\mathcal{R}} \subset V_{\mathcal{R}} \times V_{\mathcal{R}} \), are
directed, defined by a source state, \( \bm{\mathrm{x}}_{\mathrm{s}} \), and a
target state, \( \bm{\mathrm{x}}_{\mathrm{t}} \), and denoted by an ordered
pair, \( ( \bm{\mathrm{x}}_{\mathrm{s}}, \bm{\mathrm{x}}_{\mathrm{t}} ) \). The
edges in the forward tree, \( E_{\mathcal{F}} \), are embedded in the valid
region of the state space and represent valid connections between sampled
states. The edges in the reverse tree, \( E_{\mathcal{R}} \), are not
necessarily embedded in the valid region of the state space and may lead
through invalid states.

The true connection cost between two states is denoted by the function
\( c \colon X \times X \to [0, \infty) \) and admissible estimates of this cost
are denoted by the function \( \hat{c} \colon X \times X \to [0, \infty) \),
i.e.,
\( \forall \bm{\mathrm{x}}_{i}, \bm{\mathrm{x}}_{j} \in X, \hat{c}\left(
  \bm{\mathrm{x}}_{i}, \bm{\mathrm{x}}_{j} \right) \leq c\left(
  \bm{\mathrm{x}}_{i}, \bm{\mathrm{x}}_{j} \right) \). Admissible cost
heuristics to come to a specific state from the start are denoted by the
function \( \hat{g} \colon X \to [0, \infty) \) and often defined as
\( \hat{g}\left( \bm{\mathrm{x}} \right) \coloneqq \hat{c}\left(
  \bm{\mathrm{x}}_{\mathrm{start}}, \bm{\mathrm{x}} \right) \). Admissible cost
heuristics to go from a specific state to a goal are denoted by the function
\( \hat{h} \colon X \to [0, \infty) \) and often defined as
\( \hat{h}\left( \bm{\mathrm{x}} \right) \coloneqq
\min_{\bm{\mathrm{x}}_{\mathrm{goal}} \in X_{\mathrm{goal}}} \left\{
  \hat{c}\left( \bm{\mathrm{x}}, \bm{\mathrm{x}}_{\mathrm{goal}} \right)
\right\} \).

The cost to come to a specific state from the start through the forward tree is
denoted by the function \( g_{\mathcal{F}} \colon X \to [0, \infty) \). It is
well-defined for states that have an associated vertex in the forward tree and
taken as infinity for states that do not.

Admissible estimates of the cost of a path from the start to a goal constrained
to go through a specific state is denoted by the function
\( \hat{f} \colon X \to [0, \infty) \) and often defined as
\( \hat{f}(\bm{\mathrm{x}}) \coloneqq \hat{g}(\bm{\mathrm{x}}) +
\hat{h}(\bm{\mathrm{x}}) \). This function defines the informed set, i.e., the
set of states that can improve the current solution,
\( X_{\hat{f}} \coloneqq \{ \bm{\mathrm{x}} \in X \,|\,
\hat{f}(\bm{\mathrm{x}}) < c_{\mathrm{current}} \} \), where
\( c_{\mathrm{current}} \) is the cost of the current
solution~\citep{gammell_tro2018}.

Square brackets denote a label, e.g., \( l[\bm{\mathrm{x}}] \in \mathbb{R} \)
refers to a real number, \( l \), associated with the state
\( \bm{\mathrm{x}} \). Labels keep their values until they are updated, i.e.,
they are used in \ac{AIT*} and \ac{EIT*} similar to how \( g \)-values are used
in A*.

The compounding operations, \( A \leftarrow A \cup B \) and
\( A \leftarrow A \setminus B \), are respectively denoted by
\( A \xleftarrow{\scriptscriptstyle +} B \) and \( A \xleftarrow{\scriptscriptstyle -} B \), where
\( A \) and \( B \) are subsets of a common set.

\subsubsection{\acs{EIT*}-specific Notation}%
\label{sec:eit-notation}

Potentially inadmissible effort heuristics between two states are denoted by
\( \bar{e}\colon X \times X \to [0, \infty) \). These heuristics estimate the
computational effort required to find and validate a path between two states,
e.g., the number of necessary collision detections on the path. Potentially
inadmissible effort heuristics between each state and the start are denoted by
the function \( \bar{d}\colon X \to [0, \infty) \) and often defined as
\( \bar{d}\left( \bm{\mathrm{x}} \right) \coloneqq
\bar{e}\left( \bm{\mathrm{x}}, \bm{\mathrm{x}}_{\mathrm{start}} \right) \).
Potentially inadmissible cost heuristics between two states are
denoted by the function \( \bar{c}\colon X \times X \to [0, \infty) \). It is
assumed that this estimate is never lower than its admissible counterpart,
i.e.,
\( \forall\; \bm{\mathrm{x}}_{1}, \bm{\mathrm{x}}_{2} \in X, \hat{c}\left(
  \bm{\mathrm{x}}_{1}, \bm{\mathrm{x}}_{2} \right) \leq \bar{c}\left(
  \bm{\mathrm{x}}_{1}, \bm{\mathrm{x}}_{2} \right) \).

\subsection{\acf{AIT*}}%
\label{sec:adaptively-informed-trees}

\ac{AIT*} improves on \ac{BIT*} by using the same increasingly dense \ac{RGG}
approximation but searching it with an asymmetric bidirectional search which
calculates and exploits a more accurate cost heuristic that is specific to each
\ac{RGG} approximation \refp[Figure]{fig:aitstar-step-by-step}. This results in
a more efficient search with fewer evaluated edges when an admissible cost
heuristic is available \textit{a priori}
(Figures~\ref{fig:example-bitstar-path-length},~\subref{fig:example-aitstar-path-length},
Extension 1),
and can improve initial solution times and convergence rates.

\ac{AIT*} consists of three high-level steps:
\begin{enumerate*}[(i), itemjoin={{, }}, itemjoin*={{; and }}]
\item improving the \ac{RGG} approximation (sampling;
  \reft[Section]{sec:ait-approximation})
\item updating the heuristic (reverse search; \reft[Section]{sec:ait-reverse-search})
\item finding valid paths in the current \ac{RGG} approximation (forward search;
  \reft[Section]{sec:ait-forward-search}),
\end{enumerate*}
as shown by Algorithm~\ref{alg:conceptual}. The full technical details of
\ac{AIT*} are given in
Algorithms~\ref{alg:aitstar:technical}--\ref{alg:aitstar:prune}.

The reverse search of \ac{AIT*} is a version of \ac{LPA*} that calculates
accurate cost heuristics by combining an admissible cost heuristic between
multiple states into a more accurate cost heuristic between each state and the
goal. The calculated cost heuristic is admissible for the current \ac{RGG} and
leverages information implicit in the observed distribution of valid
states. This reverse search is computationally inexpensive because it does not
perform collision detection on the edges.

If the reverse search finishes without reaching the start, then the start and
goal are not in the same connected component of the current \ac{RGG}
approximation. \ac{AIT*} skips the forward search in this case and directly
improves the \ac{RGG} approximation. This ensures that \ac{AIT*} does not spend
computational effort searching a graph that it knows cannot contain a solution.

The forward search of \ac{AIT*} is an edge-queue version of A* which
efficiently exploits the calculated heuristic and evaluates few edges that do
not contribute to a solution when admissible cost heuristic are available
\textit{a priori}. If the forward search detects a collision on an edge in the
reverse search tree, then \ac{LPA*} updates the heuristic by efficiently
repairing this tree. The forward search then continues with the updated
heuristic until the optimal solution on the current \ac{RGG} approximation is
found or another collision is detected on an edge in the reverse search
tree. This process is repeated as time allows to almost-surely asymptotically
converge towards the optimal solution in an anytime manner
\refp[Figure]{fig:aitstar-step-by-step}.

\begin{figure*}
  \begin{subfigure}[b]{0.195\textwidth}
    \input{figures/3-algorithms/3-1-adaptively-informed-trees/step-by-step/steps/1-iter-000215}%
    \caption{}%
    \label{fig:aitstar-step-1}%
  \end{subfigure}%
  \hfill%
  \begin{subfigure}[b]{0.195\textwidth}%
    \input{figures/3-algorithms/3-1-adaptively-informed-trees/step-by-step/steps/2-iter-000283}%
    \caption{}%
    \label{fig:aitstar-step-2}%
  \end{subfigure}%
  \hfill%
  \begin{subfigure}[b]{0.195\textwidth}%
    \input{figures/3-algorithms/3-1-adaptively-informed-trees/step-by-step/steps/3-iter-000693}%
    \caption{}%
    \label{fig:aitstar-step-3}%
  \end{subfigure}%
  \hfill%
  \begin{subfigure}[b]{0.195\textwidth}%
    \input{figures/3-algorithms/3-1-adaptively-informed-trees/step-by-step/steps/4-iter-000707}%
    \caption{}%
    \label{fig:aitstar-step-4}%
  \end{subfigure}%
  \hfill%
  \begin{subfigure}[b]{0.195\textwidth}%
    \input{figures/3-algorithms/3-1-adaptively-informed-trees/step-by-step/steps/5-iter-003136}%
    \caption{}%
    \label{fig:aitstar-step-5}%
  \end{subfigure}%
  \caption{Five snapshots of AIT*'s search when minimizing path length. The
    start and goal states are represented by a black dot~(\tikzinlinestart) and
    circle~(\tikzinlinegoal), respectively. Sampled states are represented by
    small black dots~(\tikzinlinestate). State space obstacles are indicated
    with gray obstacles~(\tikzinlineobstacle). The forward search tree is shown
    with black lines~(\tikzinlineforwardedge) and the reverse search tree with
    gray lines~(\tikzinlinereverseedge). The current best solution is
    highlighted in yellow~(\tikzinlinesolution). AIT* starts by initializing
    the \ac{RGG} approximation and calculating an approximation-specific
    admissible cost heuristic with a reverse search without collision
    detection~(\subref{fig:aitstar-step-1}). AIT* exploits the calculated
    heuristic with its forward search and repairs the reverse search tree
    whenever the forward search reveals that it contains an invalid
    edge~(\subref{fig:aitstar-step-2}). When the forward search finds the
    resolution-optimal solution, the \ac{RGG} approximation is improved by sampling and pruning and the heuristic is updated on this improved approximation~(\subref{fig:aitstar-step-3}). This updated heuristic is again exploited with the next forward search and repaired when found to use invalid edges~(\subref{fig:aitstar-step-4}). \ac{AIT*} repeats these steps until stopped and almost-surely asymptotically converges towards the optimal solution~(\subref{fig:aitstar-step-5}).}%
  \label{fig:aitstar-step-by-step}\vspace*{-1.0em}
\end{figure*}

%%% Local Variables:
%%% mode: latex
%%% TeX-master: "../../../../main"
%%% End:

\subsubsection{Approximation}%
\label{sec:ait-approximation}

\ac{AIT*} incrementally approximates the state space by sampling batches of \(
m \) valid states~(Alg.~\ref{alg:aitstar:technical},
line~\ref{alg:aitstar:improve-approximation:sample}). States are sampled
uniformly in the informed set, using informed sampling~\citep{gammell_tro2018}
when possible. These samples are viewed as a series of increasingly dense,
edge-implicit \ac{RGG}s where bidirectional edges are defined either by a
connection radius, \( r \), or by the \emph{mutual} \( k \)-nearest
neighbors~\citep[Alg.~\ref{alg:aitstar:neighbors},
line~\ref{alg:aitstar:neighbors:nearest}; mutual \( k \)-nearest as in][]{janson_ijrr2015}. The connection parameters, \( r \) and \( k \), scale as in \acs{PRM*}~\citep{karaman_ijrr2011}, using the measure of the informed set as in~\citet{gammell_ijrr2020},
\begin{align*}
  r(q) &\coloneqq 2 \eta {\left( 1 + \frac{1}{n} \right)}^{\frac{1}{n}} {\left(
         \frac{\min\left\{ \lambda\left( X \right), \lambda\left( X_{\hat{f}}
         \right) \right\}}{\lambda\left( B_{1, n} \right)} \right)}^{\frac{1}{n}} \\ &\qquad\quad{\left( \frac{\log\left( q \right)}{q} \right)}^{\frac{1}{n}} \\
  k(q) &\coloneqq \eta \, \mathrm{e} \left( 1 + \frac{1}{n} \right) \log\left( q \right),
\end{align*}
where \( q \) is the number of sampled states in the informed set,
\( \eta > 1 \) is a tuning parameter, \( \lambda(\cdot) \) denotes the Lebesgue
measure, and \( B_{1, n} \) is the \( n \)-dimensional unit ball.

The \( r \)-disc strategy can result in better performance than the \( k
\)-nearest version but the computation of the \( r \)-disc radius must be
adjusted to the properties of the state space~\citep{kleinbort_wafr2016,
kleinbort_afr2020}.
Faster-decreasing radii are presented in~\citet{janson_ijrr2015, janson_ijrr2018}, \citet{solovey_ijrr2020}, and \citet{tsao_icra2020}, but are not used in \ac{AIT*} and \ac{EIT*} for direct comparison to existing algorithms as they are presented in the literature.

\ac{AIT*} considers the combination of both this \ac{RGG} definition and any existing connections in the forward search tree and ignores edges known to be invalid~(Alg.~\ref{alg:aitstar:neighbors}, lines~\ref{alg:aitstar:neighbors:forward-children} and~\ref{alg:aitstar:neighbors:invalid}). \ac{RGG} complexity is reduced by pruning samples that are not in the informed set~(Alg.~\ref{alg:aitstar:technical}, line~\ref{alg:aitstar:improve-approximation:prune} and Alg.~\ref{alg:aitstar:prune}).

\subsubsection{Reverse Search}%
\label{sec:ait-reverse-search}

\ac{AIT*} calculates an accurate cost heuristic between each processed state and the goal that is admissible for the current \ac{RGG} approximation. It does this by calculating the \textit{a priori} admissible cost heuristic, \( \hat{c}\left( \,\cdot\,,\,\cdot\, \right) \), over the connectivity of this approximation.

This is achieved by processing the \ac{RGG} approximation with a version of \ac{LPA*}
that is rooted at the goal and uses the admissible \textit{a priori} cost
heuristics, \( \hat{c}\left( \,\cdot\,,\,\cdot\, \right) \), as edge costs
without detecting collisions on the edges. The resulting reverse search tree
can be updated efficiently because \ac{LPA*} stores the cost of a state when it
was first connected or last rewired and when it was last expanded, denoted by
\( \hat{h}_{\mathrm{con}}[\bm{\mathrm{x}}] \) and
\( \hat{h}_{\mathrm{exp}}[\bm{\mathrm{x}}] \), respectively. These are the
\( g \) and \( v \) values in a forward \ac{LPA*} search~\citep{aine_ai2016}.

The queue of the reverse \ac{LPA*} search in \ac{AIT*} is denoted by
\( \mathcal{Q}_{\mathcal{R}} \) and lexicographically ordered according to
\begin{align*}
  \mathtt{key}_{\mathcal{R}}^{\text{AIT*}}(\bm{\mathrm{x}}) \coloneqq \Big( &\min\left\{ \hat{h}_{\mathrm{con}}[\bm{\mathrm{x}}],  \hat{h}_{\mathrm{exp}}[\bm{\mathrm{x}}] \right\} + \hat{g}(\bm{\mathrm{x}}), \\
  &\min\left\{ \hat{h}_{\mathrm{con}}[\bm{\mathrm{x}}],  \hat{h}_{\mathrm{exp}}[\bm{\mathrm{x}}] \right\}\Big),
\end{align*}
where \( \hat{g}(\bm{\mathrm{x}}) \) denotes an admissible \textit{a priori}
cost heuristic between a state, \( \bm{\mathrm{x}} \), and the start. This key
is used to extract the next edge from the reverse queue
(Alg.~\ref{alg:aitstar:technical},
lines~\ref{alg:aitstar:iterate-reverse-search:get-best-vertex}
and~\ref{alg:aitstar:iterate-reverse-search:pop-best-vertex}).

An uninitialized LPA* search is used to calculate the heuristic on the first
batch of samples and after each new batch is
added. This is more efficient than incrementally updating the heuristic with
\ac{LPA*} for the large changes in the graph that result from increasing its
resolution~\citep{koenig_ai2004, likhachev_icaps2005b, aine_ai2016}. An
uninitialized \ac{LPA*} search is started by clearing the reverse search tree
(except for the goals), setting the \( \hat{h}_{\mathrm{con}} \) and
\( \hat{h}_{\mathrm{exp}} \) values of all states to infinity (again, except
for the goals), and inserting the goal states into the reverse queue~(Alg.~\ref{alg:aitstar:technical},
lines~\ref{alg:aitstar:technical:initialize-reverse},~\ref{alg:aitstar:technical:reinitialize-reverse-tree-begin},
and~\ref{alg:aitstar:technical:reinitialize-reverse-tree-end}).

The heuristic is updated whenever an edge in the reverse search tree is found to be invalid by removing this edge from the \ac{RGG} approximation and repairing the reverse search tree with \ac{LPA*}. This is accomplished by updating the cost-to-go of the source state of the invalid edge and then running \ac{LPA*} to update the cost of all affected states as necessary~(Alg.~\ref{alg:aitstar:technical}, lines~\ref{alg:aitstar:iterate-reverse-search:get-best-vertex}--\ref{alg:aitstar:iterate-reverse-search:end} and~\ref{alg:aitstar:iterate-forward-search:blacklist}--\ref{alg:aitstar:iterate-forward-search:invalidate-reverse-branch} and Alg.~\ref{alg:aitstar:invalidate-reverse-branch}).

The reverse search is suspended when the total potential solution cost of the best state in the reverse queue is greater than or equal to that of the best edge in the forward queue and the target of the best edge in the forward queue is consistent (Alg.~\ref{alg:aitstar:technical} lines~\ref{alg:aitstar:technical:continue-reverse-search-1} and~\ref{alg:aitstar:technical:continue-reverse-search-2}). This guarantees that no other edge in the forward queue would be better if the reverse search was continued~\citep{strub_phd2021}.
The reverse search is also suspended when the reverse or forward queue is empty or when all edges in the forward queue have consistent targets with a reverse-key value less than or equal to the minimum reverse-key in the reverse queue, but these conditions are omitted from Algorithm~\ref{alg:aitstar:technical} for clearer structure.

\begin{algorithm}[t]
  \caption{\acf{AIT*}}%
  \label{alg:aitstar:technical}
  \small
  \( \currentcost \leftarrow \infty \)\;
  \( \sampledstates \leftarrow \goalstates \cup \{ \startstate \}  \)\;
  \( \fwdvertices \leftarrow \startstate \)\texttt{;} \(
  \fwdedges \leftarrow \emptyset \)\texttt{;} \( \fwdqueue \leftarrow
  \expandedge{\startstate} \)\;\label{alg:aitstar:technical:initialize-forward}
  \( \revvertices \leftarrow \goalstates \)\texttt{;} \(
  \revedges \leftarrow \emptyset \)\texttt{;} \( \revqueue \leftarrow \goalstates \)\;\label{alg:aitstar:technical:initialize-reverse}
  \Repeat{\normalfont\tt stopped}{\label{alg:aitstar:technical:repeat-begin}
    \uIf{\normalfont \( \displaystyle\min_{\state \in
        \revqueue} \set*{\aitrevkey{\state}} < \min_{\edge{\sourcestate}{\targetstate}
        \in \fwdqueue}
      \set*{\aitfwdkey{\sourcestate}{\targetstate}}
      \)\label{alg:aitstar:technical:continue-reverse-search-1}\\\textbf{or} target of best edge in forward queue is inconsistent \label{alg:aitstar:technical:continue-reverse-search-2}}{
      \( \displaystyle\bm{\mathrm{x}} \leftarrow \mathop{\arg\,\min}_{\state
        \in \revqueue} \set*{ \aitrevkey{\state} } \)\;\label{alg:aitstar:iterate-reverse-search:get-best-vertex}
      \( \revqueue \setsubtract \bm{\mathrm{x}} \)\;\label{alg:aitstar:iterate-reverse-search:pop-best-vertex}
      \eIf{\( \hat{h}_{\mathrm{con}}\left[\bm{\mathrm{x}}\right] < \hat{h}_{\mathrm{exp}}\left[\bm{\mathrm{x}}\right] \)}{
        \( \hat{h}_{\mathrm{exp}}\left[\bm{\mathrm{x}}\right] \leftarrow \hat{h}_{\mathrm{con}}\left[\bm{\mathrm{x}}\right] \)\;\label{alg:aitstar:updateheuristic:line:hexp}
      }{
        \( \hat{h}_{\mathrm{exp}}\left[\bm{\mathrm{x}}\right] \leftarrow \infty \)\;
        \( \updatestate{\bm{\mathrm{x}}}\)\;
      }
      \For{\normalfont\textbf{all }\( \state[i] \in \neighborstates{\bm{\mathrm{x}}} \)}{
        \( \updatestate{\state[i]} \)\;\label{alg:aitstar:iterate-reverse-search:end}
      }
    }
    \uElseIf{\normalfont \( \displaystyle\hspace*{-1.0em} \min_{\edge{\sourcestate}{\targetstate} \in \fwdqueue} \hspace*{-0.2em}\set*{ \fwdctc{\sourcestate} +
  \adedgecost{\sourcestate}{\targetstate} + \concost{\targetstate} } < \currentcost \)}{\label{alg:aitstar:technical:continue-forward-search}
      \( \displaystyle\edge{\sourcestate}{\targetstate} \leftarrow \mathop{\arg\,\min}_{\edge{\sourcestate}{\targetstate} \in \fwdqueue}
  \set*{\aitfwdkey{\sourcestate}{\targetstate}} \)\;\label{alg:aitstar:iterate-forward-search:get-best-edge}
      \( \fwdqueue \setsubtract \left( \sourcestate, \targetstate \right) \)\;\label{alg:aitstar:iterate-forward-search:pop-best-edge}
      \uIf{\( (\sourcestate, \targetstate) \in \fwdedges \)}{\label{alg:aitstar:iterate-forward-search:is-edge-in-tree}
        \( \fwdqueue \setadd \expandedge{\targetstate}\)\;\label{alg:aitstar:iterate-forward-search:expand-edge-in-tree}
      }
      \ElseIf{\( \fwdctc{\sourcestate} + \adedgecost{\sourcestate}{\targetstate} < \fwdctc{\targetstate} \)}{\label{alg:aitstar:iterate-forward-search:can-edge-possibly-improve-tree}
        \uIf{\normalfont\(\isvalid{\sourcestate, \targetstate}\)}{\label{alg:aitstar:iterate-forward-search:collision-detection}
          \If{\( \fwdctc{\sourcestate} + \edgecost{\sourcestate}{\targetstate} + \hat{h}_{\mathrm{con}}\left[\targetstate\right] < \currentcost \)}{\label{alg:aitstar:iterate-forward-search:can-edge-actually-improve-solution}
            \If{\( \fwdctc{\sourcestate} + \edgecost{\sourcestate}{\targetstate} < \fwdctc{\targetstate} \)}{\label{alg:aitstar:iterate-forward-search:can-edge-actually-improve-tree}
              \eIf{\( \targetstate \not\in \fwdvertices \)}{\label{alg:aitstar:iterate-forward-search:add-state-to-tree-begin}
                \( \fwdvertices \setadd \targetstate \)\;\label{alg:aitstar:iterate-forward-search:add-state-to-tree-end}
          }{
            \( \fwdedges \setsubtract (\fwdparent{\targetstate}, \targetstate) \)\;\label{alg:aitstar:iterate-forward-search:rewiring}
          }
          \( \fwdedges \setadd (\sourcestate, \targetstate) \)\;\label{alg:aitstar:iterate-forward-search:add-edge-to-tree}
          \( \fwdqueue \setadd \expandedge{\targetstate} \)\;\label{alg:aitstar:iterate-forward-search:expand-child-state}
          \( \currentcost \leftarrow \min_{\goalstate \in \goalstates}\left\{ \fwdctc{\goalstate} \right\} \)\;\label{alg:aitstar:iterate-forward-search:update-solution-cost}
        }
      }
    }
    \Else{
      \( \invedges \setadd \set*{ \edge{\sourcestate}{\targetstate},
        \edge{\targetstate}{\sourcestate} } \)\;\label{alg:aitstar:iterate-forward-search:blacklist}
      \If{\normalfont \( \edge{\sourcestate}{\targetstate} \in \revedges \)}{\label{alg:aitstar:iterate-forward-search:reverse-tree-check}
        \( \invalidaterevbranch{\sourcestate} \)\;\label{alg:aitstar:iterate-forward-search:invalidate-reverse-branch}
      }
    }
  }
    }
    \Else{
      \( \prunestates{\sampledstates} \)\;\label{alg:aitstar:improve-approximation:prune}
      \( \sampledstates \setadd \samplestates{m, \currentcost}
      \)\;\label{alg:aitstar:improve-approximation:sample}
      \( \revvertices \leftarrow \goalstates \)\texttt{;} \(
      \revedges \leftarrow \emptyset \)\texttt{;} \( \revqueue \leftarrow
      \goalstates \)\;\label{alg:aitstar:technical:reinitialize-reverse-tree-begin}
      \( \forall \state \in \sampledstates \setminus \goalstates,\; \concost{\state} = \expcost{\state} = \infty \)\;\label{alg:aitstar:technical:reinitialize-reverse-tree-end}
      \( \fwdqueue \leftarrow \expandedge{\startstate} \)\;\label{alg:aitstar:reinitialize-queues:expand-start}
    }
  }\label{alg:aitstar:technical:repeat-end}
\end{algorithm}%

%%% Local Variables:
%%% mode: latex
%%% TeX-master: "../../../../main"
%%% End:

\subsubsection{Forward Search}%
\label{sec:ait-forward-search}

\ac{AIT*} finds solutions to a planning problem by building a search tree
rooted at the start with an edge-queue version of A* that uses the heuristic
calculated by the reverse search. The edge-queue of the forward search is
denoted by \( \mathcal{Q}_{\mathcal{F}} \) and lexicographically ordered by
\begin{align*}
  \mathtt{key}_{\mathcal{F}}^{\text{AIT*}}\left( \bm{\mathrm{x}}_{\mathrm{s}}, \bm{\mathrm{x}}_{\mathrm{t}} \right) \coloneqq \Big(&g_{\mathcal{F}}(\bm{\mathrm{x}}_{\mathrm{s}}) + \hat{c}(\bm{\mathrm{x}}_{\mathrm{s}}, \bm{\mathrm{x}}_{\mathrm{t}}) + \hat{h}_{\mathrm{con}}[\bm{\mathrm{x}}_{\mathrm{t}}],\\
  &g_{\mathcal{F}}(\bm{\mathrm{x}}_{\mathrm{s}}) + \hat{c}(\bm{\mathrm{x}}_{\mathrm{s}}, \bm{\mathrm{x}}_{\mathrm{t}}),\; g_{\mathcal{F}}(\bm{\mathrm{x}}_{\mathrm{s}})\Big),
\end{align*}
similar to how the edge-queue is ordered in \ac{BIT*}.

A forward search iteration begins by testing if the forward queue contains an edge that can possibly improve the current solution (Alg.~\ref{alg:aitstar:technical}, line~\ref{alg:aitstar:technical:continue-forward-search}).
If it does, then the edge with the lowest \( \mathtt{key}_{\mathcal{F}}^{\text{AIT*}} \)-value is extracted from the forward queue~(Alg.~\ref{alg:aitstar:technical}, lines~\ref{alg:aitstar:iterate-forward-search:get-best-edge} and~\ref{alg:aitstar:iterate-forward-search:pop-best-edge}). If this edge is already in the forward tree, then its target is expanded into the forward queue and the iteration is complete~(Alg.~\ref{alg:aitstar:technical}, lines~\ref{alg:aitstar:iterate-forward-search:is-edge-in-tree} and~\ref{alg:aitstar:iterate-forward-search:expand-edge-in-tree}).

If the edge is not in the forward tree but can possibly improve it, then it is checked for validity which is computationally expensive~(Alg.~\ref{alg:aitstar:technical}, lines~\ref{alg:aitstar:iterate-forward-search:can-edge-possibly-improve-tree} and~\ref{alg:aitstar:iterate-forward-search:collision-detection}). If the edge is invalid, then it is added to the set of invalid edges (Alg.~\ref{alg:aitstar:technical}, line~\ref{alg:aitstar:iterate-forward-search:blacklist}) and if it is also part of the reverse tree, then the heuristic is updated with the reverse search~(Alg.~\ref{alg:aitstar:technical}, lines~\ref{alg:aitstar:iterate-reverse-search:get-best-vertex}--\ref{alg:aitstar:iterate-reverse-search:end},~\ref{alg:aitstar:iterate-forward-search:reverse-tree-check},~\ref{alg:aitstar:iterate-forward-search:invalidate-reverse-branch}, and Alg.~\ref{alg:aitstar:invalidate-reverse-branch}). If the edge is valid, then its true cost is evaluated which may also be computationally expensive and it is checked whether the edge can actually improve the current solution and forward tree~(Alg.~\ref{alg:aitstar:technical}, lines~\ref{alg:aitstar:iterate-forward-search:can-edge-actually-improve-solution} and~\ref{alg:aitstar:iterate-forward-search:can-edge-actually-improve-tree}).

If the edge can improve the current solution and forward search tree, then its target is added to this tree if it is not already in it~(Alg.~\ref{alg:aitstar:technical}, lines~\ref{alg:aitstar:iterate-forward-search:add-state-to-tree-begin} and~\ref{alg:aitstar:iterate-forward-search:add-state-to-tree-end}). If it is already in the forward search tree, then the new edge constitutes a rewiring and the old edge is removed from the tree~(Alg.~\ref{alg:aitstar:technical}, line~\ref{alg:aitstar:iterate-forward-search:rewiring}). The new edge is added to the forward tree and its target is expanded regardless of whether the target was already in the tree or not~(Alg.~\ref{alg:aitstar:technical}, lines~\ref{alg:aitstar:iterate-forward-search:add-edge-to-tree} and \ref{alg:aitstar:iterate-forward-search:expand-child-state}).

\begin{algorithm}[t]
  \caption{\small\acs{AIT*}: \( \neighborstates{\state} \)}%
  \label{alg:aitstar:neighbors}
  \small
  \( \neighborvertices \leftarrow \neareststates{\state} \) \;\label{alg:aitstar:neighbors:nearest}
  \( \neighborvertices \setadd
  \{ \fwdparent{\state} \hspace*{-0.1em}\cup\hspace*{-0.1em} \fwdchildren{\state} \} \hspace*{-0.2em}\setminus\hspace*{-0.2em}
  \neighborvertices \)\;\label{alg:aitstar:neighbors:forward-children}
  \( \neighborvertices \setsubtract \set*{\state[i] \in \neighborvertices
    \given \edge{\state}{\state[i]} \in \invedges } \)\;\label{alg:aitstar:neighbors:invalid}
  \Return{ \( \neighborvertices \) }
\end{algorithm}%

%%% Local Variables:
%%% mode: latex
%%% TeX-master: "../../../../main"
%%% End:

\begin{algorithm}[t]
  \caption{\small\acs{AIT*}: \( \expandedge{\state} \)}%
  \label{alg:aitstar:expand-edge}
  \small
  \( \outedges \leftarrow \emptyset \)\;
  \For{\normalfont\textbf{all} \( \state[i] \in \neighborstates{\state} \)}{
    \( \outedges \setadd ( \state, \state[i] ) \)\;
  }
  \Return{ \( \outedges \) }
\end{algorithm}%

%%% Local Variables:
%%% mode: latex
%%% TeX-master: "../../../../main"
%%% End:

\begin{algorithm}[t]
  \caption{\small\acs{AIT*}: \( \updatestate{\state} \)}%
  \label{alg:aitstar:update-state}
  \small
  \If{\( \bm{\mathrm{x}} \neq \startstate \)}{
    \( \sourcestate \leftarrow \hspace*{-1em}\mathop{\arg\,\min}_{\state[i] \in
      \neighborstates{\state}} \set*{
      \expcost{\state[i]} + \adedgecost{\state}{\state[i]} } \)\;\label{alg:aitstar:update-state:cost-check}
    \eIf{\( \state \in \revvertices \)}{
      \( \revedges \setsubtract (\revparent{\state}, \state) \)\;
    }{
      \( \revvertices \setadd \state \)\;
    }
    \( \revedges \setadd (\sourcestate, \state) \)\;
    \( \concost{\state} \leftarrow \expcost{\sourcestate} + \adedgecost{\state}{\sourcestate}\label{alg:aitstar:update-state:cost-update} \)\;
    \uIf{\( \concost{\state} \neq \expcost{\state} \)}{
      \If{\( \state \not\in \revqueue \)}{
        \( \revqueue \setadd \state \)
      }
    }
    \ElseIf{\( \state \in \revqueue \)}{
      \( \revqueue \setsubtract \state \)
    }
  }
\end{algorithm}

%%% Local Variables:
%%% mode: latex
%%% TeX-master: "../../../../main"
%%% End:

\begin{algorithm}[t]
  \caption{\small\acs{AIT*}: \( \invalidaterevbranch{\state} \)}%
  \label{alg:aitstar:invalidate-reverse-branch}
  \small
  \If{\normalfont \( \state \not\in \goalstates \)}{\label{alg:aitstar:invalidate-reverse-branch:begin}
    \( \concost{\state} \leftarrow \infty \) \;
    \( \revedges \setsubtract \edge{\revparent{\state}}{\state} \) \;
  }
  \( \expcost{\state} \leftarrow \infty \) \;
  \( \revqueue \setsubtract \state \) \;
  \For{\normalfont \( \childstate \in \revchildren{\state} \)}{
    \( \invalidaterevbranch{\childstate} \)\;
  }
  \( \updatestate{\state} \)\;\label{alg:aitstar:invalidate-reverse-branch:end}
\end{algorithm}%

%%% Local Variables:
%%% mode: latex
%%% TeX-master: "../../../main"
%%% End:

\begin{algorithm}[t]
  \caption{\small\acs{AIT*}: \( \prunestates{\vertices, \edges, \sampledstates} \)}%
  \label{alg:aitstar:prune}
  \small
    \( \sampledstates \leftarrow \set*{\state \in \sampledstates \given
    \adsolcost{\state} \leq \currentcost} \) \;
  \( \fwdvertices \leftarrow \set*{\state \in \fwdvertices \given
    \adsolcost{\state} \leq \currentcost} \) \;
  \( \fwdedges \leftarrow \set*{\edge{\sourcestate}{\targetstate} \in \fwdedges
  \given \max \set*{\adsolcost{\sourcestate}, \adsolcost{\targetstate}} \leq \currentcost} \) \;
\end{algorithm}%

%%% Local Variables:
%%% mode: latex
%%% TeX-master: "../../../../main"
%%% End:

A forward search iteration finishes by updating the current solution cost~(Alg.~\ref{alg:aitstar:technical}, line~\ref{alg:aitstar:iterate-forward-search:update-solution-cost}). In practice this is done efficiently by only checking the goals in the forward tree.

The entire forward search terminates when it is guaranteed that the optimal
solution in the current \ac{RGG} approximation is found.
This occurs when no edge in the forward queue can possibly improve the
current solution~(Alg.~\ref{alg:aitstar:technical},
\lnereft{alg:aitstar:technical:continue-forward-search}).
The forward search also terminates when the start and goal are not in the same
connected component of the \ac{RGG} approximation.
This occurs when the reverse search tree does not reach any edge in the forward
queue, but this condition is omitted from Algorithm~\ref{alg:aitstar:technical} for clearer structure.

The three steps of \ac{AIT*}, i.e., improving the \ac{RGG} approximation, updating the heuristic with the reverse search, and finding valid paths with the forward search, are repeated for as long as computational time allows or until a suitable solution is found. This results in increasingly accurate cost heuristics for increasingly efficient searches of increasingly accurate \ac{RGG} approximations and will almost-surely asymptotically converge to the optimal solution in the limit of infinite samples \refp[Section]{sec:ait-analysis}.

\subsection{\acf{EIT*}}%
\label{sec:effort-informed-trees}

Informed planning algorithms guided by admissible cost heuristics, such as
\ac{BIT*}, \ac{ABIT*}, and \ac{AIT*}, need effective \textit{a priori}
admissible cost heuristics to provide benefits over uninformed algorithms. Such
heuristics may not exist because the available admissible cost heuristics may
be too computationally expensive or too inaccurate to be effective, even for
\ac{AIT*}. \ac{EIT*} builds on \ac{AIT*} by exploiting problem-specific
information in a way that leverages informative admissible cost heuristics when
they are available but can still search problems effectively when they are
not. It achieves this by leveraging additional types of problem-specific
information, including information on the computational effort required to
validate a path. This generalizes asymptotically optimal informed path planning
algorithms to a broader class of problems that include those without effective
\textit{a priori} cost heuristics.

\ac{EIT*} consists of the same three high-level steps as \ac{AIT*}:
\begin{enumerate*}[(i), itemjoin={{, }}, itemjoin*={{; and }}]
\item improving the \ac{RGG} approximation (sampling;
  \reft[Section]{sec:ait-approximation})
\item updating the heuristics (reverse search;
  \reft[Section]{sec:eit-reverse-search})
\item finding valid paths in the current \ac{RGG} approximation (forward search; \reft[Section]{sec:eit-forward-search}),
\end{enumerate*}
as shown in Algorithm~\ref{alg:conceptual}. Identically to \ac{AIT*}, \ac{EIT*}
also approximates the state space with a series of increasingly dense \ac{RGG}s
and skips the forward search if the reverse search terminates without reaching
the start. In contrast to \ac{AIT*}, the reverse search of \ac{EIT*} includes
adaptive sparse collision detection on the edges and calculates both
problem-specific path-cost and search-effort heuristics which are exploited in
an anytime manner by the forward search of \ac{EIT*}. The full technical
details of \ac{EIT*} are given in Algorithms~\ref{alg:eitstar:technical}
and~\ref{alg:eitstar:get-best-forward-edge} using the same \( \neighborstates*{} \), \( \expandstate*{}
\), and \( \prunestates*{} \) subroutines as
\ac{AIT*} (Algs.~\ref{alg:aitstar:neighbors},~\ref{alg:aitstar:expand-edge}, and~\ref{alg:aitstar:prune}).

\begin{figure*}
  \begin{subfigure}[b]{0.195\textwidth}
    \input{figures/3-algorithms/3-2-effort-informed-trees/step-by-step/steps/1-iter-000003}%
    \caption{}%
    \label{fig:eitstar-step-1}%
  \end{subfigure}%
  \hfill%
  \begin{subfigure}[b]{0.195\textwidth}%
    \input{figures/3-algorithms/3-2-effort-informed-trees/step-by-step/steps/2-iter-000007}%
    \caption{}%
    \label{fig:eitstar-step-2}%
  \end{subfigure}%
  \hfill%
  \begin{subfigure}[b]{0.195\textwidth}%
    \input{figures/3-algorithms/3-2-effort-informed-trees/step-by-step/steps/3-iter-000012}%
    \caption{}%
    \label{fig:eitstar-step-3}%
  \end{subfigure}%
  \hfill%
  \begin{subfigure}[b]{0.195\textwidth}%
    \input{figures/3-algorithms/3-2-effort-informed-trees/step-by-step/steps/4-iter-000019}%
    \caption{}%
    \label{fig:eitstar-step-4}%
  \end{subfigure}%
  \hfill%
  \begin{subfigure}[b]{0.195\textwidth}%
    \input{figures/3-algorithms/3-2-effort-informed-trees/step-by-step/steps/5-iter-000079}%
    \caption{}%
    \label{fig:eitstar-step-5}%
  \end{subfigure}%
  \caption{Five snapshots of EIT*'s search when optimizing obstacle
    clearance. The start and goal specifications are represented by a black
    dot~(\tikzinlinestart) and circle~(\tikzinlinegoal), respectively. Sampled
    states are represented by small black dots~(\tikzinlinestate). State space
    obstacles are indicated with gray rectangles~(\tikzinlineobstacle). The
    forward search tree is shown with black lines~(\tikzinlineforwardedge) and
    the reverse search tree with gray lines~(\tikzinlinereverseedge). The
    current best solution is highlighted in yellow~(\tikzinlinesolution). EIT*
    starts by initializing the \ac{RGG} approximation and calculating
    approximation-specific cost and effort heuristics with a reverse search
    without collision detection~(\subref{fig:eitstar-step-1}). EIT* exploits
    the calculated heuristics to guide its forward search and repairs the
    reverse search tree whenever the forward search reveals that it contains an
    invalid edge. The forward search is initially ordered on the least
    calculated effort-to-go from a state to the goal, which results in fast
    initial solution times~(\subref{fig:eitstar-step-2}). Once the initial
    solution is found, the forward search uses the calculated cost heuristics to
    find the resolution-optimal solution on the current \ac{RGG} approximation~(\subref{fig:eitstar-step-3}). Having found the resolution optimal solution on an \ac{RGG} approximation, \ac{EIT*} improves this approximation, updates the heuristic, and aims to find the next best resolution-optimal solution with minimal computational effort~(\subref{fig:eitstar-step-4}). This process is repeated until the algorithm is stopped and will almost-surely asymptotically converge towards the optimal solution~(\subref{fig:eitstar-step-5}).}%
  \label{fig:eitstar-step-by-step}
\end{figure*}

%%% Local Variables:
%%% mode: latex
%%% TeX-master: "../../../../main"
%%% End:

\subsubsection{Reverse Search}%
\label{sec:eit-reverse-search}

\ac{EIT*} calculates an admissible cost heuristic, an inadmissible cost
heuristic, and an inadmissible effort heuristic for each \ac{RGG}
approximation. The calculated admissible cost heuristic is a lower bound on the
optimal cost of a path from a state to the goal and is denoted by the label
\( \hat{h}[\,\cdot\,] \). The calculated inadmissible cost heuristic
approximates the cost of an optimal path from a state to the goal and is
denoted by the label \( \bar{h}[\,\cdot\,] \). This inadmissible cost heuristic
is often more accurate than its admissible analogue because it can capture more
problem-specific knowledge, including information that may overestimate the
true cost. The calculated inadmissible effort heuristic approximates the
computational effort required to find and validate a path from a state to the
goal and is denoted by the label \( \bar{e}[\,\cdot\,] \). An example of such a
heuristic is the number of collision checks required to validate a path, which
is available and informative for all planning problems as it only depends on
path length and collision detection resolution and not on the optimization
objective.

These heuristics are computed as in \ac{AIT*} with a reverse search that
combines \textit{a priori} heuristics between multiple states into more
accurate heuristics between each state and the goal. The calculated admissible
cost heuristic, \( \hat{h}[\,\cdot\,] \), is computed by combining \textit{a
  priori} admissible cost heuristics, \( \hat{c}(\,\cdot\,,\,\cdot\,) \), with
a reverse search that preserves the admissibility of the heuristic between each
state and the goal. The calculated inadmissible cost and effort heuristics,
\( \bar{h}[\,\cdot\,] \) and \( \bar{e}[\,\cdot\,] \), are similarly computed
with the inadmissible \textit{a priori} cost and effort heuristics,
\( \bar{c}\left(\,\cdot\,,\,\cdot\,\right) \) and
\( \bar{e}\left(\,\cdot\,,\,\cdot\,\right) \). All three heuristics always have
a value of zero for any goal state.

The reverse search of \ac{EIT*} is an edge-queue version of A* with adaptive
sparse collision detection.
% The calculated heuristics are updated in \ac{EIT*}
% when the forward search detects an invalid edge in the reverse search tree by
% removing the invalid edge from the \ac{RGG} approximation and restarting the
% reverse search with increased collision detection resolution.
Collision detection is traditionally considered a computationally expensive
operation in sampling-based planning~\citep{hauser_icra2015, kleinbort_afr2020} but
this is due to the com\-pu\-tational cost of validating valid
edges~\citep{sanchez_rr2003}. Detecting invalid edges with sparse collision
detection is computationally cheaper and was found to be of similar
computational cost to other operations in the reverse search when solving the
problems presented in \reft[Section]{sec:experimental-results}.

The queue of the reverse A* search in \ac{EIT*} is denoted by
\( \mathcal{Q}_{\mathcal{R}} \) and ordered lexicographically according to
\begin{align*}
  \mathtt{key}_{\mathcal{R}}^{\text{EIT*}}\left( \bm{\mathrm{x}}_{\mathrm{s}}, \bm{\mathrm{x}}_{\mathrm{t}} \right) \coloneqq \Big( &\hat{h}\left[ \bm{\mathrm{x}}_{\mathrm{s}} \right] + \hat{c}\left( \bm{\mathrm{x}}_{\mathrm{s}}, \bm{\mathrm{x}}_{\mathrm{t}} \right) + \hat{g}\left( \bm{\mathrm{x}}_{\mathrm{t}} \right), \\
                                                                                                                     &\bar{e}\left[ \bm{\mathrm{x}}_{\mathrm{s}} \right] + \bar{e}\left( \bm{\mathrm{x}}_{\mathrm{s}}, \bm{\mathrm{x}}_{\mathrm{t}} \right) + \bar{d}\left( \bm{\mathrm{x}}_{\mathrm{t}} \right) \Big),
\end{align*}
where \( \hat{g}(\bm{\mathrm{x}}_{\mathrm{t}}) \) and
\( \bar{d}(\bm{\mathrm{x}}_{\mathrm{t}}) \) denote admissible \textit{a priori}
cost and inadmissible \textit{a priori} effort heuristics for a path from the
target state, \( \bm{\mathrm{x}}_{\mathrm{t}} \), to the start. The two parts
of the key represent the total potential solution cost of a path through an
edge and the total potential computational effort required to validate a path
through an edge, respectively. The first part of the key ensures the
admissibility of the calculated cost heuristic and the second part of the key
ensures tiebreaks in favor of lower estimated effort, which is important if
only the trivially admissible cost heuristic is available, i.e.,
\( \forall \bm{\mathrm{x}}, \bm{\mathrm{x}}^{\prime} \in X,
\hat{c}(\bm{\mathrm{x}}, \bm{\mathrm{x}}^{\prime} ) \equiv \hat{g}(
\bm{\mathrm{x}} ) \equiv 0 \).

New heuristics are calculated when the \ac{RGG} approximation is initialized or
improved and updated when the forward search detects that the heuristics were calculated with an invalid edge. If the heuristics are calculated because of an initialized or improved approximation, then the resolution of the adaptive sparse collision detection is reset to a user-specified parameter (Alg.~\ref{alg:eitstar:technical}, lines~\ref{alg:eitstar:technical:initialize-cd-resolution} and~\ref{alg:eitstar:improve-approximation:reinitialize-cd-resolution}). If they are updated because of an invalid edge, then the resolution of the sparse collision detection in the reverse search is increased~(Alg.~\ref{alg:eitstar:technical}, line~\ref{alg:eitstar:iterate-forward-search:update-cd-resolution}).

Each iteration of the reverse search extracts the edge with the lowest \(
\mathtt{key}_{\mathcal{R}}^{\text{EIT*}} \)-value from the reverse queue
(Alg.~\ref{alg:eitstar:technical},
lines~\ref{alg:eitstar:iterate-reverse-search:get-best-rev-edge}
and~\ref{alg:eitstar:iterate-reverse-search:remove-best-rev-edge}) and checks
\( d \) evenly distributed states along this edge for collision
(Alg.~\ref{alg:eitstar:technical},
line~\ref{alg:eitstar:iterate-reverse-search:could-be-valid}). If a collision
is found, then the edge is added to the set of invalid edges
(Alg.~\ref{alg:eitstar:technical},
line~\ref{alg:eitstar:iterate-reverse-search:remember-invalid-edge}), otherwise
it is used to improve the inadmissible cost- and effort heuristics, if possible
(Alg.~\ref{alg:eitstar:technical},
lines~\ref{alg:eitstar:iterate-reverse-search:update-inad-cost} and~\ref{alg:eitstar:iterate-reverse-search:update-inad-effort}).

The edge is then checked if it can improve the admissible cost heuristic of its target~(Alg.~\ref{alg:eitstar:technical}, line~\ref{alg:eitstar:iterate-reverse-search:admissible-cost-test}). If it can, then the heuristic is updated and the target is either rewired or added to the reverse search tree~(Alg.~\ref{alg:eitstar:technical}, lines~\ref{alg:eitstar:iterate-reverse-search:admissible-cost-update}--\ref{alg:eitstar:iterate-reverse-search:insert-state-in-tree}). The reverse search iteration is completed by expanding the outgoing edges of the target into the reverse queue.

Similar to \ac{AIT*}, the reverse search is suspended when the total potential solution cost of the best edge in the reverse queue is greater than or equal to that of the best edge in the forward queue and the target of the best edge in the forward queue is closed~(Alg.~\ref{alg:eitstar:technical} lines~\ref{alg:eitstar:technical:continue-reverse-search-1} and~\ref{alg:eitstar:technical:continue-reverse-search-2}). This guarantees that no other edge in the forward queue would be better if the reverse search was continued~\citep{strub_phd2021}.
The reverse search is also suspended when the reverse or forward queue is empty, when all edges in the forward queue have closed targets, or when the inflation factor is infinity and any edge in the forward queue has a target in the reverse tree, but these conditions are omitted from Algorithm~\ref{alg:eitstar:technical} for clearer structure.

\subsubsection{Forward Search}%
\label{sec:eit-forward-search}

The forward search of \ac{EIT*} is an edge-queue version of \ac{AEES} which
exploits the cost and effort heuristics calculated by the reverse search of
\ac{EIT*} in an anytime manner. It leverages problem-specific cost and effort
information and results in effective searches with fast initial solution times
even when no admissible cost heuristics are available \textit{a priori}
(\reft[Figures]{fig:evaluated-edges}\subref{fig:example-eitstar-path-length},~\subref{fig:example-eitstar-obstacle-clearance},
Extension 2).

\ac{AEES} searches the same \ac{RGG} approximation multiple times with
successively tighter suboptimality bounds. It initially prioritizes quickly
finding any solution over efficiently finding the resolution-optimum, which
improves anytime performance. \ac{AEES} is especially useful when no
informative admissible cost heuristic is available \textit{a priori} because it
can exploit an effort heuristic to guide its search. Once an initial solution
is found, \ac{EIT*} uses both the calculated admissible and inadmissible cost
heuristics to improve the tree until it finds the resolution-optimum.

The forward search of \ac{EIT*} orders its queue by considering
\begin{enumerate*}[(i), itemjoin={{, }}, itemjoin*={{; and }}]
\item a lower bound on the optimal solution cost
\item an estimate of the optimal solution cost
\item and an estimate of the minimum remaining effort to validate a solution
  within the current suboptimality bound.
\end{enumerate*}

\paragraph{\small Optimal cost bound}

A lower bound on the optimal solution cost in the current \ac{RGG}
approximation is computed as
\begin{align*}
  \mathop{\min}_{(\bm{\mathrm{x}}_{\mathrm{s}},
  \bm{\mathrm{x}}_{\mathrm{t}}) \in \mathcal{Q}_{\mathcal{F}}}
  \hat{s}\left( \bm{\mathrm{x}}_{\mathrm{s}}, \bm{\mathrm{x}}_{\mathrm{t}}
  \right)
\end{align*}
where
\( \hat{s} \colon X_{\mathrm{sampled}} \times X_{\mathrm{sampled}} \to [0, \infty)
\) estimates the solution cost through an edge as
\begin{align*}
\hat{s}\left( \bm{\mathrm{x}}_{\mathrm{s}}, \bm{\mathrm{x}}_{\mathrm{t}} \right) \coloneqq g_{\mathcal{F}}\left( \bm{\mathrm{x}}_{\mathrm{s}} \right) + \hat{c}\left( \bm{\mathrm{x}}_{\mathrm{s}}, \bm{\mathrm{x}}_{\mathrm{t}} \right) + \hat{h}\left[ \bm{\mathrm{x}}_{\mathrm{t}} \right],
\end{align*}
where \( g_{\mathcal{F}}\left( \bm{\mathrm{x}}_{\mathrm{s}} \right) \) is the
cost to come to the source of the edge,
\( \hat{c}\left( \bm{\mathrm{x}}_{\mathrm{s}}, \bm{\mathrm{x}}_{\mathrm{t}}
\right) \) is the admissible cost heuristic of the edge, and
\( \hat{h}\left[ \bm{\mathrm{x}}_{\mathrm{t}} \right] \) is the calculated
admissible cost heuristic to go from the target of the edge. At least one edge
in the forward queue has an optimally connected source state~\citep[Lemma 1,
][]{hart_tssc1968}. The edge with the smallest \( \hat{s} \)-value in the queue
is therefore a lower bound on the optimal solution cost in the current \ac{RGG}
approximation. It is denoted as
\begin{align*}
  (\bm{\mathrm{x}}_{\mathrm{s}}^{\hat{s}}, \bm{\mathrm{x}}_{\mathrm{t}}^{\hat{s}}) \coloneqq \mathop{\arg\min}_{(\bm{\mathrm{x}}_{\mathrm{s}}, \bm{\mathrm{x}}_{\mathrm{t}}) \in \mathcal{Q}_{\mathcal{F}}} \hat{s}\left( \bm{\mathrm{x}}_{\mathrm{s}}, \bm{\mathrm{x}}_{\mathrm{t}} \right).
\end{align*}

\paragraph{\small Optimal cost estimate}

A more accurate, but possibly inadmissible, estimate of the optimal solution
cost in the current \ac{RGG} approximation is computed as
\begin{align*}
  \mathop{\min}_{(\bm{\mathrm{x}}_{\mathrm{s}}, \bm{\mathrm{x}}_{\mathrm{t}}) \in \mathcal{Q}_{\mathcal{F}}} \bar{s}\left( \bm{\mathrm{x}}_{\mathrm{s}}, \bm{\mathrm{x}}_{\mathrm{t}} \right),
\end{align*}
where
\( \bar{s} \colon X_{\mathrm{sampled}} \times X_{\mathrm{sampled}} \to [0, \infty)
\) is also an estimate of the solution cost through an edge but with the
inadmissible heuristics \( \bar{c} \) and \( \bar{h} \) instead of the
admissible heuristics \( \hat{c} \) and \( \hat{h} \). It is defined as
\begin{align*}
  \bar{s}\left( \bm{\mathrm{x}}_{\mathrm{s}}, \bm{\mathrm{x}}_{\mathrm{t}} \right) \coloneqq g_{\mathcal{F}}(\bm{\mathrm{x}}_{\mathrm{s}}) + \bar{c}\left( \bm{\mathrm{x}}_{\mathrm{s}}, \bm{\mathrm{x}}_{\mathrm{t}} \right) + \bar{h}\left[ \bm{\mathrm{x}}_{\mathrm{t}} \right].
\end{align*}
This estimate is often more accurate than the admissible lower bound because it
can use information that may overestimate the true cost. The edge that leads to
this possibly inadmissible estimate of optimal solution cost is denoted as
\begin{align*}
  (\bm{\mathrm{x}}_{\mathrm{s}}^{\bar{s}},
  \bm{\mathrm{x}}_{\mathrm{t}}^{\bar{s}}) \coloneqq
  \mathop{\arg\min}_{(\bm{\mathrm{x}}_{\mathrm{s}},
  \bm{\mathrm{x}}_{\mathrm{t}}) \in \mathcal{Q}_{\mathcal{F}}} \bar{s}\left( \bm{\mathrm{x}}_{\mathrm{s}}, \bm{\mathrm{x}}_{\mathrm{t}} \right).
\end{align*}

\begin{algorithm}[t]
  \caption{\acf{EIT*}}%
  \label{alg:eitstar:technical}
  \small
  \( \currentcost \leftarrow \infty \)\;
  \( \sampledstates \leftarrow \goalstates \cup \{ \startstate \}  \)\;
  \( \fwdvertices \leftarrow \startstate \)\texttt{;} \(
  \fwdedges \leftarrow \emptyset \)\texttt{;} \( \fwdqueue \leftarrow \emptyset\)\;
  \( \revvertices \leftarrow \goalstates \)\texttt{;} \( \closedrevvertices \leftarrow \emptyset \)\texttt{;} \( \revedges \leftarrow \emptyset \)\texttt{;} \( \revqueue \leftarrow \emptyset \)\;\label{alg:eitstar:technical:initialize-reverse-tree}
  \( \inflationfactor \leftarrow \updateinflationfactor \)\;\label{alg:eitstar:technical:initialize-inflation-factor}
  \( \cdresolution \leftarrow \updatecdresolution \)\;\label{alg:eitstar:technical:initialize-cd-resolution}
  \( \fwdqueue \leftarrow \expandstate{\startstate} \)\texttt{;}
  \( \revqueue \leftarrow \expandstate{\goalstates} \)\;\label{alg:eitstar:initialize-queues:expand-start-and-goals}
  \Repeat{\normalfont\tt stopped}{\label{alg:eitstar:technical:repeat-begin}
    \uIf{\normalfont \hspace*{-1em}\( \displaystyle\min_{\edge{\sourcestate}{\targetstate} \in \revqueue}
      \hspace*{-0.1em}\set*{ \eitrevkey{\sourcestate}{\targetstate} } < \hspace*{-1.0em}\min_{\edge{\sourcestate}{\targetstate} \in \fwdqueue}
      \hspace*{-0.1em}\set*{ \eitfwdkey{\sourcestate}{\targetstate} }
      \)\label{alg:eitstar:technical:continue-reverse-search-1}\\
    \textbf{or} target of best edge in forward queue is not closed}{\label{alg:eitstar:technical:continue-reverse-search-2}
      \( \edge{\sourcestate}{\targetstate} \leftarrow \mathop{\arg\min}_{\edge{\sourcestate}{\targetstate} \in \revqueue}
      \set*{ \eitrevkey{\sourcestate}{\targetstate} } \)\;\label{alg:eitstar:iterate-reverse-search:get-best-rev-edge}
      \( \revqueue \setsubtract \left( \sourcestate, \targetstate \right)
      \)\;\label{alg:eitstar:iterate-reverse-search:remove-best-rev-edge}
      \( \closedrevvertices \setadd \sourcestate \)\;\label{alg:eitstar:iterate-reverse-search:close-source-state}
      \eIf{\normalfont\( \couldbevalid{\left(\sourcestate, \targetstate\right), d}
        \)}{\label{alg:eitstar:iterate-reverse-search:could-be-valid}
        \( \inadrevctc{\targetstate} \leftarrow
        \min\set*{\inadrevctc{\targetstate}, \inadrevctc{\sourcestate} +
          \inadedgecost{\targetstate}{\sourcestate}} \)\;\label{alg:eitstar:iterate-reverse-search:update-inad-cost}
        \( \inadrevetc{\targetstate} \leftarrow \min\set*{\inadrevetc{\targetstate}, \inadrevetc{\sourcestate} +
          \inadedgeeffort{\targetstate}{\sourcestate}} \)\;\label{alg:eitstar:iterate-reverse-search:update-inad-effort}
        \If{\normalfont\( \adctglabel{\targetstate}  > \adctglabel{\sourcestate} + \adedgecost{\targetstate}{\sourcestate} \)}{\label{alg:eitstar:iterate-reverse-search:admissible-cost-test}
          \( \adctglabel{\targetstate} \leftarrow \adctglabel{\sourcestate} + \adedgecost{\targetstate}{\sourcestate} \)\label{alg:eitstar:iterate-reverse-search:admissible-cost-update}\;
          \eIf{\normalfont\( \targetstate \in \revvertices \)}{\label{alg:eitstar:iterate-reverse-search:rewiring-check}
            \( \revedges \setsubtract \left( \revparent{\targetstate}, \targetstate \right) \)\;\label{alg:eitstar:iterate-reverse-search:remove-parent-from-tree}
          }{
            \( \revvertices \setadd \targetstate \)\;\label{alg:eitstar:iterate-reverse-search:insert-state-in-tree}
          }
          \( \revedges \setadd \left( \sourcestate, \targetstate \right) \)\;\label{alg:eitstar:iterate-reverse-search:insert-edge-in-tree}
          \( \revqueue \setadd \expandstate{\targetstate} \) \;\label{alg:eitstar:iterate-reverse-search:expand-target-state}
        }
      }{
        \( \invedges \setadd \set*{ \edge{\sourcestate}{\targetstate},
            \edge{\targetstate}{\sourcestate} } \)\label{alg:eitstar:iterate-reverse-search:remember-invalid-edge}
      }
    }
    \uElseIf{\normalfont \( \displaystyle\hspace*{-1.0em} \min_{\edge{\sourcestate}{\targetstate} \in \fwdqueue} \hspace*{-0.2em}\set*{ \fwdctc{\sourcestate} +
        \adedgecost{\sourcestate}{\targetstate} + \concost{\targetstate} } < \currentcost \)}{\label{alg:eitstar:technical:continue-forward-search}
      \( \edge{\sourcestate}{\targetstate} \leftarrow \getbestfwdedge{\fwdqueue} \)\;\label{alg:eitstar:iterate-forward-search:get-best-edge}
      \( \fwdqueue \setsubtract \left( \sourcestate, \targetstate \right) \)\;\label{alg:eitstar:iterate-forward-search:pop-best-edge}
      \uIf{\( (\sourcestate, \targetstate) \in \fwdedges \)}{\label{alg:eitstar:iterate-forward-search:is-edge-in-tree}
        \( \fwdqueue \setadd \expandstate{\targetstate}\)\;\label{alg:eitstar:iterate-forward-search:expand-edge-in-tree}
      }
      \ElseIf{\( \fwdctc{\sourcestate} + \adedgecost{\sourcestate}{\targetstate} < \fwdctc{\targetstate} \)}{\label{alg:eitstar:iterate-forward-search:can-edge-possibly-improve-tree}
        \uIf{\normalfont\(\isvalid{\sourcestate, \targetstate}\)}{\label{alg:eitstar:iterate-forward-search:collision-detection}
          \If{\( \fwdctc{\sourcestate} + \edgecost{\sourcestate}{\targetstate} + \hat{h}_{\mathrm{con}}\left[\targetstate\right] < \currentcost \)}{\label{alg:eitstar:iterate-forward-search:can-edge-actually-improve-solution}
            \If{\( \fwdctc{\sourcestate} + \edgecost{\sourcestate}{\targetstate} < \fwdctc{\targetstate} \)}{\label{alg:eitstar:iterate-forward-search:can-edge-actually-improve-tree}
              \eIf{\( \targetstate \not\in \fwdvertices \)}{\label{alg:eitstar:iterate-forward-search:forward-tree-check}
                \( \fwdvertices \setadd \targetstate \)\;\label{alg:eitstar:iterate-forward-search:add-state-to-tree}
              }{\label{alg:eitstar:iterate-forward-search:rewiring-else}
                \( \fwdedges \setsubtract (\fwdparent{\targetstate}, \targetstate) \)\;\label{alg:eitstar:iterate-forward-search:rewiring}
              }
              \( \fwdedges \setadd (\sourcestate, \targetstate) \)\;\label{alg:eitstar:iterate-forward-search:add-edge-to-tree}
              \( \fwdqueue \setadd \expandstate{\targetstate} \)\;\label{alg:eitstar:iterate-forward-search:expand-child-state}
              \If{\( \min_{\goalstate \in \goalstates}\set*{
                  \fwdctc{\goalstate} } < \currentcost \)}{
                \( \currentcost \leftarrow \min_{\goalstate \in \goalstates}\set*{
                  \fwdctc{\goalstate} }
                \)\;\label{alg:eitstar:iterate-forward-search:update-solution-cost}
                \( \inflationfactor \leftarrow \updateinflationfactor \)\;\label{alg:eitstar:iterate-forward-search:update-inflation-factor}
              }
            }
          }
        }
        \Else{
          \( \invedges \setadd \set*{ \edge{\sourcestate}{\targetstate},
            \edge{\targetstate}{\sourcestate} } \)\;\label{alg:eitstar:iterate-forward-search:blacklist}
          \If{\normalfont \( \edge{\sourcestate}{\targetstate} \in \revedges \)}{\label{alg:eitstar:iterate-forward-search:reverse-tree-check}
            \( \cdresolution \leftarrow \updatecdresolution{}
            \)\;\label{alg:eitstar:iterate-forward-search:update-cd-resolution}
            \( \revvertices \leftarrow \goalstates; \revedges \leftarrow \emptyset
            \)\;
            \( \revqueue \leftarrow \expandstate{\goalstates} \)\;\label{alg:eitstar:iterate-forward-search:reexpand-goal-states}
          }
        }
      }
    }
    \Else{
      \( \prunestates{\sampledstates} \)\;\label{alg:eitstar:improve-approximation:prune}
      \( \sampledstates \setadd \samplestates{m, \currentcost} \)\;\label{alg:eitstar:improve-approximation:sample}
      \( \revvertices \leftarrow \goalstates \)\texttt{;} \(  \closedrevvertices \leftarrow
      \emptyset \)\texttt{;} \(
      \revedges \leftarrow \emptyset \)\;\label{alg:eitstar:technical:reinitialize-reverse-tree}
      \( \fwdqueue \leftarrow \expandstate{\startstate} \)\texttt{;}
      \( \revqueue \leftarrow \expandstate{\goalstates}
      \)\;\label{alg:eitstar:reinitialize-queues:expand-start-and-goals}
      \( \cdresolution \leftarrow \updatecdresolution{}
      \)\;\label{alg:eitstar:improve-approximation:reinitialize-cd-resolution}
    }
  }\label{alg:eitstar:technical:repeat-end}
\end{algorithm}%

%%% Local Variables:
%%% mode: latex
%%% TeX-master: "../../../../main"
%%% End:

\begin{algorithm}[t]
  \caption{\acs{EIT*}: \( \getbestfwdedge{\fwdqueue} \)}%
  \label{alg:eitstar:get-best-forward-edge}
  \small
  \( \displaystyle \edge{\inadremeffortsrc}{\inadremefforttgt} \leftarrow
  \mathop{\arg\,\min}_{\edge{\sourcestate}{\targetstate} \in \queue[\fwdsymbol][\inflationfactor\inadsolcostlabel*{}{}]} \set*{\inadedgeeffort{\sourcestate}{\targetstate} + \inadefforttogolabel{\targetstate}} \)\;
  \( \displaystyle \edge{\inadsolcostsrc}{\inadsolcosttgt} \leftarrow
  \mathop{\arg\,\min}_{\edge{\sourcestate}{\targetstate} \in \fwdqueue} \set*{\fwdctc{\sourcestate} + \inadedgecost{\sourcestate}{\targetstate} + \inadctglabel{\targetstate}} \)\;
  \( \displaystyle \edge{\adsolcostsrc}{\adsolcosttgt} \leftarrow
  \mathop{\arg\,\min}_{\edge{\sourcestate}{\targetstate} \in \fwdqueue} \set*{\fwdctc{\sourcestate} + \adedgecost{\sourcestate}{\targetstate} + \adctglabel{\targetstate}} \)\;

  \uIf{\normalfont \( \inadsolcostlabel{\inadremeffortsrc}{\inadremefforttgt} \leq \inflationfactor \adsolcostlabel{\adsolcostsrc}{\adsolcosttgt} \)}{\label{alg:eitstar:get-best-forward-edge:best-effort-test}
    \Return \( \edge{\inadremeffortsrc}{\inadremefforttgt} \) \;\label{alg:eitstar:get-best-forward-edge:best-effort-return}
  }
  \uElseIf{\normalfont \( \inadsolcostlabel{\inadsolcostsrc}{\inadsolcosttgt} \leq \inflationfactor \adsolcostlabel{\adsolcostsrc}{\adsolcosttgt} \)}{\label{alg:eitstar:get-best-forward-edge:best-cost-test}
    \Return \( \edge{\inadsolcostsrc}{\inadsolcosttgt} \) \;\label{alg:eitstar:get-best-forward-edge:best-cost-return}
  }
  \Else{\label{alg:eitstar:get-best-forward-edge:lower-bound-cost-else}
    \Return \( \edge{\adsolcostsrc}{\adsolcosttgt} \) \;\label{alg:eitstar:get-best-forward-edge:lower-bound-cost-return}
  }
\end{algorithm}%

%%% Local Variables:
%%% mode: latex
%%% TeX-master: "../../../main"
%%% End:

\paragraph{\small Minimum effort estimate}

An estimate of the minimum remaining effort to validate a solution within the
suboptimality bound is computed as
\begin{align*}
  \mathop{\min}\limits_{(\bm{\mathrm{x}}_{\mathrm{s}},
  \bm{\mathrm{x}}_{\mathrm{t}}) \in \mathcal{Q}_{\mathcal{F}}^{w\bar{s}}}
  \bar{r}\left( \bm{\mathrm{x}}_{\mathrm{s}}, \bm{\mathrm{x}}_{\mathrm{t}} \right),
\end{align*}
where
\( \bar{r} \colon X_{\mathrm{sampled}} \times X_{\mathrm{sampled}} \to [0,
\infty) \) estimates the remaining effort to validate a solution through an
edge as,
\begin{align*}
  \bar{r}\left( \bm{\mathrm{x}}_{\mathrm{s}}, \bm{\mathrm{x}}_{\mathrm{t}}
  \right) \coloneqq \bar{e}\left( \bm{\mathrm{x}}_{\mathrm{s}}, \bm{\mathrm{x}}_{\mathrm{t}} \right) + \bar{e}\left[ \bm{\mathrm{x}}_{\mathrm{t}} \right],
\end{align*}
where
\( \bar{e}\left( \bm{\mathrm{x}}_{\mathrm{s}}, \bm{\mathrm{x}}_{\mathrm{t}}
\right) \) is the heuristic effort to validate the edge and
\( \bar{e}\left[ \bm{\mathrm{x}}_{\mathrm{t}} \right] \) is the calculated
heuristic effort to validate a solution from the target of the edge. The
minimum is taken only over the edges in the queue that are estimated to lead to
a solution within the current suboptimality factor, \( w \),
\begin{align*}
  \mathcal{Q}_{\mathcal{F}}^{w\bar{s}} \coloneqq \set*{(\bm{\mathrm{x}}_{\mathrm{s}},  \bm{\mathrm{x}}_{\mathrm{t}}) \in \mathcal{Q}_{\mathcal{F}} \,\given\, \bar{s}(\bm{\mathrm{x}}_{\mathrm{s}}, \bm{\mathrm{x}}_{\mathrm{t}}) \leq w \bar{s}\left( \bm{\mathrm{x}}_{\mathrm{s}}^{\bar{s}}, \bm{\mathrm{x}}_{\mathrm{t}}^{\bar{s}} \right)}.
\end{align*}
The edge that results in this estimate of the minimum required effort remaining
to find a solution within the current suboptimality bound is denoted as
\begin{align*}
  (\bm{\mathrm{x}}_{\mathrm{s}}^{\bar{r}},
  \bm{\mathrm{x}}_{\mathrm{t}}^{\bar{r}}) \coloneqq
  \mathop{\arg\min}\limits_{(\bm{\mathrm{x}}_{\mathrm{s}},
  \bm{\mathrm{x}}_{\mathrm{t}}) \in \mathcal{Q}_{\mathcal{F}}^{w\bar{s}}}
  \bar{r}\left( \bm{\mathrm{x}}_{\mathrm{s}}, \bm{\mathrm{x}}_{\mathrm{t}} \right).
\end{align*}

\vspace{0.5em}

\ac{EIT*} first checks if the queue contains an edge that could improve the current solution (Alg.~\ref{alg:eitstar:technical}, line~\ref{alg:eitstar:technical:continue-forward-search}). If none of the edges in the forward queue can, then the \ac{RGG} approximation is improved by pruning states that are not in the informed set and sampling more states (Alg.~\ref{alg:aitstar:prune} and Alg.~\ref{alg:eitstar:technical}, lines~\ref{alg:eitstar:improve-approximation:prune} and~\ref{alg:eitstar:improve-approximation:sample}). The reverse search tree and set of closed vertices are then reset (Alg.~\ref{alg:eitstar:technical}, line~\ref{alg:eitstar:technical:reinitialize-reverse-tree}) and the forward and reverse search queues are reinitialized by inserting the outgoing edges of the start and goals, respectively~(Alg.~\ref{alg:eitstar:technical}, line~\ref{alg:eitstar:reinitialize-queues:expand-start-and-goals}).

If at least one edge in the forward queue could improve the current solution, then the edge that is estimated to lead to the fastest improvement of the current solution is extracted from the queue (Alg.~\ref{alg:eitstar:technical}, lines~\ref{alg:eitstar:iterate-forward-search:get-best-edge} and~\ref{alg:eitstar:iterate-forward-search:pop-best-edge}, and Alg.~\ref{alg:eitstar:get-best-forward-edge}). This edge is determined with the following steps:
\begin{enumerate}
\item If the edge with the minimum remaining effort required to validate a
  solution,
  \( \left( \bm{\mathrm{x}}_{\mathrm{s}}^{\bar{r}},
    \bm{\mathrm{x}}_{\mathrm{t}}^{\bar{r}} \right) \), can possibly lead to a
  solution within the current suboptimality bound,
  \begin{align*}
    \bar{s}\left( \bm{\mathrm{x}}_{\mathrm{s}}^{\bar{r}}, \bm{\mathrm{x}}_{\mathrm{t}}^{\bar{r}} \right) \leq w \hat{s}\left( \bm{\mathrm{x}}_{\mathrm{s}}^{\hat{s}}, \bm{\mathrm{x}}_{\mathrm{t}}^{\hat{s}} \right),
  \end{align*}
  then it is selected (Alg.~\ref{alg:eitstar:get-best-forward-edge}, lines~\ref{alg:eitstar:get-best-forward-edge:best-effort-test} and~\ref{alg:eitstar:get-best-forward-edge:best-effort-return}). This edge likely improves the current solution with the least amount of computational effort.
\item If the edge that is estimated to be on the optimal solution path,
  \( \left( \bm{\mathrm{x}}_{\mathrm{s}}^{\bar{s}},
    \bm{\mathrm{x}}_{\mathrm{t}}^{\bar{s}} \right) \), can possibly lead to a
  solution within the suboptimality bound,
  \begin{align*}
    \bar{s}\left( \bm{\mathrm{x}}_{\mathrm{s}}^{\bar{s}}, \bm{\mathrm{x}}_{\mathrm{t}}^{\bar{s}} \right) \leq w \hat{s}\left( \bm{\mathrm{x}}_{\mathrm{s}}^{\hat{s}}, \bm{\mathrm{x}}_{\mathrm{t}}^{\hat{s}} \right),
  \end{align*}
  then it is selected (Alg.~\ref{alg:eitstar:get-best-forward-edge},
  lines~\ref{alg:eitstar:get-best-forward-edge:best-cost-test}
  and~\ref{alg:eitstar:get-best-forward-edge:best-cost-return}). This edge likely steps
  towards the resolution-optimal solution.
\item Otherwise the edge that provides the lower bound on the optimal solution
  cost in the current \ac{RGG} approximation,
  \( \left( \bm{\mathrm{x}}_{\mathrm{s}}^{\hat{s}},
    \bm{\mathrm{x}}_{\mathrm{t}}^{\hat{s}} \right) \), is selected. This raises
  the lower bound on the optimal solution cost in the current \ac{RGG}
  approximation and increases the number of candidates available to steps 1 and
  2 in the next iteration~(Alg.~\ref{alg:eitstar:get-best-forward-edge},
  lines~\ref{alg:eitstar:get-best-forward-edge:lower-bound-cost-else} and~\ref{alg:eitstar:get-best-forward-edge:lower-bound-cost-return}).
\end{enumerate}

The forward search in \ac{EIT*} then proceeds similarly to \ac{AIT*}. If the
selected edge is in the forward search tree, then its target state is expanded
and the forward search iteration is complete~(Alg.~\ref{alg:eitstar:technical},
lines~\ref{alg:eitstar:iterate-forward-search:is-edge-in-tree}
and~\ref{alg:eitstar:iterate-forward-search:expand-edge-in-tree}). If the selected edge is
not part of the forward search tree but can possibly improve it, then it is
checked for collisions~(Alg.~\ref{alg:eitstar:technical},
lines~\ref{alg:eitstar:iterate-forward-search:can-edge-possibly-improve-tree}
and~\ref{alg:eitstar:iterate-forward-search:collision-detection}).

If collisions are detected, then the edge is added to the invalid edges (Alg.~\ref{alg:eitstar:technical}, line~\ref{alg:eitstar:iterate-forward-search:blacklist}) and if it is in the reverse search tree, then the reverse search tree and queue are reset and the sparse collision detection resolution is updated, which will improve the accuracy of the heuristic computed by restarting the reverse search (Alg.~\ref{alg:eitstar:technical}, lines~\ref{alg:eitstar:iterate-forward-search:reverse-tree-check}--\ref{alg:eitstar:iterate-forward-search:reexpand-goal-states}). If no collisions are detected, then the true cost of the edge is evaluated to check whether it actually improves the current solution and forward search tree (Alg.~\ref{alg:eitstar:technical}, lines~\ref{alg:eitstar:iterate-forward-search:can-edge-actually-improve-solution} and~\ref{alg:eitstar:iterate-forward-search:can-edge-actually-improve-tree}).

If the edge improves the current solution and forward search tree, then its
target state is added to the tree if it is not already in it
(Alg.~\ref{alg:eitstar:technical},
lines~\ref{alg:eitstar:iterate-forward-search:forward-tree-check}
and~\ref{alg:eitstar:iterate-forward-search:add-state-to-tree}). If the target
state is already in the tree, then the edge causes a rewiring and the edge from
the old parent is removed from the tree (Alg.~\ref{alg:eitstar:technical},
lines~\ref{alg:eitstar:iterate-forward-search:rewiring-else}
and~\ref{alg:eitstar:iterate-forward-search:rewiring}). The new edge is then
added to the tree, its target state is expanded into the forward queue, and the
solution cost is updated (Alg.~\ref{alg:eitstar:technical},
lines~\ref{alg:eitstar:iterate-forward-search:add-edge-to-tree}--\ref{alg:eitstar:iterate-forward-search:update-solution-cost}). If
the edge results in an improved solution, then the suboptimality factor is
changed according to a user-specified update policy
(Alg.~\ref{alg:eitstar:technical},
lines~\ref{alg:eitstar:iterate-forward-search:update-inflation-factor}). Section~\ref{sec:experimental-results}
presents the update policy used in the experimental evaluation of \ac{EIT*}.

The entire forward search terminates when it is guaranteed that the optimal
solution in the current \ac{RGG} approximation is found.
This occurs when no edge in the forward queue can possibly improve the current
solution~(Alg.~\ref{alg:eitstar:technical},~\lnereft{alg:eitstar:technical:continue-forward-search}).
The forward search also terminates when the start and goal are not in the same
connected component of the \ac{RGG} approximation. This occurs when the reverse
search tree does not reach any edge in the forward queue, but this
condition is omitted from Algorithm~\ref{alg:eitstar:technical} for clearer
structure.

Similar to \ac{AIT*}, the three steps of \ac{EIT*}, i.e., improving the \ac{RGG} approximation, updating the heuristic with the reverse search, and finding valid paths with the forward search, are repeated for as long as computational time allows or until a suitable solution is found. This results in increasingly accurate cost and effort heuristics for increasingly efficient and effective searches of increasingly accurate approximations and will also almost-surely asymptotically converge to the optimal solution in the limit of infinite samples~\refp[Section]{sec:eit-analysis}.

\subsection{Analysis}%
\label{sec:analysis}

Any path planning algorithm that processes a sampling-based approximation with
a graph-search algorithm is almost-surely asymptotically optimal if the
approximation almost-surely contains an asymptotically optimal solution and the
graph-search algorithm is resolution-optimal. This is a sufficient but not
necessary condition. The almost-sure asymptotic optimality of \ac{AIT*} and
\ac{EIT*} follows from proven properties of their \ac{RGG} approximations and
graph-search algorithms.

\subsubsection{\acs{AIT*}}%
\label{sec:ait-analysis}

The \ac{RGG} approximation constructed by \ac{AIT*} almost-surely contains an
asymptotically optimal solution because it contains all the edges in \acs{PRM*}
for any set of samples and \acs{PRM*} is almost-surely asymptotically
optimal~\citep{karaman_ijrr2011}. \ac{AIT*}'s forward search is resolution-optimal
because A* is a resolution-optimal algorithm if it is provided with an
admissible cost heuristic~\citep{hart_tssc1968}. \ac{AIT*}'s reverse search
without collision detection results in an admissible cost-heuristic because
\ac{LPA*} is also a resolution-optimal algorithm~\citep{aine_ai2016} and because
adding collision detection cannot decrease path cost. \ac{AIT*} is therefore
almost-surely asymptotically optimal.

\subsubsection{\acs{EIT*}}%
\label{sec:eit-analysis}

The \ac{RGG} approximation constructed by \ac{EIT*} almost-surely contains an
asymptotically optimal solution because like \ac{AIT*} it also contains all the
edges in \acs{PRM*} and \acs{PRM*} is almost-surely asymptotically
optimal~\citep{karaman_ijrr2011}. EIT*'s forward search is resolution-optimal
because \ac{AEES} is a resolution-optimal algorithm when the cost heuristic is
admissible and the suboptimality factor is one~\citep{thayer_icaps2011a}. EIT*'s
reverse search with sparse collision detection results in an admissible cost
heuristic because denser collision detection cannot decrease path cost and A*
is a resolution-optimal graph-search algorithm when provided with an admissible
cost heuristic~\citep{hart_tssc1968}. \ac{EIT*} is therefore also almost-surely
asymptotically optimal.

%%% Local Variables:
%%% mode: latex
%%% TeX-master: "../main"
%%% End:

% Experimental Results
\section{Experimental Results}%
\label{sec:experimental-results}

The benefits of an asymmetric bidirectional search are shown on abstract,
robotic manipulator, and knee replacement dislocation problems
(Sections~\ref{sec:abstract-problems}--\ref{sec:knee-implants}). \ac{AIT*} and
\ac{EIT*} were compared against the \accite{OMPL}{sucan_ram2012} implementations of
RRT, RRT-Connect, RRT*, LBT-RRT, LazyPRM*, FMT*, BIT*, and
ABIT*\footnote[1]{\( {}^{1}\)Using \ac{OMPL} v1.5.0 on a laptop with 16 GB of
  RAM and an Intel i7-4910MQ (2.9 GHz) processor running Ubuntu 18.04}.

The planners were tested when optimizing path length and obstacle
clearance. Path length was optimized by minimizing the arc length of the path
in state space. Obstacle clearance was optimized by minimizing the reciprocal
of clearance integrated over the arc length of the path, \( l \),
\begin{align*}
  c(\sigma) &\coloneqq \int_{0}^{l} \frac{ 1 }{ \delta\left( \sigma(\nicefrac{s}{l}) \right) } \,\mathrm{d}s,
\end{align*}
where \( \delta\colon X \to [10^{-6}, \infty) \) is the distance of a state to the nearest obstacle, limited to be no smaller than \( 10^{-6} \),
\begin{align*}
  \delta\left( \bm{\mathrm{x}} \right) \coloneqq \max \left\{
  \mathop{\mathrm{clearance}}(\bm{\mathrm{x}}), 10^{-6} \right\}.
\end{align*}
The lower limit on \( \delta \) ensures numerical stability and that the cost
of a path is bounded by a multiple of its total variation as in
\reft[Section]{sec:cost-function-assumptions}.
This optimization objective balances the clearance and length of a path and is similar to the objectives presented by \citet{wein_ijrr2008} and \citet{agarwal_ta2018}.

The admissible cost heuristic, \( \hat{c} \), used by informed planners was the Euclidean distance for path length and the trivial zero-heuristic for obstacle clearance. The possibly inadmissible cost heuristic, \( \bar{c} \), used by \ac{EIT*} was again the Euclidean distance for path length and the reciprocal of the average clearance of the two end states for obstacle clearance,
\begin{align*}
  \bar{c}\left( \bm{\mathrm{x}}_{i}, \bm{\mathrm{x}}_{j} \right) \coloneqq
  \frac{2}{\delta\left( \bm{\mathrm{x}}_{i} \right) + \delta\left(
  \bm{\mathrm{x}}_{j} \right) }.
\end{align*}
The effort heuristic, \( \bar{e}(\,\cdot\,,\,\cdot\,) \), used by \ac{EIT*} was the number of collision checks required to validate a path for both objectives. It was computed by dividing the Euclidean distance between two states by the collision detection resolution.

The inflation factor update policy in \ac{EIT*} was configured to have an infinite inflation factor until the initial solution is found and then switch to a unity inflation factor. This results in fast initial solutions and efficient subsequent searches to improve them.
The sparse collision detection resolution update policy used in \ac{EIT*} was configured to initially search each batch with a single collision check and then double the resolution if the forward search detects a collision on an edge used in the reverse search tree.

RRT-based planners used maximum edge lengths of 0.3, 0.9, 1.25, 1.25, 2.4, and 3.0 in \( \mathbb{R}^{2}, \mathrm{SE}(3), \mathbb{R}^{7}, \mathbb{R}^{8}, \mathbb{R}^{14} \), and \( \mathbb{R}^{16} \), respectively. RRT* used informed sampling and \ac{LBT-RRT} used the default approximation factor of 0.4 but was not tested on obstacle clearance problems as it can only optimize path length.
BIT*-based planners used a batch size of \( 100 \) samples and the \( k \)-nearest connection strategy with an \ac{RGG} connection parameter of \( \eta = 1.001 \) regardless of problem dimension.

FMT* is not an anytime algorithm and requires the user to specify the number of
samples in advance. All experiments presented in this section tested
configurations of \ac{FMT*} with 10, 50, 100, 500, 1000, and 5000
samples. There are multiple lines for \ac{FMT*} in the presented plots because
median solution times and costs were computed separately for each configuration
and the results of all configurations that were able to solve a specific
problem were plotted.

\subsection{Abstract Problem}%
\label{sec:abstract-problems}

State space obstacles have complex shapes even for relatively simple
problems~\citep[e.g., Figure 1, ][]{das_tro2020}. This complexity often makes it
difficult to gain insight about the underlying reason for a planner's
performance on a given problem. Directly designing abstract state space
obstacles from basic geometries provides intuition on the performance of a
planner for a given obstacle configuration and helps the algorithmic design
process.

The basic geometries of these simple obstacle configurations make collision
detection computationally much less expensive than in real-world problems. A
simple way to simulate the more expensive collision detection of real-world
problems in this abstract setting is to increase the collision detection
resolution. The collision detection resolution was set to \( 5\cdot10^{-6} \),
which on the tested hardware makes evaluating a valid
edge in this abstract setting as computationally expensive as evaluating a
valid edge on a dual-arm manipulation problem~\refp[Section]{sec:manipulator-arms}. While admissible cost heuristics
exist for these abstract problems with clearance in state
space~\citep{strub_tr2021}, such heuristics often do not exist for real-world
problems with clearance in work space and therefore no heuristics were used for
the clearance objective in these abstract problems either.

The abstract obstacle configuration on which the planners were tested consists
of a wall with a narrow gap between the start and goal states
\refp[Figure]{fig:results:wall-gap}. This obstacle configuration illustrates
the speed with which planners find a hard-to-find optimal homotopy class when
optimizing path length. When optimizing obstacle clearance, this configuration
illustrates the challenges of searching in the absence of informative
heuristics and ordering the search according to the total potential solution
cost.

Three versions of the wall gap obstacle configuration in dimensions
\( \mathbb{R}^{2}, \mathbb{R}^{8} \), and \( \mathbb{R}^{16} \) were tested for
both objectives. The obstacle configuration shown in
\reft[Figure]{fig:results:wall-gap} was adapted to higher dimensions by extending
the obstacle such that only two homotopy classes
exist for all problems.

\reft[Figure]{fig:results:wall-gap-results} shows the performance of all
algorithms on all six instances of the problem when optimizing path length and
obstacle clearance. When optimizing path length, \ac{AIT*} and \ac{EIT*}
perform similarly to Lazy \acs{PRM*}, \ac{BIT*}, and \ac{ABIT*} and find
initial solutions at least as fast as \ac{RRT}-Connect and significantly faster
than \acs{RRT*} and \ac{LBT-RRT}~(Figures~\ref{fig:results:wall-gap-path-length-r2},
\subref{fig:results:wall-gap-path-length-r8}, and
\subref{fig:results:wall-gap-path-length-r16}). When optimizing obstacle
clearance, \ac{EIT*} outperforms all other tested asymptotically optimal
planners, including \ac{AIT*}, by again finding initial solutions as fast as
\ac{RRT}-Connect, which has a computational advantage because it does not
calculate path cost (e.g., obstacle clearance,
Figures~\ref{fig:results:wall-gap-clearance-r2},
\subref{fig:results:wall-gap-clearance-r8}, and
\subref{fig:results:wall-gap-clearance-r16}).

\subsection{Manipulator Problems}%
\label{sec:manipulator-arms}

The algorithms were also tested on path planning problems for Barrett \ac{WAM}
arms in the \accite{OpenRAVE}{diankov_phd2010}. \ac{OpenRAVE} was configured to use
the \accite{FCL}{pan_icra2012} for collision detection and clearance computation,
using \accite{OBB}{gottschalk_ccgit1996} tree and \accite{RSS}{larsen_icra2000}
volume representations, respectively. The collision detection and clearance
computation resolution was set to \( 3.6 \cdot 10^{-3} \), which resulted in a
1\% false-negative collision detection rate for invalid edges on representative
problems.

\subsubsection{Single-Arm Manipulator Problem}%
\label{sec:single-arm-planning}

Robotic manipulator arms are commonly used in pick-and-place tasks. In the
single-arm experiment, the algorithms were instructed to find paths for a
Barret \ac{WAM} arm with seven degrees of freedom to place a small cube into a
box \refp[Figure]{fig:results:one-manipulator-arm}.

\reft[Figure]{fig:results:one-arm-results} shows the performance of all
algorithms when optimizing path length and obstacle clearance. When optimizing
path length, \ac{AIT*} and \ac{EIT*} perform similarly to Lazy \acs{PRM*},
\ac{BIT*}, and \ac{ABIT*}, which all find initial solutions nearly as fast as
\ac{RRT}-Connect and significantly faster than \ac{RRT*} and \ac{LBT-RRT}
\refp[Figure]{fig:results:one-arm-path-length}. When optimizing obstacle
clearance, \ac{AIT*} and \ac{EIT*} outperform all other tested asymptotically
optimal planners but do not find initial solutions as fast as \ac{RRT}-Connect,
which again has a computational advantage because it does not calculate path
cost \refp[Figure]{fig:results:one-arm-clearance}.

\subsubsection{Dual-Arm Manipulator Problem}%
\label{sec:dual-arm-planning}

In the dual-arm planning experiment, the algorithms were instructed to find
paths for two Barret \ac{WAM} arms with a total of 14 degrees of freedom from a
start configuration at the bottom shelf to a goal configuration at the top
shelf \refp[Figure]{fig:results:two-manipulator-arms}.

\reft[Figure]{fig:results:two-arm-results} shows the performance of all
algorithms when optimizing path length and obstacle clearance. When optimizing
path length, \ac{AIT*} performs similarly to \ac{BIT*} and \ac{ABIT*} which
perform better than Lazy \acs{PRM*} and \ac{EIT*} and find initial solutions
nearly as fast as \ac{RRT}-Connect and significantly faster than \ac{RRT*} and
\ac{LBT-RRT} \refp[Figure]{fig:results:two-arm-path-length}. When optimizing
obstacle clearance, \ac{AIT*} and \ac{EIT*} again outperform all tested
asymptotically optimal planners but do not find initial solutions as fast as
\ac{RRT}-Connect, which still has a computational advantage because it does not
calculate path cost \refp[Figure]{fig:results:two-arm-clearance}.

\subsection{Knee Replacement Dislocation Problem}%
\label{sec:knee-implants}

Calculating heuristics with an asymmetric bidirectional search can also improve performance on the feasible planning problem by guiding the search towards the goal. The knee replacement dislocation problem evaluates the potential of medial dislocation for the Oxford Domed Lateral \acl{UKR}~\citep[\acs{UKR};][Figure~\ref{fig:results:knee}]{pandit_knee2010}\footnote[2]{\({}^{2}\)This experiment used an approximation of the Oxford Domed Lateral \acl{UKR} due to copyright restrictions.}\acused{UKR} by searching for a path to free the mobile bearing.

The Oxford Domed Lateral \ac{UKR} consists of metal femoral and ti\-bial components which are fixed to the bone and a mobile polythylene bearing which separates the metal components~\citep{gunther_knee1996}. Medial dislocation occurs when there is enough space between the tibial and femoral components for the mobile bearing to move onto the tibial-component wall where it may be trapped by the femoral component. The dislocation risk for different relative poses of the femoral and tibial components has been analyzed by using planning algorithms to search for paths that allow the bearing to reach a region representative of dislocation~\citep{yang_caos2020, yang_bors2021}. \reft[Figures]{fig:results:knee-approximation} and~\subref{fig:results:knee-goal-region} respectively illustrate the start state and goal region of the mobile bearing and the fixed poses of the tibial and femoral components used in this experiment.
The state space of this problem is \( \mathrm{SE}(3) \), and the mobile bearing is free to move and rotate in any direction not in collision with the fixed parts.

\reft[Figure]{fig:results:knee-results} shows the performance of all planners when optimizing path length and obstacle clearance. When optimizing path length, \ac{EIT*} outperforms all other tested planners. \ac{AIT*} is the second best performing algorithm followed by \ac{FMT*}, \ac{BIT*}, and \ac{ABIT*}. The \ac{OMPL} implementations of \ac{LBT-RRT} and \ac{RRT}-Connect did not allow goals to be defined as a region and were replaced with \ac{RRT} for this experiment. When optimizing obstacle clearance, \ac{EIT*} again outperforms all other tested planners and is the only planner that achieved a success rate of 100\%.

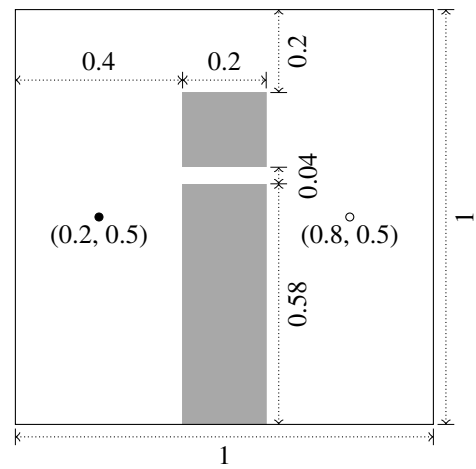
\begin{figure}[t]
  \centering
  \begin{tikzpicture}[scale = 5.5] % scale = 7.7 for ~columnwidth
% Obstacle
\draw [obstacle] (-0.1, -0.5) rectangle (0.1, 0.3);

% Boundary
\draw [boundary] (-0.5, -0.5) rectangle (0.5, 0.5);

% Antiobstacle
\draw [fill = white, draw = none] (-0.11, 0.08) rectangle (0.11, 0.12);

% Start state
\node (wallgapstart) [start] at (-0.3, 0.0) {};
\node [below = 0pt of wallgapstart, inner sep = 1pt] {\normalfont (0.2, 0.5)};

% Goal state
\node (wallgapgoal) [goal] at (0.3, 0.0) {};
\node [below = 0pt of wallgapgoal, inner sep = 1pt] {\normalfont (0.8, 0.5)};

% Measurements
\draw [<->, densely dotted] (0.13, -0.5) -- node [below, sloped, midway] {\normalfont 0.58} (0.13, 0.08);
\draw [ultra thin] (0.11, 0.08) -- (0.15, 0.08);
\draw [<->, densely dotted] (0.13, 0.08) -- node [below=4pt, sloped, midway] {\normalfont 0.04} (0.13, 0.12);
\draw [ultra thin] (0.11, 0.12) -- (0.15, 0.12);
\draw [<->, densely dotted] (0.13, 0.3) -- node [below, sloped, midway] {\normalfont 0.2} (0.13, 0.5);
\draw [ultra thin] (0.11, 0.3) -- (0.15, 0.3);
\draw [<->, densely dotted] (-0.1, 0.33) -- node [above, midway] {\normalfont 0.2} (0.1, 0.33);
\draw [ultra thin] (-0.1, 0.31) -- (-0.1, 0.35);
\draw [<->, densely dotted] (-0.5, 0.33) -- node [above, midway] {\normalfont 0.4} (-0.1, 0.33);
\draw [ultra thin] (0.1, 0.31) -- (0.1, 0.35);
\draw [<->, densely dotted] (-0.5, -0.53) -- node [below, midway] {\normalfont 1} (0.5, -0.53);
\draw [ultra thin] (-0.5, -0.55) -- (-0.5, -0.51);
\draw [ultra thin] (0.5, -0.55) -- (0.5, -0.51);
\draw [<->, densely dotted] (0.53, -0.5) -- node [sloped, below, midway] {\normalfont 1} (0.53, 0.5);
\draw [ultra thin] (0.55, -0.5) -- (0.51, -0.5);
\draw [ultra thin] (0.55, 0.5) -- (0.51, 0.5);
\end{tikzpicture}%
  \caption{A two-dimensional illustration of the wall gap experiment. The start
    and goal states are represented by a black dot (\tikzinlinestart) and
    circle (\tikzinlinegoal), respectively. State space obstacles are indicated
    with gray rectangles~(\tikzinlineobstacle). Each state space dimension was
    bounded to the interval \( [0, 1] \).\vspace{1.5em}}%
  \label{fig:results:wall-gap}
\end{figure}

%%% Local Variables:
%%% mode: latex
%%% TeX-master: "../../../main"
%%% End:

\begin{figure*}
  \begin{subfigure}[b]{0.495\textwidth}%
    \input{figures/5-experimental-results/abstract_problems/results/wall_gap_length_r2}%
    \caption{Path length in \(\mathbb{R}^{2}\)}%
    \label{fig:results:wall-gap-path-length-r2}%
  \end{subfigure}%
  \hfill%
  \begin{subfigure}[b]{0.495\textwidth}%
    \input{figures/5-experimental-results/abstract_problems/results/wall_gap_clearance_r2}%
    \caption{Obstacle clearance in \(\mathbb{R}^{2}\)}%
    \label{fig:results:wall-gap-clearance-r2}%
  \end{subfigure}%
  \\[1em]
  \begin{subfigure}[b]{0.495\textwidth}%
    \input{figures/5-experimental-results/abstract_problems/results/wall_gap_length_r8}%
    \caption{Path length in \(\mathbb{R}^{8}\)}%
    \label{fig:results:wall-gap-path-length-r8}%
  \end{subfigure}%
  \hfill%
  \begin{subfigure}[b]{0.495\textwidth}%
    \input{figures/5-experimental-results/abstract_problems/results/wall_gap_clearance_r8}%
    \caption{Obstacle clearance in \(\mathbb{R}^{8}\)}%
    \label{fig:results:wall-gap-clearance-r8}%
  \end{subfigure}%
  \\[1em]
  \begin{subfigure}[b]{0.495\textwidth}%
    \input{figures/5-experimental-results/abstract_problems/results/wall_gap_length_r16}%
    \caption{Path length in \(\mathbb{R}^{16}\)}%
    \label{fig:results:wall-gap-path-length-r16}%
  \end{subfigure}%
  \hfill%
  \begin{subfigure}[b]{0.495\textwidth}%
    \input{figures/5-experimental-results/abstract_problems/results/wall_gap_clearance_r16}%
    \caption{Obstacle clearance in \(\mathbb{R}^{16}\)}%
    \label{fig:results:wall-gap-clearance-r16}%
  \end{subfigure}%
  \\[0.5em]
  \begin{subfigure}[b]{1.0\textwidth}%
    \centering
    \begin{tikzpicture}
\begin{axis} [
  width=\textwidth,
  height=0.5\textwidth,
  unbounded coords=jump,
  xtick align=inside,
  ytick align=inside,
  anchor=north,
  hide axis,
  xmajorgrids,
  ymajorgrids,
  major grid style={densely dotted, black!20},
  xmin=0,
  xmax=10,
  ymin=0,
  ymax=10,
  xlabel style={font=\footnotesize},
  xticklabel style={font=\footnotesize},
  ylabel style={font=\footnotesize},
  yticklabel style={font=\footnotesize},
  legend style={anchor=south, legend cell align=left, legend columns=-1, at={(axis cs:5, 6)}, font=\small}
]
\addlegendimage{espblack, line width = 1.0pt, mark size=1.0pt, mark=square*}
\addlegendentry{RRT-Connect}
\addlegendimage{esppurple, line width = 1.0pt, mark size=1.0pt, mark=square*}
\addlegendentry{RRT*}
\addlegendimage{esplightpurple, line width = 1.0pt, mark size=1.0pt, mark=square*}
\addlegendentry{LBT-RRT}
\addlegendimage{esplightred, line width = 1.0pt, mark size=1.0pt, mark=square*}
\addlegendentry{LazyPRM*}
\addlegendimage{espyellow, line width = 1.0pt, mark size=1.0pt, mark=square*}
\addlegendentry{FMT*}
\addlegendimage{espblue, line width = 1.0pt, mark size=1.0pt, mark=square*}
\addlegendentry{BIT*}
\addlegendimage{esplightblue, line width = 1.0pt, mark size=1.0pt, mark=square*}
\addlegendentry{ABIT*}
\addlegendimage{esplightgreen, line width = 1.0pt, mark size=1.0pt, mark=square*}
\addlegendentry{AIT*}
\addlegendimage{espgreen, line width = 1.0pt, mark size=1.0pt, mark=square*}
\addlegendentry{EIT*}
\end{axis}
\end{tikzpicture}
  \end{subfigure}%
  \caption{The planner performances on the wall gap experiments described in
    \reft[Section]{sec:abstract-problems}
    \refp[Figure]{fig:results:wall-gap}. The success plots show the percentages
    of successful runs over time. The cost plots show the median initial
    solution times and costs as squares and the median solution costs over time
    as thick lines, both with nonparametric 99\% confidence intervals shown as
    error bars and shaded areas, respectively. Unsuccessful runs were taken as
    infinite costs. The results show that EIT* outperforms all other tested
    asymptotically optimal planners for both objectives in terms of success
    rates, median initial solution times, and median solution quality over
    time.}%
  \label{fig:results:wall-gap-results}
\end{figure*}

%%% Local Variables:
%%% mode: latex
%%% TeX-master: "../../../../main"
%%% End:

% ~/phd/tech_reports/TR-2020-MPS001/figures/5-experimental-results/manipulator_arms/
\begin{figure*}
  \begin{subfigure}[b]{0.245\textwidth}
    \includegraphics[width=\textwidth]{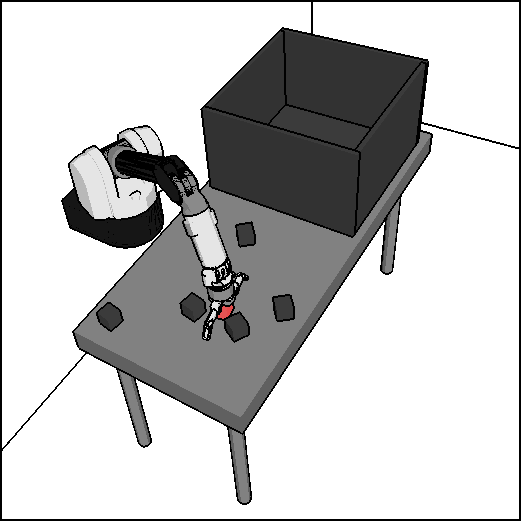}%
    \caption{Start}%
    \label{fig:results:one-arm-start}%
  \end{subfigure}%
  \hfill%
  \begin{subfigure}[b]{0.245\textwidth}%
    \includegraphics[width=\textwidth]{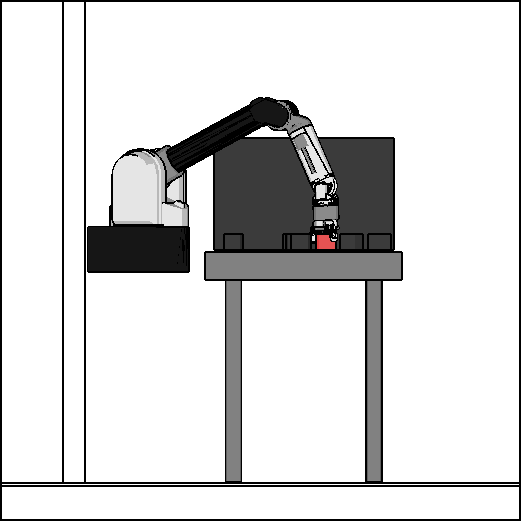}%
    \caption{Start side view}%
    \label{fig:results:one-arm-back-start}%
  \end{subfigure}%
  \hfill%
  \begin{subfigure}[b]{0.245\textwidth}%
    \includegraphics[width=\textwidth]{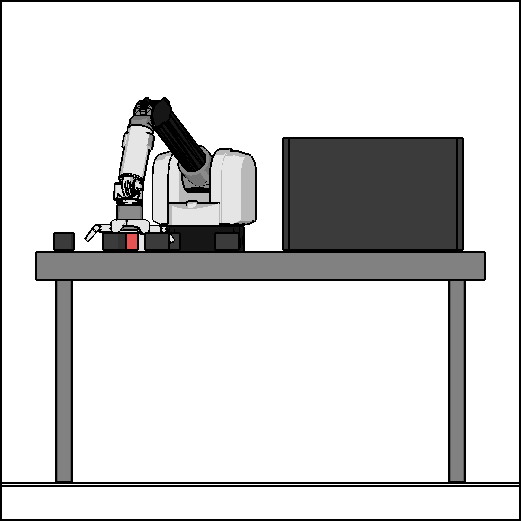}%
    \caption{Start front view}%
    \label{fig:results:one-arm-side-start}%
  \end{subfigure}%
  \hfill%
  \begin{subfigure}[b]{0.245\textwidth}%
    \includegraphics[width=\textwidth]{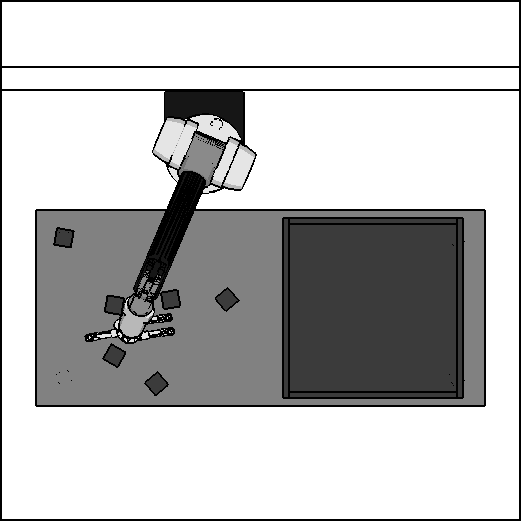}%
    \caption{Start top view}%
    \label{fig:results:one-arm-top-start}%
  \end{subfigure}%
  \\[0.2em]
  \begin{subfigure}[b]{0.245\textwidth}
    \includegraphics[width=\textwidth]{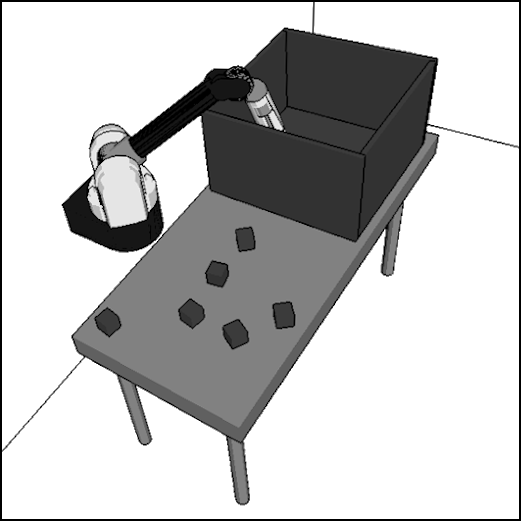}%
    \caption{Goal}%
    \label{fig:results:one-arm-goal}%
  \end{subfigure}%
  \hfill%
  \begin{subfigure}[b]{0.245\textwidth}%
    \includegraphics[width=\textwidth]{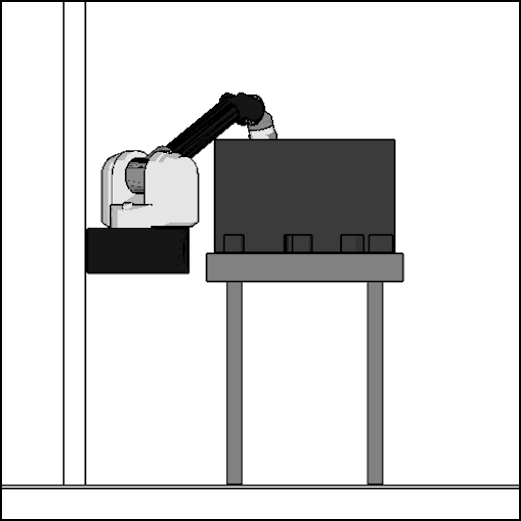}%
    \caption{Goal side view}%
    \label{fig:results:one-arm-back-goal}%
  \end{subfigure}%
  \hfill%
  \begin{subfigure}[b]{0.245\textwidth}%
    \includegraphics[width=\textwidth]{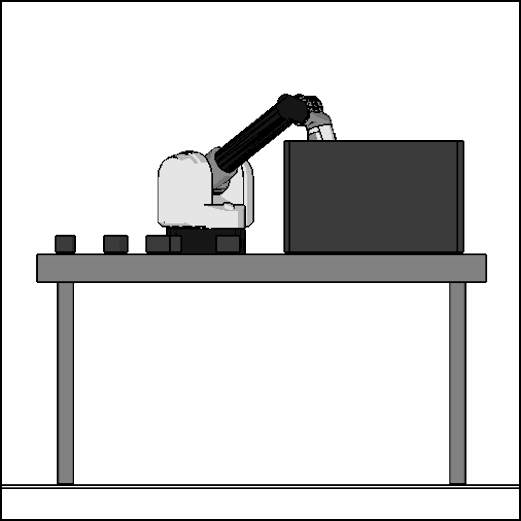}%
    \caption{Goal front view}%
    \label{fig:results:one-arm-side-goal}%
  \end{subfigure}%
  \hfill%
  \begin{subfigure}[b]{0.245\textwidth}%
    \includegraphics[width=\textwidth]{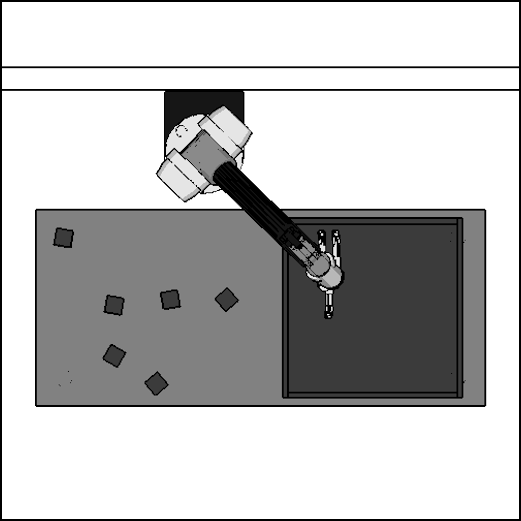}%
    \caption{Goal top view}%
    \label{fig:results:one-arm-top-goal}%
  \end{subfigure}%
  \caption{Illustrations of the single-arm manipulator problem. The top row
    shows the start configuration of the arm in position to pick up the red cube from
    the table~(\subref{fig:results:one-arm-start}--\subref{fig:results:one-arm-top-start}). The
    bottom row shows the goal configuration of the arm in position to place a
    cube in the
    box~(\subref{fig:results:one-arm-goal}--\subref{fig:results:one-arm-top-goal}).\vspace*{5em}}%
  \label{fig:results:one-manipulator-arm}
\end{figure*}

%%% Local Variables:
%%% mode: latex
%%% TeX-master: "../../../main"
%%% End:
\begin{figure*}
  \begin{subfigure}[b]{0.495\textwidth}%
    \input{figures/5-experimental-results/manipulator_arms/results/one_arm/one_arm_length}%
    \caption{Path length}%
    \label{fig:results:one-arm-path-length}%
  \end{subfigure}%
  \hfill%
  \begin{subfigure}[b]{0.495\textwidth}%
    \input{figures/5-experimental-results/manipulator_arms/results/one_arm/one_arm_clearance}%
    \caption{Obstacle clearance}%
    \label{fig:results:one-arm-clearance}%
  \end{subfigure}%
  \\[1em]
  \begin{subfigure}[b]{1.0\textwidth}%
    \centering
    \begin{tikzpicture}
\begin{axis} [
  width=\textwidth,
  height=0.5\textwidth,
  unbounded coords=jump,
  xtick align=inside,
  ytick align=inside,
  anchor=north,
  hide axis,
  xmajorgrids,
  ymajorgrids,
  major grid style={densely dotted, black!20},
  xmin=0,
  xmax=10,
  ymin=0,
  ymax=10,
  xlabel style={font=\footnotesize},
  xticklabel style={font=\footnotesize},
  ylabel style={font=\footnotesize},
  yticklabel style={font=\footnotesize},
  legend style={anchor=south, legend cell align=left, legend columns=-1, at={(axis cs:5, 6)}, font=\small}
]
\addlegendimage{espblack, line width = 1.0pt, mark size=1.0pt, mark=square*}
\addlegendentry{RRT-Connect}
\addlegendimage{esppurple, line width = 1.0pt, mark size=1.0pt, mark=square*}
\addlegendentry{RRT*}
\addlegendimage{esplightpurple, line width = 1.0pt, mark size=1.0pt, mark=square*}
\addlegendentry{LBT-RRT}
\addlegendimage{esplightred, line width = 1.0pt, mark size=1.0pt, mark=square*}
\addlegendentry{LazyPRM*}
\addlegendimage{espyellow, line width = 1.0pt, mark size=1.0pt, mark=square*}
\addlegendentry{FMT*}
\addlegendimage{espblue, line width = 1.0pt, mark size=1.0pt, mark=square*}
\addlegendentry{BIT*}
\addlegendimage{esplightblue, line width = 1.0pt, mark size=1.0pt, mark=square*}
\addlegendentry{ABIT*}
\addlegendimage{esplightgreen, line width = 1.0pt, mark size=1.0pt, mark=square*}
\addlegendentry{AIT*}
\addlegendimage{espgreen, line width = 1.0pt, mark size=1.0pt, mark=square*}
\addlegendentry{EIT*}
\end{axis}
\end{tikzpicture}%
  \end{subfigure}%
  \caption{This figure shows the planner performances on the single-arm
    manipulator experiments described in
    \reft[Section]{sec:single-arm-planning}
    \refp[Figure]{fig:results:one-manipulator-arm}. The success plots show the
    percentages of successful runs over time. The cost plots show the median
    initial solution times and costs as squares and the median solution costs
    over time as thick lines, both with nonparametric 99\% confidence intervals
    shown as error bars and shaded areas, respectively. Unsuccessful runs were
    taken as infinite costs. The results show that for the path length version
    of this simple problem, \ac{EIT*} and \ac{AIT*} have near identical
    performances to Lazy \ac{PRM}*, \ac{FMT*}, \ac{BIT*}, and \ac{ABIT*}, which
    outperform \ac{LBT-RRT} and \acs{RRT*}
    (\subref{fig:results:one-arm-path-length}). In the obstacle clearance
    version of the problem, \ac{EIT*} and \ac{AIT*} outperform all other
    almost-surely asymptotically optimal planners
    (\subref{fig:results:one-arm-clearance}).}%
  \label{fig:results:one-arm-results}
\end{figure*}

%%% Local Variables:
%%% mode: latex
%%% TeX-master: "../../../../main"
%%% End:

\begin{figure*}
  \begin{subfigure}[b]{0.245\textwidth}
    \includegraphics[width=\textwidth]{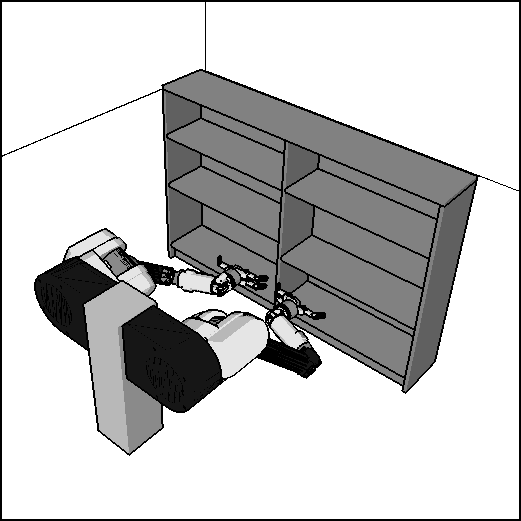}%
    \caption{Start}%
    \label{fig:results:two-arm-start}%
  \end{subfigure}%
  \hfill%
  \begin{subfigure}[b]{0.245\textwidth}%
    \includegraphics[width=\textwidth]{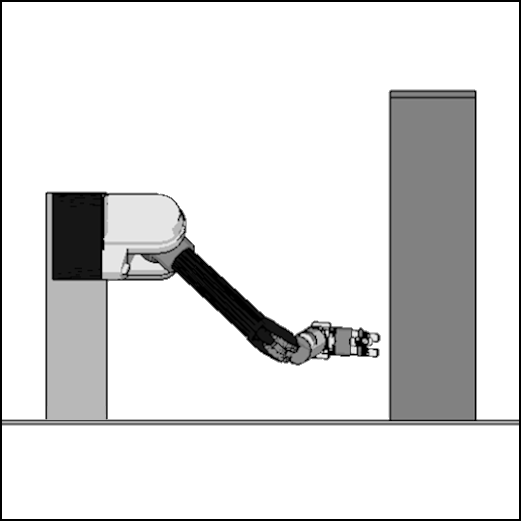}%
    \caption{Start side view}%
    \label{fig:results:two-arm-back-start}%
  \end{subfigure}%
  \hfill%
  \begin{subfigure}[b]{0.245\textwidth}%
    \includegraphics[width=\textwidth]{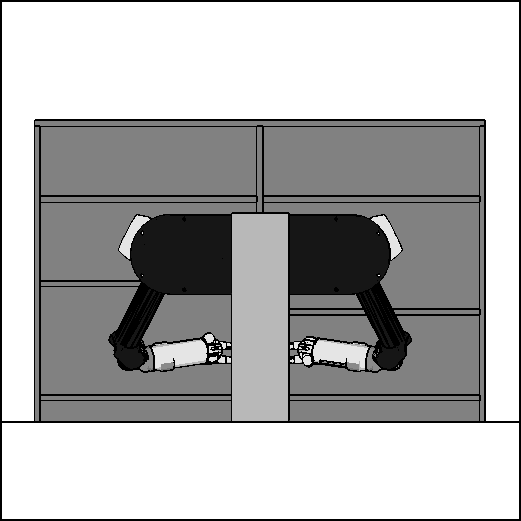}%
    \caption{Start back view}%
    \label{fig:results:two-arm-side-start}%
  \end{subfigure}%
  \hfill%
  \begin{subfigure}[b]{0.245\textwidth}%
    \includegraphics[width=\textwidth]{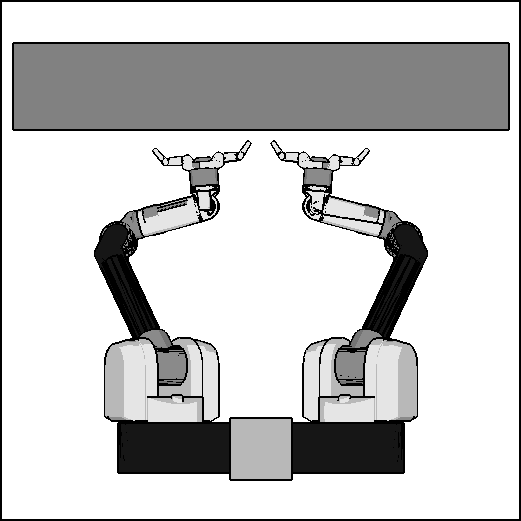}%
    \caption{Start top view}%
    \label{fig:results:two-arm-top-start}%
  \end{subfigure}%
  \\[0.2em]
  \begin{subfigure}[b]{0.245\textwidth}
    \includegraphics[width=\textwidth]{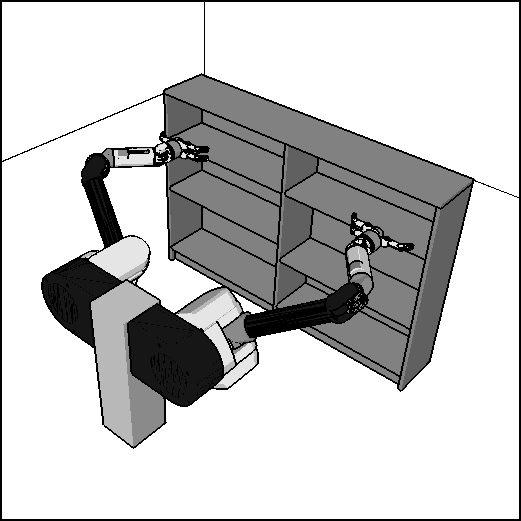}%
    \caption{Goal}%
    \label{fig:results:two-arm-goal}%
  \end{subfigure}%
  \hfill%
  \begin{subfigure}[b]{0.245\textwidth}%
    \includegraphics[width=\textwidth]{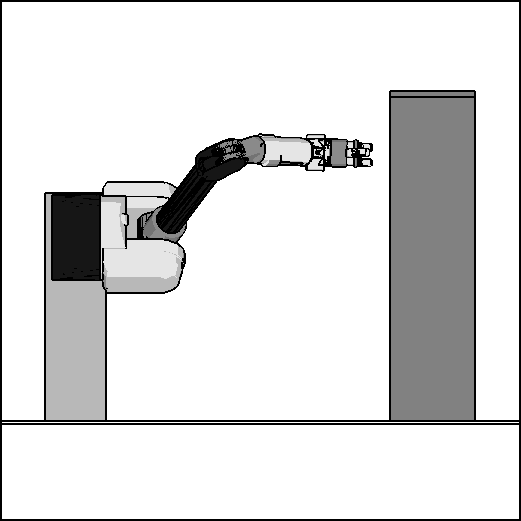}%
    \caption{Goal side view}%
    \label{fig:results:two-arm-back-goal}%
  \end{subfigure}%
  \hfill%
  \begin{subfigure}[b]{0.245\textwidth}%
    \includegraphics[width=\textwidth]{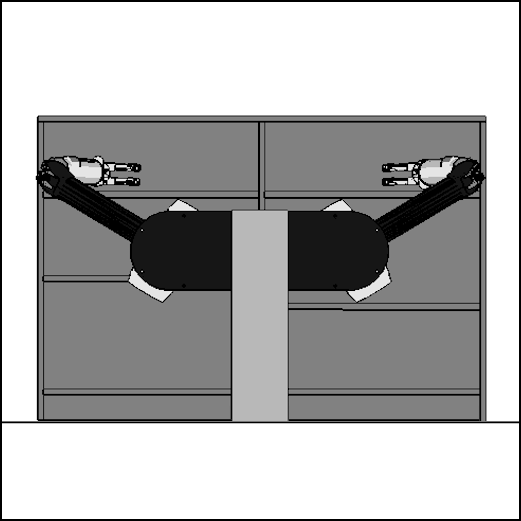}%
    \caption{Goal back view}%
    \label{fig:results:two-arm-side-goal}%
  \end{subfigure}%
  \hfill%
  \begin{subfigure}[b]{0.245\textwidth}%
    \includegraphics[width=\textwidth]{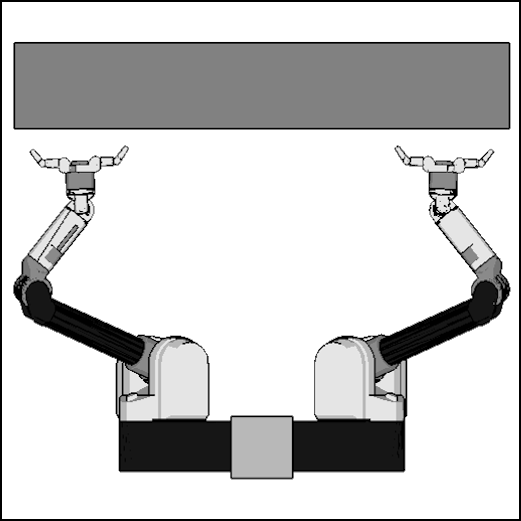}%
    \caption{Goal configuration top view}%
    \label{fig:results:two-arm-top-goal}%
  \end{subfigure}%
  \caption{Illustrations of the dual-arm manipulator problem. The
    top row shows the start configuration of the arms in position to pick up an
    object on the bottom
    shelf~(\subref{fig:results:two-arm-start}--\subref{fig:results:two-arm-top-start}). The
    bottom row shows the goal configuration of the arms in position to place an
    object on the top
    shelf~(\subref{fig:results:two-arm-goal}--\subref{fig:results:two-arm-top-goal}).\vspace*{5em}}%
  \label{fig:results:two-manipulator-arms}
\end{figure*}

%%% Local Variables:
%%% mode: latex
%%% TeX-master: "../../../main"
%%% End:
\begin{figure*}
  \begin{subfigure}[b]{0.495\textwidth}%
    \input{figures/5-experimental-results/manipulator_arms/results/two_arm/two_arm_length}%
    \caption{Path length}%
    \label{fig:results:two-arm-path-length}%
  \end{subfigure}%
  \hfill%
  \begin{subfigure}[b]{0.495\textwidth}%
    \input{figures/5-experimental-results/manipulator_arms/results/two_arm/two_arm_clearance}%
    \caption{Optimizing obstacle}%
    \label{fig:results:two-arm-clearance}%
  \end{subfigure}%
  \\[1em]
  \begin{subfigure}[b]{1.0\textwidth}%
    \centering
    \begin{tikzpicture}
\begin{axis} [
  width=\textwidth,
  height=0.5\textwidth,
  unbounded coords=jump,
  xtick align=inside,
  ytick align=inside,
  anchor=north,
  hide axis,
  xmajorgrids,
  ymajorgrids,
  major grid style={densely dotted, black!20},
  xmin=0,
  xmax=10,
  ymin=0,
  ymax=10,
  xlabel style={font=\footnotesize},
  xticklabel style={font=\footnotesize},
  ylabel style={font=\footnotesize},
  yticklabel style={font=\footnotesize},
  legend style={anchor=south, legend cell align=left, legend columns=-1, at={(axis cs:5, 6)}, font=\small}
]
\addlegendimage{espblack, line width = 1.0pt, mark size=1.0pt, mark=square*}
\addlegendentry{RRT-Connect}
\addlegendimage{esppurple, line width = 1.0pt, mark size=1.0pt, mark=square*}
\addlegendentry{RRT*}
\addlegendimage{esplightpurple, line width = 1.0pt, mark size=1.0pt, mark=square*}
\addlegendentry{LBT-RRT}
\addlegendimage{esplightred, line width = 1.0pt, mark size=1.0pt, mark=square*}
\addlegendentry{LazyPRM*}
\addlegendimage{espyellow, line width = 1.0pt, mark size=1.0pt, mark=square*}
\addlegendentry{FMT*}
\addlegendimage{espblue, line width = 1.0pt, mark size=1.0pt, mark=square*}
\addlegendentry{BIT*}
\addlegendimage{esplightblue, line width = 1.0pt, mark size=1.0pt, mark=square*}
\addlegendentry{ABIT*}
\addlegendimage{esplightgreen, line width = 1.0pt, mark size=1.0pt, mark=square*}
\addlegendentry{AIT*}
\addlegendimage{espgreen, line width = 1.0pt, mark size=1.0pt, mark=square*}
\addlegendentry{EIT*}
\end{axis}
\end{tikzpicture}%
  \end{subfigure}%
  \caption{This figure shows the planner performances on the dual-arm
    manipulator experiments described in \reft[Section]{sec:dual-arm-planning}
    \refp[Figure]{fig:results:two-manipulator-arms}. The success plots show the
    percentages of successful runs over time. The cost plots show the median
    initial solution times and costs as squares and the median solution costs
    over time as thick lines, both with nonparametric 99\% confidence intervals
    shown as error bars and shaded areas, respectively. Unsuccessful runs were
    taken as infinite costs. The results show when optimizing path length,
    \ac{AIT*} performs nearly identically to \ac{BIT*} and \ac{ABIT*}, which
    all outperform Lazy \ac{PRM}*, \ac{FMT*}, and \ac{EIT*}, and clearly
    outperform \ac{LBT-RRT} and \ac{RRT*}
    (\subref{fig:results:one-arm-path-length}). When optimizing obstacle
    clearance, \ac{EIT*} and \ac{AIT*} outperform all other almost-surely
    asymptotically optimal planners (\subref{fig:results:two-arm-clearance}).}%
  \label{fig:results:two-arm-results}
\end{figure*}

%%% Local Variables:
%%% mode: latex
%%% TeX-master: "../../../../main"
%%% End:

\begin{figure*}
  \begin{subfigure}[t]{0.245\textwidth}
    \includegraphics[width=\textwidth]{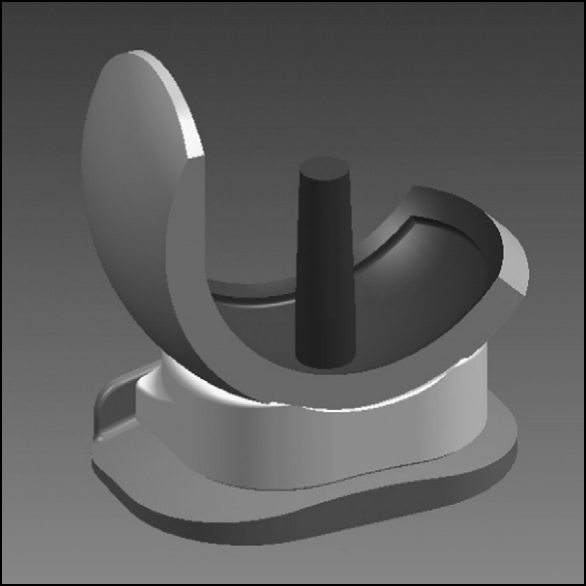}%
    \caption{Oxford Domed Lateral \acs{UKR}\\\citep{pandit_knee2010}}%
    \label{fig:results:knee-original}%
  \end{subfigure}%
  \hfill%
  \begin{subfigure}[t]{0.245\textwidth}%
    \includegraphics[width=\textwidth]{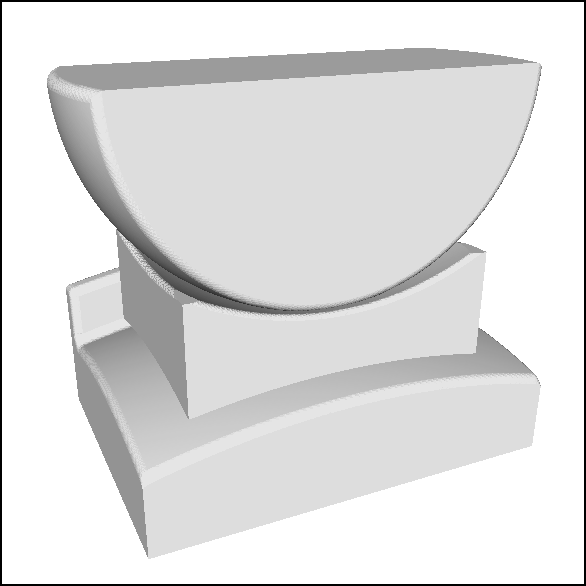}%
    \caption{Approximation used in experiments}%
    \label{fig:results:knee-approximation}%
  \end{subfigure}%
  \hfill%
  \begin{subfigure}[t]{0.245\textwidth}%
    \includegraphics[width=\textwidth]{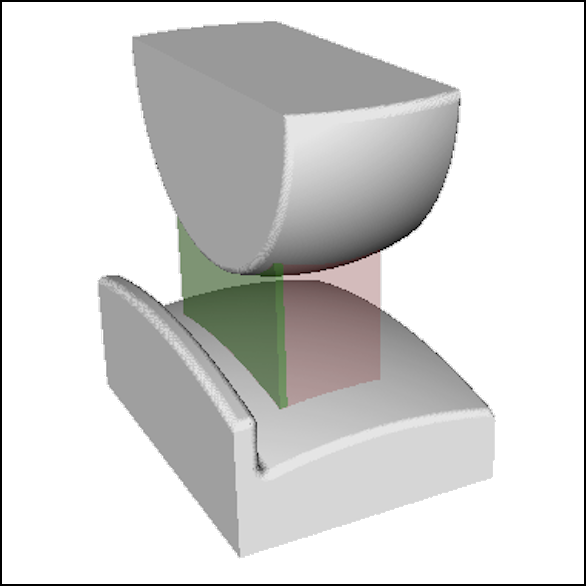}%
    \caption{Goal region used in experiments}%
    \label{fig:results:knee-goal-region}%
  \end{subfigure}%
  \hfill%
  \begin{subfigure}[t]{0.245\textwidth}%
    \includegraphics[width=\textwidth]{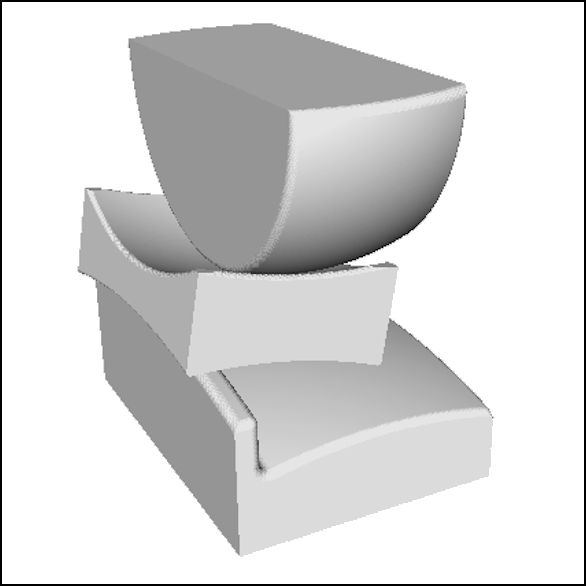}%
    \caption{Example goal configuration}%
    \label{fig:results:knee-goal-example}%
  \end{subfigure}%
  \caption{A multiview illustration of the knee replacement dislocation experiment. The 3D model of the Oxford Domed Lateral \acs{UKR} is reproduced from Figure 1 in \citet{pandit_knee2010}~(\subref{fig:results:knee-original}). The experiments presented in this paper used a simplified approximation of the Oxford Domed Lateral \acs{UKR}~(\subref{fig:results:knee-approximation}).
  The start state of the mobile bearing is between the two fixed parts~(\subref{fig:results:knee-approximation}) with the search space and goal region shown as red and green regions, respectively~(\subref{fig:results:knee-goal-region}). The goal region is the set of bearing positions in medial dislocation between the tibial and femoral components~(\subref{fig:results:knee-goal-example}).\vspace{1em}}%
  \label{fig:results:knee}
\end{figure*}

%%% Local Variables:
%%% mode: latex
%%% TeX-master: "../../../main"
%%% End:
\begin{figure*}
  \begin{subfigure}[b]{0.495\textwidth}%
    \input{figures/5-experimental-results/knee/results/knee_length}%
    \caption{Path length}%
    \label{fig:results:knee-path-length}%
  \end{subfigure}%
  \hfill%
  \begin{subfigure}[b]{0.495\textwidth}%
    \input{figures/5-experimental-results/knee/results/knee_clearance}%
    \caption{Obstacle clearance}%
    \label{fig:results:knee-clearance}%
  \end{subfigure}%
  \\[1em]
  \begin{subfigure}[b]{1.0\textwidth}%
    \centering
    \begin{tikzpicture}
\begin{axis} [
  width=\textwidth,
  height=0.5\textwidth,
  unbounded coords=jump,
  xtick align=inside,
  ytick align=inside,
  anchor=north,
  hide axis,
  xmajorgrids,
  ymajorgrids,
  major grid style={densely dotted, black!20},
  xmin=0,
  xmax=10,
  ymin=0,
  ymax=10,
  xlabel style={font=\footnotesize},
  xticklabel style={font=\footnotesize},
  ylabel style={font=\footnotesize},
  yticklabel style={font=\footnotesize},
  legend style={anchor=south, legend cell align=left, legend columns=-1, at={(axis cs:5, 6)}, font=\small}
]
\addlegendimage{black!50, line width = 1.0pt, mark size=1.0pt, mark=square*}
\addlegendentry{RRT}
\addlegendimage{esppurple, line width = 1.0pt, mark size=1.0pt, mark=square*}
\addlegendentry{RRT*}
\addlegendimage{esplightred, line width = 1.0pt, mark size=1.0pt, mark=square*}
\addlegendentry{LazyPRM*}
\addlegendimage{espyellow, line width = 1.0pt, mark size=1.0pt, mark=square*}
\addlegendentry{FMT*}
\addlegendimage{espblue, line width = 1.0pt, mark size=1.0pt, mark=square*}
\addlegendentry{BIT*}
\addlegendimage{esplightblue, line width = 1.0pt, mark size=1.0pt, mark=square*}
\addlegendentry{ABIT*}
\addlegendimage{esplightgreen, line width = 1.0pt, mark size=1.0pt, mark=square*}
\addlegendentry{AIT*}
\addlegendimage{espgreen, line width = 1.0pt, mark size=1.0pt, mark=square*}
\addlegendentry{EIT*}
\end{axis}
\end{tikzpicture}%
  \end{subfigure}%
  \caption{The planner performances on the knee replacement dislocation problem
    described in \reft[Section]{sec:knee-implants}
    \refp[Figure]{fig:results:knee}. The success plots show the percentages of
    successful runs over time. The cost plots show the median initial solution
    times and costs as squares and the median solution costs over time as thick
    lines, both with nonparametric 99\% confidence intervals shown as error
    bars and shaded areas, respectively. Unsuccessful runs were taken as
    infinite costs. The results show that when optimizing path length,
    \ac{EIT*} and \ac{AIT*} outperform all other planners in terms of success
    rates, median initial solution times, and median solution quality over
    time. Lazy \ac{PRM}* finds solutions as fast as \ac{EIT*} for some random
    sequences of samples, but reaches the time limit before finding any
    solution for other random sequences. When optimizing obstacle clearance,
    \ac{EIT*} again outperforms all other tested planners in terms of the above
    measures. \ac{RRT}-Connect and \ac{LBT-RRT} were not tested in this
    experiment, as their available OMPL implementations do not support goal
    regions.}%
  \label{fig:results:knee-results}
\end{figure*}

%%% Local Variables:
%%% mode: latex
%%% TeX-master: "../../../../main"
%%% End:

%%% Local Variables:
%%% mode: latex
%%% TeX-master: "../main"
%%% End:

% Discussion
\section{Discussion}%
\label{sec:discussion}

The experiments presented in \reft[Section]{sec:experimental-results} demonstrate
the benefits of sampling-based path planning with an asymmetric bidirectional
search. This section discusses the results of these experiments, elaborates on
the algorithmic differences between \ac{AIT*} and \ac{EIT*}, and presents
possible extensions to asymmetric bidirectional algorithms in sampling-based
path planning and beyond.

\reft[Section]{sec:experimental-results} shows the performance of \ac{AIT*},
\ac{EIT*}, and eight other sampling-based algorithms on a diverse set of twelve
problems optimizing two objectives. When optimizing path length, the
experiments show that \ac{AIT*} and \ac{EIT*} are competitive to the other
tested planners in terms of initial solution times, success rates, and solution
quality over time.

When optimizing obstacle clearance, the experiments show that \ac{EIT*}
outperforms all other tested asymptotically optimal planners by finding initial
solutions faster, reaching 100\% success rates sooner, and providing the
highest quality solution for most of the time. \ac{AIT*} is often the
second-best performing planner on this objective and even competitive to
\ac{EIT*} on the robotic arm experiments.

The batch size of \ac{AIT*} and \ac{EIT*} was kept constant for all presented
experiments, while the performance of \ac{RRT}-based planners was tuned to the
problem dimension by adjusting the maximum edge length.
This shows that \ac{AIT*} and \ac{EIT*} can perform well without
problem-specific batch sizes but can also motivate future research to investigate advanced batch-size calculations, including variable and adaptive batch sizes.

The experiments presented in \reft[Section]{sec:experimental-results} keep the
collision detection resolution constant within each problem. This resolution
determines the false negative collision rate, i.e., the percentage of edges
that are considered valid but in reality are not. What is considered an
acceptable false negative collision rate depends on the application of the
planning algorithm. Experiments not presented in this paper showed that the
\emph{relative} performance of \ac{AIT*} and \ac{EIT*} compared to other
algorithms improves as edge evaluation becomes more computationally expensive,
e.g., due to finer collision detection resolution or more complex
analysis. This may be because \ac{AIT*} and \ac{EIT*} often fully evaluate
fewer edges than other algorithms and their performance is therefore not as
sensitive to the collision detection resolution.
If edge evaluation is computationally inexpensive, e.g., due to coarse
collision detection resolution, then the benefits of the accurate heuristics calculated in \ac{AIT*} and \ac{EIT*} may not justify the computational cost required to calculate them and other sampling-based planners may perform better in these cases.

Edge evaluation is also computationally expensive for systems with kinodynamic constraints, when full edge evaluation requires solving a two-point \ac{BVP}.
The accurate heuristics calculated by \ac{AIT*} and \ac{EIT*} can reduce the number of \acp{BVP} solved, but \ac{AIT*} and \ac{EIT*} require exact solutions to the \acp{BVP}.
Other almost-surely asymptotically optimal algorithms do not require exact \ac{BVP} solutions even for problems with kinodynamic constraints~\citep{li_ijrr2016, hauser_tro2016, kleinbort_icra2020, shome_icra2021}.

\ac{AIT*} and \ac{EIT*} use different algorithms for the reverse and forward searches of their asymmetric bidirectional search~(Table~\ref{tbl:forward-and-reverse-searches}). The change in forward search algorithms from A* in \ac{AIT*} to \ac{AEES} in \ac{EIT*} is motivated by the benefits of effort heuristics, which cannot be exploited with A*, and justify the computational overhead induced by the increased complexity of \ac{AEES}. The change in reverse search algorithms from \ac{LPA*} in \ac{AIT*} to A* in \ac{EIT*} is motivated by the observations that repairing the reverse search tree with \ac{LPA*} is only more efficient than restarting A* for small changes in the search tree~\citep[between 1\% and 2\%; Table 1,][]{aine_ai2016}, and that detecting invalid edges with the forward search often results in larger changes in the reverse search tree. Implementing increasingly dense collision detection in an \ac{LPA*} reverse search would additionally require either further bookkeeping to keep track of which portion of the tree was checked with which resolution or result in duplicated collision detection effort on some of the edges.

The presented asymmetric bidirectional approach in which two searches inform
each other with complementary information can potentially be beneficial in all
problem domains where full edge evaluation is computationally expensive, e.g.,
because of computationally expensive true edge cost computation or complex
collision detection. If this edge cost computation can inexpensively be
approximated by a heuristic, then the reverse search can combine such
heuristics between multiple states into more accurate heuristics between each
state and the goal, similar to \ac{AIT*} and \ac{EIT*}.

%%% Local Variables:
%%% mode: latex
%%% TeX-master: "../main"
%%% End:

% Conclusion
\section{Conclusion}%
\label{sec:conclusion}

Informed path planning algorithms can use problem-specific knowledge in the
form of heuristics to improve their performance, but selecting appropriate
heuristics is difficult. This is because heuristics are most effective when
they are both accurate and computationally inexpensive to evaluate, which are
often conflicting characteristics. Many informed planners additionally can only
use problem-specific knowledge if it is expressible as admissible heuristics,
which is not always possible for all optimization objectives.

This paper presents two almost-surely asymptotically optimal path planning
algorithms that address these challenges by using asymmetric bidirectional
searches that simultaneously calculate and exploit accurate, problem-specific
heuristics. \ac{AIT*} uses an inexpensive reverse search to combine admissible
\textit{a priori} cost heuristics between two states into a more accurate but
still admissible cost heuristic between each state and the goal. This heuristic
is exploited to make \ac{AIT*}'s forward search more efficient and repaired
whenever the forward search detects that it uses invalid edges. In this way,
information is passed between both directions of the bidirectional search, as
each search informs the other.

\ac{EIT*} builds on \ac{AIT*} by additionally calculating inadmissible cost and
effort heuristics with its reverse search. This additional knowledge about the
computational effort to validate a path can be calculated for any optimization
objective and exploited by \ac{EIT*}'s forward search to order its search in a
solution-oriented manner. This improves anytime performance when the cost of a
path does not correlate well with the computational effort required to validate
it and allows \ac{EIT*} to find initial solutions quickly, even if an
admissible cost heuristic is not available for an optimization objective.

The benefits of simultaneously calculating and exploiting ever more accurate
heuristics through an asymmetric bidirectional search are demonstrated on
twelve diverse problems in abstract, robotic, and biomedical domains. \ac{EIT*}
outperforms all other tested asymptotically optimal planners when optimizing
obstacle clearance and performs competitively when optimizing path
length. \ac{AIT*} is often the second best performing asymptotically optimal
planner when optimizing obstacle clearance and also performs competitively when
optimizing path length.

Information on the \ac{OMPL} implementations of \ac{AIT*} and \ac{EIT*} as well
as the software framework for running the experiments and creating the
corresponding plots will be available at \url{https://robotic-esp.com/code/}.

%%% Local Variables:
%%% mode: latex
%%% TeX-master: "../main"
%%% End:

% Acknowledgments
\section{Acknowledgments}%
\label{sec:acknowledgements}

This research was funded by UK Research and Innovation and EPSRC through Robotics and Artificial Intelligence for Nuclear (RAIN) [EP/R026084/1] and ACE-OPS:\@ From Autonomy to Cognitive assistance in Emergency OPerationS [EP/S030832/1]. We thank Irene Yang and Stephen J. Mellon for discussions and guidance regarding medial dislocations of the Oxford Domed Lateral \ac{UKR} implant.
We also thank the editorial board for considering this manuscript and our reviewers for their time and constructive criticism.

%%% Local Variables:
%%% mode: latex
%%% TeX-master: "../main"
%%% End:

% Multimedia appendix
\begin{appendices}
  \section{Multimedia Extensions}%
\label{sec:multimedia-extensions}

A YouTube playlist that complements this article can be found at
\href{https://youtube.com/playlist?list=PLbaQBz4TuPczfN6PN6NkfmlnXpcf79Aq\_}{\small\texttt{https://youtube.com/playlist?list=
PLbaQBz4TuPczfN6PN6NkfmlnXpcf79Aq\_}}.

\begin{table}[h!]
    \begin{tabular}{rlp{5cm}}
        \toprule
        Extension & Type & Description \\
        \midrule
        1         & Video & AIT* compared to RRT*, FMT*, and BIT*,
                            optimizing path length \\
        2         & Video & EIT* compared to RRT*, BIT*, and AIT*,
                            optimizing obstacle clearance \\
        \bottomrule
      \end{tabular}\vspace{0.5em}
    \caption{Index to multimedia extensions.}
\end{table}
\vspace{-2em}

%%% Local Variables:
%%% mode: latex
%%% TeX-master: "../main"
%%% End:

\end{appendices}

% Bibliography
\bibliographystyle{SageH}
\interlinepenalty=10000
\bibliography{\string~/bibliography/bibliography}

\end{document}